\documentclass[10pt,twocolumn,letterpaper]{article}

\usepackage[pagenumbers]{cvpr} 

\usepackage[T1]{fontenc}
\usepackage{amsthm,mathtools}
\usepackage{multirow,array}
\usepackage{tikz}
\usetikzlibrary{arrows.meta,positioning,fit,calc,backgrounds,patterns,shapes.geometric}
\usepackage{pgfplots}
\pgfplotsset{compat=1.18}
\usepackage[ruled,vlined,linesnumbered]{algorithm2e}
\SetAlFnt{\footnotesize}\SetAlCapFnt{\footnotesize}\SetAlCapNameFnt{\footnotesize}
\SetKwComment{Comment}{$\triangleright$\ }{}

\SetCommentSty{algcomment}
\graphicspath{{figs/}}

\newtheorem{proposition}{Proposition}
\newtheorem{lemma}{Lemma}
\newtheorem{corollary}{Corollary}

\definecolor{nncol}{HTML}{E69F00}
\definecolor{mapcol}{HTML}{0072B2}
\definecolor{hocol}{HTML}{56B4E9}

\newcommand{\best}[1]{\textbf{#1}}
\newcommand{\second}[1]{\underline{#1}}

\pgfplotsset{
  paper/.style={
    axis line style={black!60, line width=0.5pt}, tick style={black!60, line width=0.4pt},
    major tick length=2pt, ymajorgrids, grid style={black!12, line width=0.3pt}, xmajorgrids=false,
    tick label style={font=\scriptsize}, label style={font=\scriptsize}, title style={font=\scriptsize},
    legend style={font=\scriptsize, draw=none, fill=white, fill opacity=0.85, text opacity=1, inner sep=1pt,
                  row sep=-1pt},
    legend cell align=left, every axis plot/.append style={line width=0.8pt},
    axis lines*=left, clip=false},
}

\newcommand{\oursname}{RecaP}
\newcommand{\ours}{\mbox{\oursname}\xspace}
\newcommand{\NN}{NN\xspace}
\newcommand{\RX}{\mbox{\oursname}\textsubscript{0}\xspace}
\newcommand{\recapplotsubscript}[1]{\ensuremath{{}_{\text{#1}}}}
\newcommand{\RXplot}{\oursname\recapplotsubscript{0}}
\newcommand{\MAP}{\mbox{\oursname}\xspace}
\newcommand{\MAPo}{\mbox{\oursname}\textsubscript{own}\xspace}
\newcommand{\MAPx}{\mbox{\oursname}\textsubscript{cross}\xspace}
\newcommand{\KDNN}{NN+KD\xspace}
\newcommand{\KDMAP}{\mbox{\oursname}+KD\xspace}
\newcommand{\lpipsA}{LPIPS-A\xspace}
\newcommand{\INh}{\mbox{IN-100}\xspace}
\newcommand{\INk}{\mbox{IN-1k}\xspace}

\newcommand{\R}{\mathbb{R}}
\newcommand{\E}{\mathbb{E}}
\newcommand{\Hm}{\mathbf{H}}
\newcommand{\T}{^{\!\top}}
\DeclareMathOperator*{\argmin}{arg\,min}
\DeclareMathOperator*{\argmax}{arg\,max}
\DeclareMathOperator{\softmax}{softmax}

\definecolor{cvprblue}{rgb}{0.21,0.49,0.74}
\usepackage[pagebackref,breaklinks,colorlinks,allcolors=cvprblue]{hyperref}

\usepackage[capitalize,noabbrev]{cleveref}

\newcommand{\appxtag}{App.}
\newcommand{\appxtags}{Apps.}
\newcommand{\Appxtag}{Appendix}
\newcommand{\Appxtags}{Appendices}
\newcommand{\maintag}{Main}
\newcommand{\Maintag}{Main}
\newcommand{\appxsecstyle}{\arabic}
\newif\ifappxSnumbers
\appxSnumbersfalse

\newif\ifinappendix
\protected\def\toappx{\ifinappendix\else\appxtag~\fi}
\protected\def\Toappx{\ifinappendix\else\Appxtag~\fi}
\protected\def\tomain{\ifinappendix\maintag~\fi}
\protected\def\Tomain{\ifinappendix\Maintag~\fi}
\protected\def\appxsec{\ifinappendix Sec.\else\appxtag\fi}
\protected\def\appxsecs{\ifinappendix Secs.\else\appxtags\fi}
\protected\def\Appxsec{\ifinappendix Section\else\Appxtag\fi}
\protected\def\Appxsecs{\ifinappendix Sections\else\Appxtags\fi}

\makeatletter
\newcommand{\appendixrefs}{%
  \global\inappendixtrue
  \renewcommand{\thesection}{\appxsecstyle{section}}%
  \ifappxSnumbers\gdef\appx@numprefix{S}\else\global\let\appx@numprefix\@empty\fi
  \appx@renumber{figure}\appx@renumber{table}\appx@renumber{equation}\appx@renumber{algocf}%
  \appx@renumber{theorem}\appx@renumber{proposition}\appx@renumber{lemma}\appx@renumber{corollary}%
  \crefalias{section}{appendix}%
  \crefalias{subsection}{subappendix}%
  \crefalias{subsubsection}{subsubappendix}%
  \crefalias{figure}{appfigure}%
  \crefalias{table}{apptable}%
  \crefalias{equation}{appequation}%
  \crefalias{algocf}{appalgorithm}%
  \crefalias{algorithm}{appalgorithm}%
  \crefalias{theorem}{apptheorem}%
  \crefalias{proposition}{appproposition}%
  \crefalias{lemma}{applemma}%
  \crefalias{corollary}{appcorollary}}
\newcommand{\appx@renumber}[1]{%
  \@ifundefined{c@#1}{}{%
    \setcounter{#1}{0}%
    \@namedef{the#1}{\appx@numprefix\arabic{#1}}%
    \@namedef{theH#1}{S\arabic{#1}}}}

\newcommand{\appx@names}[6]{%
  \crefname{#1}{\tomain #3}{\tomain #4}\Crefname{#1}{\Tomain #5}{\Tomain #6}%
  \crefname{#2}{\toappx #3}{\toappx #4}\Crefname{#2}{\Toappx #5}{\Toappx #6}}
\AtEndPreamble{%
  \appx@names{section}{appendix}{Sec.}{Secs.}{Section}{Sections}%
  \appx@names{figure}{appfigure}{Fig.}{Figs.}{Figure}{Figures}%
  \appx@names{table}{apptable}{Tab.}{Tabs.}{Table}{Tables}%
  \appx@names{equation}{appequation}{Eq.}{Eqs.}{Equation}{Equations}%
  \appx@names{algorithm}{appalgorithm}{Alg.}{Algs.}{Algorithm}{Algorithms}%
  \appx@names{theorem}{apptheorem}{Thm.}{Thms.}{Theorem}{Theorems}%
  \appx@names{proposition}{appproposition}{Prop.}{Props.}{Proposition}{Propositions}%
  \appx@names{lemma}{applemma}{Lem.}{Lems.}{Lemma}{Lemmas}%
  \appx@names{corollary}{appcorollary}{Cor.}{Cors.}{Corollary}{Corollaries}%
  \crefname{appendix}{\appxsec}{\appxsecs}\Crefname{appendix}{\Appxsec}{\Appxsecs}%
  \creflabelformat{appequation}{\textup{(#2#1#3)}}%
}
\makeatother

\def\confName{CVPR}
\def\confYear{2027}

\title{Same Tokenizer, Better Tokens:\\Choosing Image Tokens by MAP Inference}

\author{
Piotr Kubaty$^{1,2}$\quad
Jacek Tabor$^{1,3}$\quad
\L{}ukasz Struski$^{1}$\\[0.5em]
{\normalsize $^{1}$Faculty of Mathematics and Computer Science, Jagiellonian University, Krak\'{o}w, Poland}\\
{\normalsize $^{2}$Doctoral School of Exact and Natural Sciences, Jagiellonian University, Krak\'{o}w, Poland}\\
{\normalsize $^{3}$Centre for Credible Artificial Intelligence, Warsaw University of Technology, Poland}
}

\begin{document}
\twocolumn[{%
\renewcommand\twocolumn[1][]{#1}%
\maketitle
\begin{center}
\captionsetup{type=figure}
\def\tzPsnrNNA{20.21}\def\tzPsnrMapA{21.42}%
\def\tzPsnrNNB{18.73}\def\tzPsnrMapB{20.34}%

\providecommand{\oursLab}[1]{\textcolor{mapcol}{\textbf{#1}}}%
\providecommand{\nnLab}[1]{\textcolor{nncol!85!black}{\textbf{#1}}}%
\begin{tikzpicture}[font=\footnotesize]
\foreach \x/\lab in {0/{Original}, 5.82/{\oursLab{\oursname{} (ours)}}, 11.64/{\nnLab{Nearest neighbour (\NN)}}}
  \node[anchor=south, inner sep=2pt, text height=1.5ex, text depth=0.4ex] at (\x+2.85,0.04) {\lab};
\node[anchor=north west, inner sep=0, draw=black!40, line width=0.5pt] at (0,0)
  {\includegraphics[width=5.7cm, height=3.1cm]{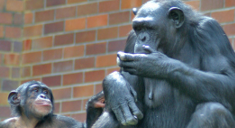}};
\node[anchor=north west, inner sep=0, draw=mapcol, line width=1.2pt] at (5.82,0)
  {\includegraphics[width=5.7cm, height=3.1cm]{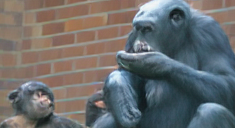}};
\node[anchor=north west, inner sep=0, draw=nncol, line width=1.0pt] at (11.64,0)
  {\includegraphics[width=5.7cm, height=3.1cm]{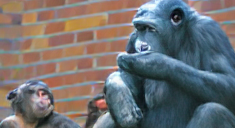}};
\node[anchor=north west, inner sep=0, draw=black!40, line width=0.5pt] at (0,-3.23)
  {\includegraphics[width=5.7cm, height=3.1cm]{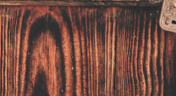}};
\node[anchor=north west, inner sep=0, draw=mapcol, line width=1.2pt] at (5.82,-3.23)
  {\includegraphics[width=5.7cm, height=3.1cm]{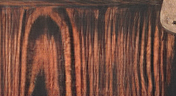}};
\node[anchor=north west, inner sep=0, draw=nncol, line width=1.0pt] at (11.64,-3.23)
  {\includegraphics[width=5.7cm, height=3.1cm]{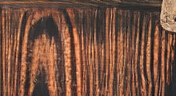}};
\foreach \x/\y/\v in {5.82/-3.10/\tzPsnrMapA, 11.64/-3.10/\tzPsnrNNA, 5.82/-6.33/\tzPsnrMapB, 11.64/-6.33/\tzPsnrNNB}
  \node[anchor=south east, text=white, fill=black, fill opacity=0.6, text opacity=1, inner sep=1.5pt]
    at (\x+5.66,\y+0.04) {PSNR \v\,dB};
\end{tikzpicture}
\captionof{figure}{\textbf{Reconstruction with a frozen LlamaGen VQ-16 tokenizer.}
  \ours ($\lambda=0.01$) and \NN assign different codes to the same images. \NN changes the larger chimpanzee's eye
  and mouth, while \ours keeps their shape. PSNR is measured on the whole image; crop selection is described in
  \cref{sec:supp-every-tok}. \Cref{fig:teaser-gen} shows the effect on generation.}
\label{fig:teaser}

\end{center}
\vspace{0.6em}
}]

\begin{abstract}
Discrete visual tokens encode images for reconstruction and serve as training targets for autoregressive image generators. Vector-quantized tokenizers typically assign each encoder feature its nearest code independently. The decoder uses these codes jointly, so a sequence of individually nearest codes can reconstruct an image less faithfully than an alternative assignment. A local model of the reconstruction loss yields a closest-vector problem with a decoder-dependent target and metric. Reconstruction-only search improves PSNR across the frozen tokenizers we test. Different code sequences can achieve similar reconstruction quality, leaving room to favour more predictable ones.

We introduce RecaP (Reconstruction and Predictability), which selects codes for a frozen tokenizer by balancing how faithfully they reconstruct the image with how likely their sequence is under a pretrained autoregressive prior.
On ImageNet, LlamaGen GPT-B trained on RecaP tokens achieves lower early-checkpoint FID against training images than GPT-B trained on nearest-neighbour tokens. The difference is smaller against held-out images. On an ImageNet subset, the prior improves generation at comparable reconstruction FID.
Our code is available at: \url{https://github.com/gmum/recap}.
\end{abstract}

\section{Introduction}\label{sec:intro}

Discrete visual tokens provide a finite vocabulary for autoregressive image
generation~\cite{esser2021taming,sun2024llamagen,tian2024var}. A visual tokenizer
encodes each image as a sequence of indices into a codebook of learned vectors.
Its decoder uses the selected codes to reconstruct the image, and the
generator learns to predict the same indices. At generation time, the decoder
turns sampled sequences back into images. The choice of codes determines both what the decoder can recover
(\cref{fig:teaser}) and what the generator is trained to predict.

To choose the codes, a vector-quantized tokenizer first maps the image to a
sequence of continuous features. Nearest-neighbour (\NN) assignment selects
the closest codebook entry at each position with an independent
lookup~\cite{vandenoord2017neural,esser2021taming}. Recent work has improved
how codebooks are learned~\cite{zhu2024simvq,shi2024ibq} and shown how to train
larger visual tokenizers~\cite{xiong2025gigatok}.
We instead ask how much reconstruction quality we can recover by changing the
assignment rule while keeping the encoder, codebook and decoder fixed.

Proximity to an encoder feature gives only a partial picture of what will
reconstruct well. The decoder can respond differently to equally large
changes in different latent directions. The effect of changing one code can
also depend on the codes at other positions. A local quadratic model of the
reconstruction loss gives a closest-vector problem whose target can differ
from the encoder output and whose metric reflects the decoder's sensitivity
(\cref{sec:theory}). Reconstruction-only search improves PSNR across the
pretrained tokenizers we test at fixed codebook size and sequence length
(\cref{tab:recon}). The same pattern holds on COCO and DIV2K for the
tokenizers evaluated there (\cref{tab:coco}).

Reconstruction search can find substantially different token sequences at
similar PSNR (\cref{sec:objective}). This gives us room to change the
generator's training targets while preserving reconstruction quality. Prior
work encourages predictable tokens during tokenizer
training~\cite{ramanujan2025worse,wang2025larp}. We use predictability when
choosing codes with a frozen tokenizer.

\ours uses a pretrained autoregressive prior to favour sequences whose codes
are predictable in context. We combine the prior with a likelihood based on
the reconstruction loss and search over nearby codebook entries by maximum
a posteriori inference. Running the search with reconstruction loss alone
gives us a control for testing what the prior adds when the resulting
sequences are used to train a generator.

\begin{figure}[t]
\centering
\makebox[\linewidth][l]{\footnotesize\textbf{(a)} Preparing the training tokens}\par
\vspace{1pt}
\begin{tikzpicture}[
  x=1pt, y=1pt, font=\scriptsize,
  arr/.style={-{Latex[length=3.4pt,width=2.6pt]}, draw=black!60, line width=0.5pt},
  frozen/.style={fill=black!9, draw=black!40, line width=0.5pt, line join=round},
  choice/.style={rounded corners=2pt, line width=0.65pt, align=center, inner xsep=2pt, inner ysep=2.5pt},
  hdr/.style={inner sep=1pt, text=black!55},
  lbl/.style={inner sep=1pt, align=center},
  snow/.pic={\foreach \ang in {0,60,...,300}{
    \draw[black!55, line width=0.35pt] (0,0) -- (\ang:2.2)
      (\ang:1.3) -- +({\ang+45}:0.8) (\ang:1.3) -- +({\ang-45}:0.8);}}]
  \node[inner sep=0] (input) at (17,-18) {\includegraphics[width=34pt]{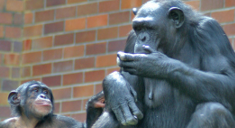}};
  \node[lbl, anchor=north] at (17,-28) {training image};
  \filldraw[frozen] (40,-30) -- (56,-25) -- (56,-11) -- (40,-6) -- cycle;
  \node at (48,-18) {$E$};
  \pic at (48,-1) {snow};
  \draw[arr] (input.east) -- (39.5,-18);
  \coordinate (split) at (63,-18);
  \draw[black!60, line width=0.5pt] (56.5,-18) -- (split);
  \fill[black!60] (split) circle[radius=0.85pt];
  \node[choice, draw=nncol, fill=nncol!10, minimum width=80pt] (nn) at (111,0)
    {\textbf{\NN{} assignment}\\nearest code};
  \node[choice, draw=mapcol, fill=mapcol!7, minimum width=80pt] (recap) at (111,-36)
    {\textbf{\oursname{} search}\\loss through frozen $D$\\$+$ prior: released GPT-B};
  \draw[arr, draw=nncol] (split) |- (nn.west);
  \draw[arr, draw=mapcol] (split) |- (recap.west);
  \foreach \rowy/\armcolor in {0/nncol,-36/mapcol} {
    \foreach \i in {0,...,3} \foreach \j in {0,...,3} {
      \filldraw[fill=black!13, draw=white, line width=0.45pt] (160+5*\i,\rowy-10+5*\j) rectangle ++(5,5);}
    \foreach \i/\j in {0/3,1/0,2/2,3/1,3/3} {
      \filldraw[fill=\armcolor, draw=white, line width=0.45pt] (160+5*\i,\rowy-10+5*\j) rectangle ++(5,5);}
  }
  \draw[arr, draw=nncol] (nn.east) -- (158,0);
  \draw[arr, draw=mapcol] (recap.east) -- (158,-36);
  \node[choice, draw=nncol, fill=nncol!10, minimum width=40pt] (gnn) at (211,0) {\textbf{GPT-B}\\same recipe};
  \node[choice, draw=mapcol, fill=mapcol!7, minimum width=40pt] (gmap) at (211,-36) {\textbf{GPT-B}\\same recipe};
  \draw[arr, draw=nncol] (182,0) -- (gnn.west);
  \draw[arr, draw=mapcol] (182,-36) -- (gmap.west);
  \node[hdr, anchor=south] at (111,14) {assignment rule};
  \node[hdr, anchor=south] at (170,14) {token maps};
  \node[hdr, anchor=south] at (211,14) {training};
\end{tikzpicture}\par
\vspace{4pt}
\makebox[\linewidth][l]{\footnotesize\textbf{(b)} GPT-B trained on each token set}\par
\vspace{2pt}
\includegraphics{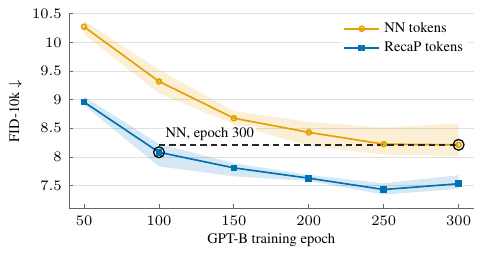}
\caption{\textbf{Code assignment changes the generator's training targets.}
  (a)~The same frozen tokenizer encodes each view using \NN or \ours. Coloured cells mark differences between the
  resulting token maps. Both GPT-B models use the same training recipe.
  (b)~On \INk at cfg~2.0, \ours at epoch~100 reaches the FID-10k of \NN at epoch~300 (dashed line); FID-10k against
  training images, mean and range over three seeds. At epoch~100, gFID-50k improves by 16\% on \INk ($p=0.003$) and
  9.5\% on \INh (\cref{tab:gen}).}
\label{fig:teaser-gen}
\end{figure}
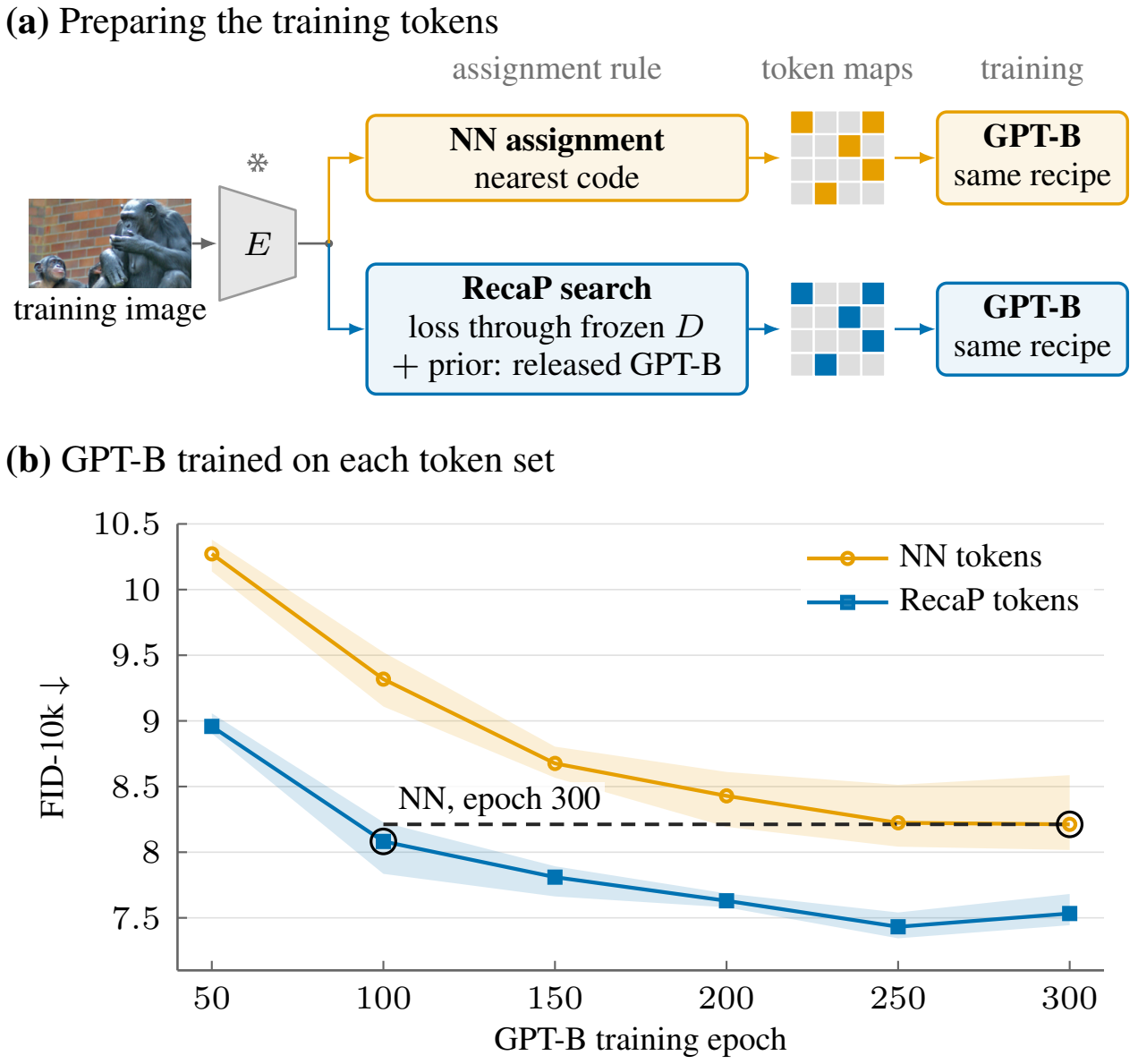

We train class-conditional LlamaGen GPT-B on ImageNet with the same training
recipe for both token sets. \ours lowers early-checkpoint FID against the
training images, including at matched guidance
(\cref{fig:teaser-gen,tab:robust}). The difference is smaller against
validation images (\cref{tab:gen}). On an ImageNet subset, adding the prior
lowers generation FID at similar reconstruction FID. Giving the prior more
weight makes the selected tokens easier to predict under that prior but
worsens generation FID (\cref{tab:mech}).

Preparing \ours tokens adds a one-time encoding cost. Reusing the tokens
across runs can reduce the cost per generator at comparable training-reference
FID (\cref{sec:cost}).

\noindent\textbf{Contributions}
\begin{itemize}\setlength\itemsep{1pt}
\item \textbf{The nearest-neighbour assignment gap.}
We formulate code assignment as a closest-vector problem under a local model
of the reconstruction loss and characterize when nearest-neighbour assignment
is exact in that model. Using reconstruction-only search, we measure how much
reconstruction quality can be recovered by changing code assignments in
frozen pretrained tokenizers.

\item \textbf{Token selection for autoregressive generation.}
We introduce \ours, which combines the reconstruction loss with an
autoregressive prior to select training targets for an image generator.
In controlled training experiments, we measure the effect of \ours on GPT-B
and isolate the contribution of the prior on an ImageNet subset.
\end{itemize}

\begin{figure*}[t]
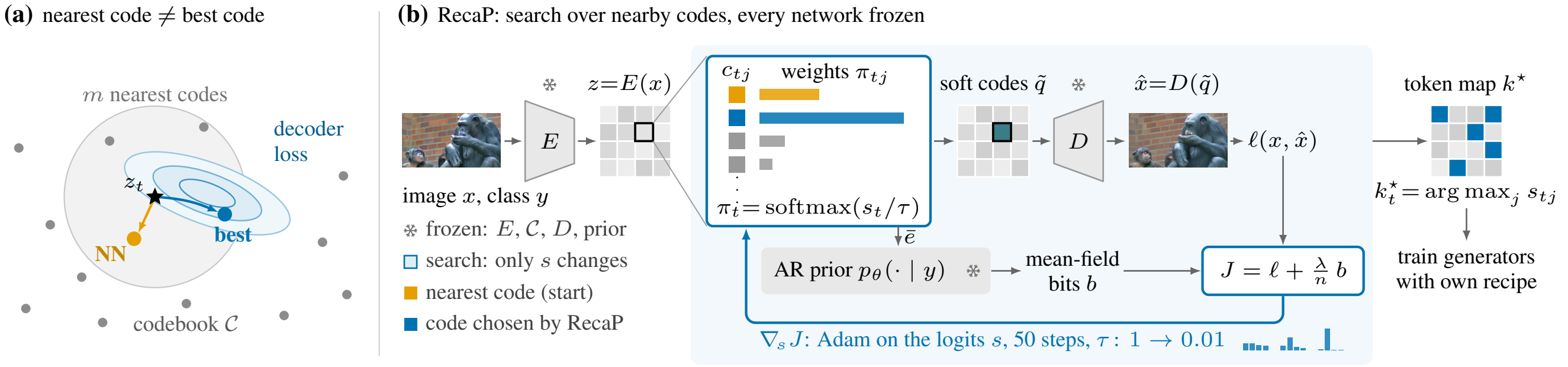

\centering
\begin{tikzpicture}[font=\scriptsize,
  arr/.style={-{Latex[length=1.2mm,width=0.9mm]}, line width=0.5pt, draw=black!60},
  frozen/.style={fill=black!9, draw=black!40, line width=0.5pt, line join=round},
  lbl/.style={inner sep=1pt},
  snow/.pic={\foreach \ang in {0,60,...,300}{
      \draw[black!55, line width=0.35pt] (0,0) -- (\ang:0.075)
        (\ang:0.045) -- +({\ang+45}:0.028) (\ang:0.045) -- +({\ang-45}:0.028);}}]

\node[anchor=base west, inner sep=0] at (0,3.78) {\textbf{\footnotesize(a)}\, nearest code $\neq$ best code};
\begin{scope}[x=0.78cm, y=0.78cm, xshift=-0.1cm, yshift=0.15cm]
  \filldraw[fill=black!5, draw=black!22, line width=0.4pt] (2.3,2.15) circle (1.3);
  \node[anchor=south, text=black!55, lbl] at (2.3,3.46) {$m$ nearest codes};
  \begin{scope}[shift={(3.05,2.2)}, rotate=-20]
    \fill[mapcol!6] (0,0) ellipse (1.25 and 0.44);
    \draw[mapcol!40, line width=0.45pt] (0,0) ellipse (1.25 and 0.44);
    \fill[mapcol!13] (0,0) ellipse (0.82 and 0.29);
    \draw[mapcol!55, line width=0.45pt] (0,0) ellipse (0.82 and 0.29);
    \draw[mapcol!75, line width=0.45pt] (0,0) ellipse (0.42 and 0.15);
  \end{scope}
  \node[mapcol!80!black, anchor=west, align=left, lbl] at (3.95,2.95) {decoder\\loss};
  \foreach \p in {(0.45,0.55),(1.55,0.35),(2.75,0.62),(4.15,0.45),(0.8,1.75),(4.7,1.5),(0.35,2.85),(1.65,2.95),
                  (3.0,3.15),(5.0,2.45),(5.05,0.75),(1.25,1.0)}
    \fill[black!45] \p circle (1.5pt);
  \node[black!65, anchor=north, lbl] at (2.75,0.5) {codebook $\mathcal C$};
  \fill[nncol] (2.0,1.55) circle (2.3pt);
  \fill[mapcol] (3.3,1.9) circle (2.3pt);
  \draw[-{Latex[length=1.2mm,width=0.9mm]}, nncol, line width=0.8pt] (2.3,2.15) -- (2.06,1.63);
  \draw[-{Latex[length=1.2mm,width=0.9mm]}, mapcol, line width=0.8pt] (2.3,2.15) to[bend left=10] (3.21,1.93);
  \node[star, star points=5, star point ratio=2.2, fill=black, inner sep=0pt, minimum size=6pt] at (2.3,2.15) {};
  \node[anchor=south east, lbl] at (2.22,2.18) {$z_t$};
  \node[nncol!85!black, anchor=north east, font=\scriptsize\bfseries, lbl] at (1.95,1.5) {NN};
  \node[mapcol, anchor=north, font=\scriptsize\bfseries, lbl] at (3.42,1.78) {best};
\end{scope}
\draw[black!18, line width=0.4pt] (4.2,0.05) -- (4.2,3.9);

\node[anchor=base west, inner sep=0] at (4.4,3.78)
  {\textbf{\footnotesize(b)}\, \oursname: search over nearby codes, every network frozen};
\fill[mapcol!5, rounded corners=3pt] (7.68,-0.05) rectangle (15.3,3.52);

\node[inner sep=0, anchor=south west] (xin) at (4.45,2.15)
  {\includegraphics[width=1.1cm, height=0.6cm]{chimp_orig.png}};
\node[lbl, anchor=west] at (4.42,1.82) {image $x$, class $y$};
\draw[arr] (5.6,2.45) -- (5.8,2.45);
\filldraw[frozen] (5.83,2.0) -- (6.38,2.2) -- (6.38,2.7) -- (5.83,2.9) -- cycle;
\node at (6.105,2.45) {$E$};
\pic at (6.105,3.08) {snow};
\draw[arr] (6.42,2.45) -- (6.63,2.45);
\foreach \i in {0,...,3} \foreach \j in {0,...,3} {
  \pgfmathtruncatemacro{\g}{7+4*mod(7*\i+5*\j+1,4)}
  \filldraw[fill=black!\g, draw=white, line width=0.5pt] (6.66+0.2*\i,2.05+0.2*\j) rectangle ++(0.2,0.2);}
\draw[black, line width=0.7pt] (7.06,2.45) rectangle ++(0.2,0.2);
\node[lbl] at (7.0,3.08) {$z{=}E(x)$};

\draw[black!40, line width=0.35pt] (7.26,2.65) -- (7.86,3.4) (7.26,2.45) -- (7.86,1.5);
\filldraw[fill=white, draw=mapcol, line width=0.8pt, rounded corners=2pt] (7.86,1.5) rectangle (10.36,3.4);
\node[lbl] at (8.2,3.2) {$c_{tj}$};
\node[lbl] at (9.3,3.2) {weights $\pi_{tj}$};
\foreach \y/\col/\pp in {2.97/nncol/0.38, 2.71/mapcol/0.92, 2.45/black!40/0.16, 2.19/black!40/0.08} {
  \fill[\col] (8.11,\y-0.09) rectangle (8.29,\y+0.09);
  \fill[\col, opacity=0.85] (8.45,\y-0.06) rectangle (8.45+1.75*\pp,\y+0.06);}
\node[lbl, font=\tiny] at (8.2,2.01) {$\vdots$};
\node[lbl] at (9.11,1.7) {$\pi_t{=}\softmax(s_t/\tau)$};

\draw[arr] (10.4,2.45) -- (10.63,2.45);
\foreach \i in {0,...,3} \foreach \j in {0,...,3} {
  \pgfmathtruncatemacro{\g}{7+4*mod(3*\i+2*\j+2,4)}
  \filldraw[fill=black!\g, draw=white, line width=0.5pt] (10.66+0.2*\i,2.05+0.2*\j) rectangle ++(0.2,0.2);}
\filldraw[fill=mapcol!75!nncol, draw=black, line width=0.7pt] (11.06,2.45) rectangle ++(0.2,0.2);
\node[lbl] at (11.06,3.08) {soft codes $\tilde q$};
\draw[arr] (11.5,2.45) -- (11.71,2.45);
\filldraw[frozen] (11.74,2.2) -- (12.29,2.0) -- (12.29,2.9) -- (11.74,2.7) -- cycle;
\node at (12.015,2.45) {$D$};
\pic at (12.015,3.08) {snow};
\draw[arr] (12.33,2.45) -- (12.54,2.45);
\node[inner sep=0, anchor=south west] (xhat) at (12.57,2.15)
  {\includegraphics[width=1.1cm, height=0.6cm]{chimp_recap.png}};
\node[lbl] at (13.12,3.08) {$\hat x{=}D(\tilde q)$};
\node[lbl] (L) at (14.3,2.45) {$\ell(x,\hat x)$};
\draw[arr] (13.71,2.45) -- (L.west);

\node[frozen, draw=none, rounded corners=2pt, minimum height=0.52cm, inner xsep=4pt] (P) at (9.75,1.0)
  {AR prior $p_\theta(\cdot\mid y)$\kern10pt};
\pic at ($(P.east)+(-0.2,0)$) {snow};
\coordinate (ebar) at (10.0,1.5);
\draw[arr] (ebar) -- (ebar |- P.north);
\node[anchor=west, lbl] at (10.02,1.38) {$\bar e$};
\node[align=center, lbl] (B) at (11.95,1.0) {mean-field\\bits $b$};
\node[draw=mapcol, line width=0.8pt, fill=white, rounded corners=2pt, minimum width=1.8cm, minimum height=0.52cm]
  (J) at (14.3,1.0) {$J=\ell+\tfrac{\lambda}{n}\,b$};
\draw[arr] (P) -- (B);
\draw[arr] (B) -- (J);
\draw[arr] (L) -- (J);

\draw[-{Latex[length=1.3mm,width=1mm]}, mapcol, line width=0.8pt, rounded corners=3pt]
  (J.south) -- (14.3,0.42) -- (8.3,0.42) -- (8.3,1.48);
\node[anchor=north west, text=mapcol, lbl] (gl) at (8.4,0.37)
  {$\nabla_{\!s}J$: Adam on the logits $s$, 50 steps, $\tau\colon 1\to0.01$};
\begin{scope}[shift={($(gl.south east)+(0.15,0.05)$)}]
  \foreach \i/\h in {0/0.30,1/0.28,2/0.22,3/0.20} \fill[mapcol!80] (\i*0.075,0) rectangle ++(0.055,\h*0.26);
  \foreach \i/\h in {0/0.20,1/0.55,2/0.15,3/0.10} \fill[mapcol!80] (0.42+\i*0.075,0) rectangle ++(0.055,\h*0.26);
  \foreach \i/\h in {0/0.04,1/0.92,2/0.02,3/0.02} \fill[mapcol!80] (0.84+\i*0.075,0) rectangle ++(0.055,\h*0.26);
\end{scope}

\pic at (4.55,1.45) {snow};
\node[anchor=west, text=black!75, lbl] at (4.68,1.45) {frozen: $E$, $\mathcal C$, $D$, prior};
\filldraw[fill=mapcol!12, draw=mapcol, line width=0.5pt] (4.48,1.03) rectangle (4.62,1.17);
\node[anchor=west, text=black!75, lbl] at (4.68,1.1) {search: only $s$ changes};
\fill[nncol] (4.48,0.68) rectangle (4.62,0.82);
\node[anchor=west, text=black!75, lbl] at (4.68,0.75) {nearest code (start)};
\fill[mapcol] (4.48,0.33) rectangle (4.62,0.47);
\node[anchor=west, text=black!75, lbl] at (4.68,0.4) {code chosen by \oursname};

\draw[arr] (15.3,2.45) -- (15.92,2.45);
\foreach \i in {0,...,3} \foreach \j in {0,...,3} {
  \pgfmathtruncatemacro{\g}{7+4*mod(5*\i+3*\j,4)}
  \filldraw[fill=black!\g, draw=white, line width=0.5pt] (15.95+0.2*\i,2.05+0.2*\j) rectangle ++(0.2,0.2);}
\foreach \i/\j in {2/2,0/3,3/1,1/0,3/3}
  \filldraw[fill=mapcol, draw=white, line width=0.5pt] (15.95+0.2*\i,2.05+0.2*\j) rectangle ++(0.2,0.2);
\node[lbl] at (16.35,3.08) {token map $k^\star$};
\node[lbl] at (16.35,1.86) {$k^\star_t{=}\argmax_j s_{tj}$};
\draw[arr] (16.35,1.7) -- (16.35,1.33);
\node[align=center, lbl] at (16.35,1.0) {train generators\\with own recipe};
\end{tikzpicture}
\caption{\textbf{Decoder-aware code assignment and the \ours search.}
  (a)~In the local model, the reconstruction loss has a shifted target and an anisotropic metric, so a code
  farther from the encoder feature $z_t$ can reconstruct better than the nearest one.
  (b)~Encoder, codebook, decoder and prior stay frozen. For each position $t$ (zoomed in), the decoder reads a
  mixture of the $m=16$ nearest codes. Gradient steps on $J$ (blue track) update only the mixture logits $s$, as
  $\tau$ decreases and the weights sharpen. The final argmax gives one discrete token map per view; blue cells mark
  codes that differ from \NN.}
\label{fig:method}
\end{figure*}

\section{Related work}\label{sec:related}

Visual tokenizers such as VQ-VAE and VQGAN assign each encoder feature to a code
independently~\cite{vandenoord2017neural,esser2021taming}. Later work enlarges the
codebook~\cite{zhu2024simvq,shi2024ibq,zhu2024vqganlc}, uses implicit
grids~\cite{mentzer2024fsq,yu2024lfq,luo2024openmagvit2} or stacks residual
codes~\cite{lee2022rqvae,jia2025mgvq}. Other work changes the token
layout~\cite{yu2024titok}, scales the tokenizer~\cite{xiong2025gigatok,chang2025fvq}
or learns a new quantizer~\cite{fang2026vqtransplant}. CRT, LARP, AliTok and GEAR
shape tokens for generation during training~\cite{ramanujan2025worse,wang2025larp,wu2026alitok,lin2026gear}.
\ours uses predictability to select codes at encoding time with a frozen tokenizer.

\noindent\textbf{Encoding-time optimization.}
Iterative inference reduces the amortization gap in
VAEs~\cite{cremer2018inference,kim2018semi,marino2018iterative}, and neural image
codecs refine latents at encoding time~\cite{campos2019content,yang2020improving,rdvq2026}.
\citet{laobeyer2025highly} optimize codes of a frozen 1D tokenizer for editing and
training-free generation. Soft quantization and relaxed posteriors provide related
optimization tools~\cite{agustsson2017soft,jang2017gumbel,murata2024g2d2,rout2025aps}.
We measure the reconstruction gap left by the released assignment rule and train
generators on the reassigned tokens.

\noindent\textbf{Rate--distortion assignment.}
Our objective takes the form of entropy-constrained VQ~\cite{chou1989entropy}, with a
learned decoder and an autoregressive prior. Related assignment criteria appear in
video coding~\cite{ortega1998rate}, speech coding~\cite{schroeder1985celp} and text
segmentation~\cite{kudo2018subword,meister2026langmap}. The local assignment problem
also connects to lattice decoding and second-order weight quantization, including
Babai's nearest-plane rule and its use in
GPTQ~\cite{babai1986lovasz,chen2026gptqbabai,frantar2023gptq,nagel2020adaround,chee2023quip,chen2026mpq}.

\noindent\textbf{Training targets for generators.}
VA-$\pi$ combines a decoder likelihood and an autoregressive prior to tune a
generator on fixed tokens~\cite{liao2025vapi}, and feature-space Fr\'echet loss
supervises generator post-training~\cite{zhang2026fdloss}. \ours applies its objective
when choosing the training tokens. Related ways to change training targets include
reassigning speech units~\cite{kamper2021dpdp,kamper2023word,visser2025dpdp},
sampling multiple token maps or image crops~\cite{lee2022rqvae,sun2024llamagen},
and distilling a teacher's next-token distribution at each training
step~\cite{hinton2015distilling,kim2016sequence,peruzzo2026varkd}.
We compare additional crops, multiple maps and distillation in \cref{sec:ablation}.

\noindent\textbf{FID evaluation.}
FID has finite-sample bias and depends on the evaluation
protocol~\cite{heusel2017gans,chong2020effectively,parmar2022aliased,kynkaanniemi2023role,zheng2025rae,stein2023exposing}.
We derive the bias when each reconstruction is paired with its original.
\section{\texorpdfstring{\ours}{\oursname}: tokenization as MAP inference}\label{sec:method}

We first analyse when independent nearest-code lookup solves the reconstruction
problem. We then use reconstruction and an autoregressive prior to choose the
token maps on which a generator trains.

\subsection{Setup}\label{sec:setup}
For image $x$, encoder $E$ returns one feature per position, $z=E(x)=(z_1,\dots,z_n)$. A token map
$k\in[K]^n$ selects vectors $q(k)=(c_{k_1},\dots,c_{k_n})$ from $\mathcal C=\{c_1,\dots,c_K\}\subset\R^d$,
and the decoder reconstructs $D(q(k))$. The released \NN rule sets
$k_t^{\textsc{nn}}=\argmin_j\|z_t-c_j\|$ independently at each position (IBQ selects the largest inner product).
Let $\ell(k)=\ell\big(x,D(q(k))\big)$ be the reconstruction loss without the adversarial term.
The \emph{assignment gap} is $\ell(k^{\textsc{nn}})-\min_k\ell(k)$. We quantify the recoverable improvement by comparing the PSNR of
the best map found by search with that of the \NN map.

\def\fzXa{0}\def\fzYa{112}\def\fzXb{64}\def\fzYb{64}\def\fzXc{96}\def\fzYc{176}\def\fzXd{160}\def\fzYd{128}%
\def\fzDesc{A COCO image}%

\providecommand{\oursLab}[1]{\textcolor{mapcol}{\textbf{#1}}}%
\providecommand{\nnLab}[1]{\textcolor{nncol!85!black}{\textbf{#1}}}%
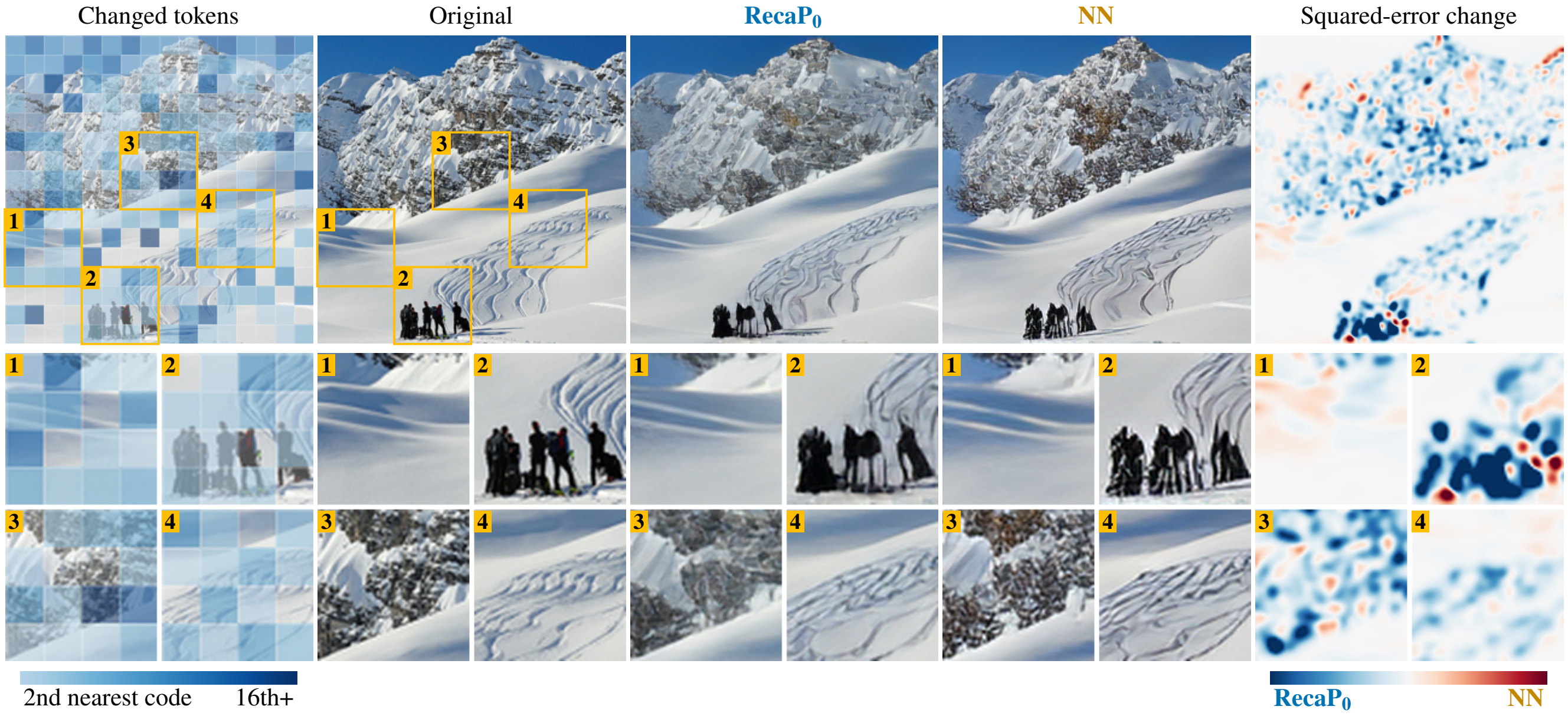
\begin{figure*}[t]
\centering
\definecolor{zoomcol}{RGB}{255,190,0}
\pgfmathsetlengthmacro{\mapunit}{\linewidth/1296}
\begin{tikzpicture}[x=\mapunit, y=\mapunit,
  num/.style={anchor=north west, font=\scriptsize\bfseries, text=black, fill=zoomcol, inner sep=1pt}]
  \useasboundingbox (0,-48) rectangle (1296,556);
  \node[anchor=south west, inner sep=0] at (0,264)
    {\includegraphics[width=\dimexpr\mapunit*256\relax]{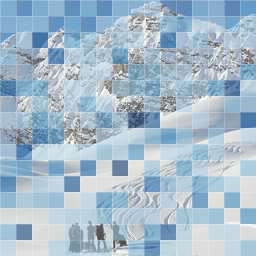}};
  \node[anchor=south west, inner sep=0] at (0,0)
    {\includegraphics[width=\dimexpr\mapunit*256\relax]{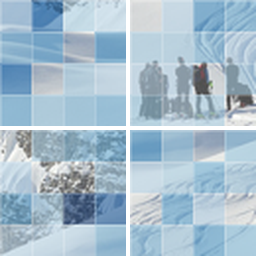}};
  \node[anchor=south west, inner sep=0] at (260,264)
    {\includegraphics[width=\dimexpr\mapunit*256\relax]{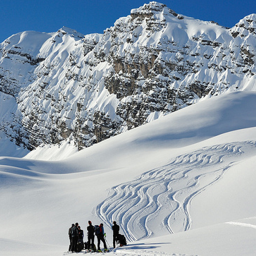}};
  \node[anchor=south west, inner sep=0] at (260,0)
    {\includegraphics[width=\dimexpr\mapunit*256\relax]{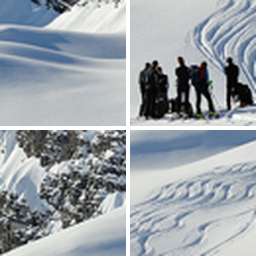}};
  \node[anchor=south west, inner sep=0] at (520,264)
    {\includegraphics[width=\dimexpr\mapunit*256\relax]{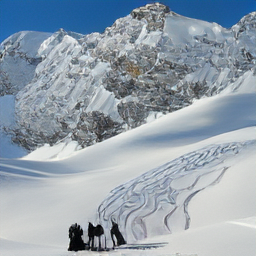}};
  \node[anchor=south west, inner sep=0] at (520,0)
    {\includegraphics[width=\dimexpr\mapunit*256\relax]{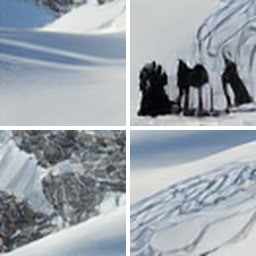}};
  \node[anchor=south west, inner sep=0] at (780,264)
    {\includegraphics[width=\dimexpr\mapunit*256\relax]{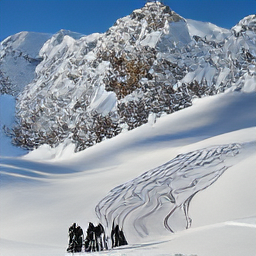}};
  \node[anchor=south west, inner sep=0] at (780,0)
    {\includegraphics[width=\dimexpr\mapunit*256\relax]{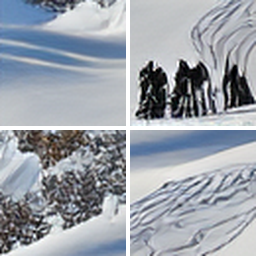}};
  \node[anchor=south west, inner sep=0] at (1040,264)
    {\includegraphics[width=\dimexpr\mapunit*256\relax]{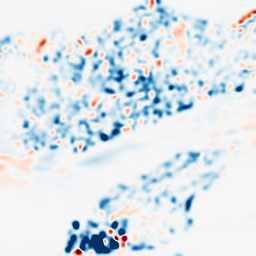}};
  \node[anchor=south west, inner sep=0] at (1040,0)
    {\includegraphics[width=\dimexpr\mapunit*256\relax]{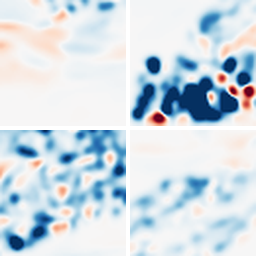}};
  \foreach \col/\heading in {0/{Changed tokens}, 1/{Original}, 2/{\oursLab{\RXplot}}, 3/{\nnLab{\NN}},
                           4/{Squared-error change}}
    \node[anchor=south, font=\footnotesize, inner sep=2pt, text height=1.5ex, text depth=0.4ex] at ({128+260*\col},520) {\heading};
  \foreach \n/\wx/\wy in {1/\fzXa/\fzYa, 2/\fzXb/\fzYb, 3/\fzXc/\fzYc, 4/\fzXd/\fzYd} {
    \foreach \col in {0,1} {
      \draw[zoomcol, line width=0.7pt] ({260*\col+\wx},{264+\wy-64}) rectangle ({260*\col+\wx+64},{264+\wy});
      \node[num] at ({260*\col+\wx},{264+\wy}) {\n};
    }
  }
  \foreach \n/\cx/\cy in {1/0/256, 2/130/256, 3/0/126, 4/130/126}
    \foreach \col in {0,...,4}
      \node[num] at ({260*\col+\cx},\cy) {\n};
  \node[anchor=north, inner sep=0] (rank) at (128,-8)
    {\includegraphics[width=\dimexpr\mapunit*230\relax, height=4pt]{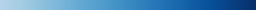}};
  \node[anchor=north west, font=\footnotesize, inner sep=1pt] at (rank.south west) {2nd nearest code};
  \node[anchor=north east, font=\footnotesize, inner sep=1pt] at (rank.south east) {16th+};
  \node[anchor=north, inner sep=0] (error) at (1168,-8)
    {\reflectbox{\includegraphics[width=\dimexpr\mapunit*230\relax, height=4pt]{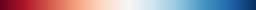}}};
  \node[anchor=north west, font=\footnotesize, inner sep=1pt] at (error.south west) {\oursLab{\RXplot}};
  \node[anchor=north east, font=\footnotesize, inner sep=1pt] at (error.south east) {\nnLab{\NN}};
\end{tikzpicture}
\caption{\textbf{Where changing tokens improves reconstruction.}
  \fzDesc{} (LlamaGen VQ-16): some patches whose token \RX leaves unchanged (clear cells) also show lower squared
  error. The left column shades each changed token by the rank of its new code by distance to the encoder feature;
  the darkest shade covers ranks 16 and above. The right column shows the smoothed change in squared error, blue
  favouring \RX and red favouring \NN. Bottom row: the four 64\,px windows with the most changed tokens, magnified
  2$\times$.}
\label{fig:token-map-main}
\end{figure*}

\subsection{Why nearest neighbour leaves a gap}\label{sec:theory}
The decoder can respond differently across latent directions and couple codes at different positions.
We approximate $\ell(q)=\ell(x,D(q))$ around $q=z$, with gradient $g$ and Gauss--Newton matrix
$\Hm\succ0$ (damped if needed). Writing $\|v\|_{\Hm}^2=v\T\Hm v$, we obtain
\begin{equation}
\ell(q)\approx\ell(z)-\tfrac12 g\T\Hm^{-1}g+\tfrac12\|q-t\|_{\Hm}^2,\quad t=z-\Hm^{-1}g.
\label{eq:quad}
\end{equation}
Minimizing \cref{eq:quad} over $q\in\mathcal C^n$ gives a closest-vector problem with target $t$ and metric $\Hm$.
\NN uses $z$ as the target and measures distance with $I$. For distinct encoder features, it is exact in the local
model for every codebook only if $g=0$ and $\Hm$ has isotropic diagonal blocks and zero off-diagonal blocks
(\cref{lem:supp-nn-exact}). This keeps the target at $z$ and makes the loss separable and isotropic within each position.

To isolate the role of the metric, we fix the target at $t$ and use a
high-resolution codebook model. In this setting, the expected error ratio
between independent Euclidean lookup and sequential nearest-plane
assignment~\cite{babai1986lovasz,agrell2002closest} splits into two factors.
One captures directional sensitivity within a position and the other captures
coupling between positions. Under this model, denser codes reduce absolute
error, while the relative penalty for using the wrong metric stays the same
(\cref{prop:amgm,cor:fischer}). \Cref{sec:supp-proofs} also checks these effects
on trained tokenizers. The search below uses the reconstruction loss of the
frozen decoder directly.

\subsection{The \texorpdfstring{\ours}{\oursname} objective}\label{sec:objective}
Different code maps can reconstruct an image nearly equally well. Two runs of reconstruction-only search from
perturbed starts agree on only 55\% of token positions at equal PSNR. This gives the prior room to favour more
predictable sequences while retaining reconstruction quality.
For a class label $y$, a pretrained autoregressive generator defines the prior $p_\theta(k\mid y)$. We encode $x$ by minimizing
\begin{equation}
k^\star=\argmin_{k\in\mathcal K(x)}\;\ell\big(x,D(q(k))\big)+\frac{\lambda}{n}\big(-\log_2 p_\theta(k\mid y)\big).
\label{eq:map}
\end{equation}
The candidate set $\mathcal K(x)$ combines nearby codes at each position, ranked
as in the tokenizer's lookup. The weight $\lambda$ sets the trade-off between reconstruction
loss and bits per token under the prior. For $\lambda>0$, the objective gives MAP
inference over $\mathcal K(x)$ under a reconstruction-based likelihood with a
code-independent normalizer (\cref{sec:supp-transport}). At $\lambda=0$, \RX searches
using reconstruction loss alone and can refine the result outside $\mathcal K(x)$.
Unless stated otherwise, \ours uses $\lambda=0.01$ and the released generator as prior.

\subsection{Annealed relaxation}\label{sec:relax}
We approximate \cref{eq:map} by letting the decoder read a mixture of candidate
codes at each position (\cref{fig:method}b). The weights
$\pi_t=\softmax(s_t/\tau)$ define the soft code
$\tilde q_t=\sum_j\pi_{tj}c_{tj}$. We initialize $s_t$ so that its argmax gives the
\NN code. Adam~\cite{kingma2015adam} updates the logits as the temperature $\tau$
decreases and the mixtures sharpen. The final argmax gives a discrete code at each
position.

The prior reads the expected embeddings of preceding tokens and scores each
candidate at the current position. The mean-field rate averages their negative
log-probabilities under $\pi_t$. It is differentiable and reduces to the discrete rate when each
position selects a single code. \Cref{alg:relax,sec:supp-relax,sec:supp-impl} give
the algorithm, its derivation and the settings. \Cref{sec:negative} compares
simpler ways to search for reconstruction codes.

\begin{table*}[t]
\caption{\textbf{The assignment gap on twelve released tokenizers}; only the assignment changes.
  \RX searches using reconstruction loss alone (\cref{sec:supp-impl}). rFID uses ImageNet-1k
  validation images (50k, or the first 10k for TiTok) and each tokenizer's evaluator in \cref{tab:impl}.
  FID$_\infty$ estimates clean-fid rFID at infinite sample size using the closed form
  in \cref{sec:supp-bias} ($^\S$ uses the $1/N$ fit). Its values are directly comparable
  to rFID only for clean-fid rows. Changes are relative to \NN. Bold and underlining
  mark the largest and second-largest improvements in each change column.}
\label{tab:recon}
\centering\small
\setlength{\tabcolsep}{4.2pt}
\begin{tabular*}{\textwidth}{@{\extracolsep{\fill}}llrrrrrrr@{}}
\toprule
 & & \multicolumn{3}{c}{rFID-50k $\downarrow$} & \multicolumn{2}{c}{FID$_\infty$ $\downarrow$} & & \\
\cmidrule(lr){3-5}\cmidrule(lr){6-7}
Tokenizer & $K$, tokens & \NN & \RX & change & \NN & \RX & $\Delta$PSNR (dB) $\uparrow$ & $\Delta$\lpipsA $\downarrow$ \\
\midrule
LlamaGen VQ-16~\cite{sun2024llamagen} & $2^{14}$, 16$\times$16 & 2.25 & 1.30 & \best{$-$42\%} & 1.73 & 0.86 & +1.27 & $-$5.4\% \\
SimVQ~\cite{zhu2024simvq} (128\,px) & $2^{10}$, 16$\times$16 & 3.67 & 2.60 & $-$29\% & 3.54 & 2.51 & +1.22 & \second{$-$9.2\%} \\
SimVQ (128\,px) & $2^{18}$, 16$\times$16 & 2.00 & 1.43 & $-$28\% & 1.91 & 1.34 & +1.04 & \best{$-$13.1\%} \\
Open-MAGVIT2-L~\cite{luo2024openmagvit2} & $2^{18}$ LFQ, 16$\times$16 & 1.25 & 1.01 & $-$19\% & 0.89$^\S$ & 0.64$^\S$ & +1.06 & $-$1.1\% \\
IBQ~\cite{shi2024ibq} & $2^{14}$, 16$\times$16 & 1.38 & 1.08 & $-$21\% & 0.97$^\S$ & 0.69$^\S$ & +1.17 & $-$0.2\% \\
IBQ & $2^{18}$, 16$\times$16 & 1.01 & 0.85 & $-$16\% & 0.64$^\S$ & 0.48$^\S$ & +1.09 & 0.0\% \\
FVQ~\cite{chang2025fvq} & $2^{18}$, 16$\times$16 & 1.15 & 0.80 & \second{$-$30\%} & 0.73 & 0.43 & +1.14 & $-$3.5\% \\
TiTok-S-128~\cite{yu2024titok} (10k) & $2^{12}$, 128 (1D) & 4.27 & 3.76 & $-$12\% & 1.66 & 1.46 & \best{+1.86} & $-$8.7\% \\
TiTok-SL-256 (10k) & $2^{13}$, 256 (1D) & 2.87 & 2.36 & $-$18\% & 0.85 & 0.60 & +0.90 & $-$1.7\% \\
GigaTok B-L-2D~\cite{xiong2025gigatok} & $2^{14}$, 16$\times$16 & 0.87 & 0.72 & $-$17\% & 0.48 & 0.36 & +1.25 & $-$3.5\% \\
GigaTok B-L & $2^{14}$, 256 (1D) & 0.81 & 0.72 & $-$11\% & 0.44$^\S$ & 0.38$^\S$ & \second{+1.29} & $-$5.8\% \\
GigaTok B-L + DINO disc. & $2^{14}$, 256 (1D) & 0.54 & 0.60 & +10\% & 0.17$^\S$ & 0.25$^\S$ & +1.28 & $-$5.0\% \\
\bottomrule
\end{tabular*}
\end{table*}

\subsection{Tokens for generation}\label{sec:gen-method}
The prior scores sequences during encoding and can change the distribution
of the resulting training targets. We encode each training view once and can
reuse those token maps across runs. Each downstream generator trains on the
selected maps and samples using its usual recipe. Within the candidate sets,
exact MAP minimizes expected reconstruction loss among encoders with the same
joint distribution of classes and code maps (\cref{prop:transport}).
\section{Experiments}\label{sec:exp}

We compare search with \NN to measure how much reconstruction quality reassignment
recovers. We then train generators on the selected tokens and use \RX to test
what adding the prior contributes.

\subsection{Setup}\label{sec:exp-setup}
For reconstruction, we compare \NN and \RX on the same ImageNet validation images
across twelve frozen tokenizers from seven families. For generation, we train the
same class-conditional LlamaGen GPT-B on different tokens of the same images and
views, with the same training recipe and budget. We test a 100-class subset
(\INh) and full ImageNet (\INk). The default views are the centre crop and its
horizontal flip. \Cref{sec:supp-impl} gives implementation details.

Reconstruction FID (rFID) compares reconstructions with their originals. We also
report PSNR and both raw and bias-corrected rFID, since pairing each image
with its reconstruction affects finite-sample bias (\cref{sec:supp-bias}).
Generation FID (gFID) uses training images as the reference, and held-out gFID
uses validation images. For the main gFID-50k comparison, we choose each run's
classifier-free guidance (cfg) scale on 10k samples.
The assignment objective measures rate under the frozen prior. To measure
how predictable the selected targets are for the trained generator, we report
its validation negative log-likelihood (NLL) in bits per token on its own
assignments.

We mark analyses outside the pre-registered comparisons as post hoc. Generation
results use three training seeds per variant unless stated otherwise. Third
seeds were added after observing the two-seed results (\cref{sec:supp-gen}).
The main text reports unadjusted, two-sided Welch tests. The appendix also
reports Holm-adjusted tests for several pre-registered comparison families.
An anonymized implementation of \ours is included in the supplementary material.

\subsection{Reconstruction}\label{sec:recon}
\def\fvXA{77}\def\fvXeA{141}\def\fvYbA{141}\def\fvYtA{205}%
\def\fvNNA{134}\def\fvRXA{42}\def\fvHFA{104}%
\def\fvDescA{a DIV2K image}\def\fvSetA{DIV2K}\def\fvWhatA{the parrot's feathers}%
\def\fvXB{86}\def\fvXeB{150}\def\fvYbB{138}\def\fvYtB{202}%
\def\fvNNB{110}\def\fvRXB{47}\def\fvHFB{104}%
\def\fvDescB{an ImageNet validation image}\def\fvSetB{ImageNet validation}\def\fvWhatB{the hummingbird's throat feathers}%

\providecommand{\oursLab}[1]{\textcolor{mapcol}{\textbf{#1}}}%
\providecommand{\nnLab}[1]{\textcolor{nncol!85!black}{\textbf{#1}}}%
\begin{figure*}[t]
\centering
\definecolor{zoomcol}{RGB}{255,190,0}
\pgfmathsetlengthmacro{\fivunit}{\linewidth/1312}
\begin{tikzpicture}[x=\fivunit, y=\fivunit, energy/.style={font=\footnotesize, inner sep=0pt}]
  \foreach \col/\heading in {0/{Image}, 1/{Original, 4$\times$}, 2/{\oursLab{\RXplot}}, 3/{\oursLab{\RXplot{} + energy}},
                           4/{\nnLab{\NN}}}
    \node[anchor=south, font=\footnotesize, inner sep=2pt, text height=1.5ex, text depth=0.4ex] at ({264*\col+128},552) {\heading};
  \node[anchor=south west, inner sep=0] at (0,296)
    {\includegraphics[width=\dimexpr\fivunit*256\relax]{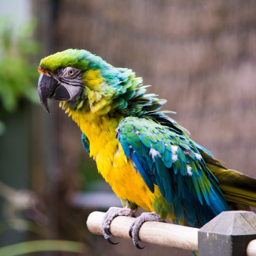}};
  \node[anchor=south west, inner sep=0] at (264,296)
    {\includegraphics[width=\dimexpr\fivunit*256\relax]{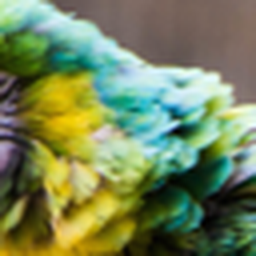}};
  \node[anchor=south west, inner sep=0] at (528,296)
    {\includegraphics[width=\dimexpr\fivunit*256\relax]{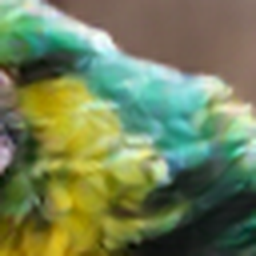}};
  \node[anchor=south west, inner sep=0] at (792,296)
    {\includegraphics[width=\dimexpr\fivunit*256\relax]{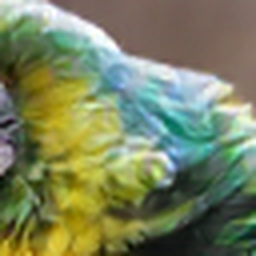}};
  \node[anchor=south west, inner sep=0] at (1056,296)
    {\includegraphics[width=\dimexpr\fivunit*256\relax]{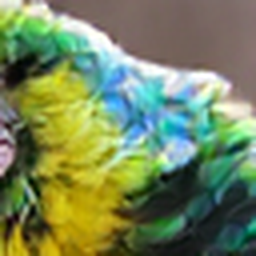}};
  \node[anchor=south west, inner sep=0] at (0,0)
    {\includegraphics[width=\dimexpr\fivunit*256\relax]{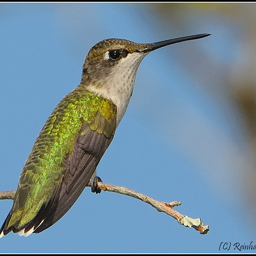}};
  \node[anchor=south west, inner sep=0] at (264,0)
    {\includegraphics[width=\dimexpr\fivunit*256\relax]{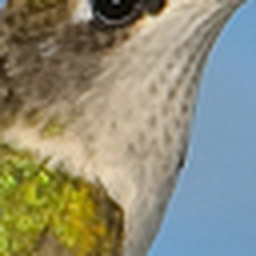}};
  \node[anchor=south west, inner sep=0] at (528,0)
    {\includegraphics[width=\dimexpr\fivunit*256\relax]{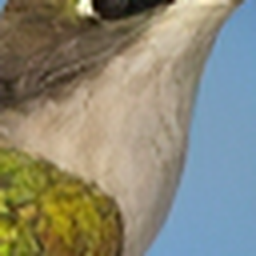}};
  \node[anchor=south west, inner sep=0] at (792,0)
    {\includegraphics[width=\dimexpr\fivunit*256\relax]{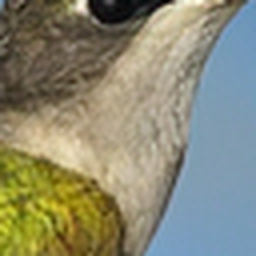}};
  \node[anchor=south west, inner sep=0] at (1056,0)
    {\includegraphics[width=\dimexpr\fivunit*256\relax]{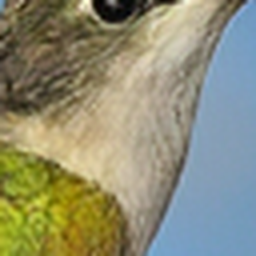}};
  \foreach \row/\y/\x/\xe/\yb/\yt/\enn/\erx/\ehf in {A/296/\fvXA/\fvXeA/\fvYbA/\fvYtA/\fvNNA/\fvRXA/\fvHFA,
                                                  B/0/\fvXB/\fvXeB/\fvYbB/\fvYtB/\fvNNB/\fvRXB/\fvHFB} {
    \begin{scope}[shift={(0,\y)}]
      \draw[zoomcol, line width=0.9pt] (\x,\yb) rectangle (\xe,\yt);
      \draw[zoomcol, line width=0.9pt] (264,0) rectangle (520,256);
      \draw[zoomcol, line width=0.4pt, opacity=0.85] (\xe,\yt) -- (264,256) (\xe,\yb) -- (264,0);
      \node[energy, anchor=north east, align=right] (elab) at (252,-8) {Finest-scale energy:};
      \foreach \xx/\v in {392/100, 656/\erx, 920/\ehf, 1184/\enn}
        \node[energy, anchor=center] at (\xx,0 |- elab.center) {\v\%};
    \end{scope}
  }
\end{tikzpicture}
\caption{\textbf{Fine texture after reassignment.}
  \RX smooths \fvWhatA{} (top) and \fvWhatB{} (bottom); adding the local energy term ($\gamma=0.02$) brings their
  finest-scale energy closer to the original's. Percentages are relative to the original window. 64\,px windows
  (yellow boxes) at 4$\times$ zoom of \ifx\fvSetA\fvSetB two \fvSetA{} images\else\fvDescA{} (top) and \fvDescB{}
  (bottom)\fi{} (LlamaGen VQ-16), selected for the largest energy recovery among textured windows.}
\label{fig:texture-main}
\end{figure*}

Reconstruction-only search improves PSNR by 0.9--1.9\,dB across all twelve
tokenizers at fixed codebook size and sequence length (\cref{tab:recon}).
At least 99.7\% of images improve. rFID decreases by 11--42\% for eleven
tokenizers, and the bias-corrected estimate decreases for the same eleven.
\lpipsA, which uses a different feature network from the LPIPS-VGG term used
in search, decreases or stays constant. The pre-registered COCO and DIV2K
comparisons also show PSNR gains for the tokenizers tested there (\cref{tab:coco}).

On 504 ImageNet validation images with LlamaGen VQ-16, \RX changes 64\% of
positions. Patches whose token stays fixed account for 32\% of the total
squared-error reduction, consistent with the decoder mixing information across
positions (\cref{fig:token-map-main,sec:supp-token-map}).

\noindent\textbf{Comparison with simpler search.}
With a budget of 50 decoder passes per method, straight-through refinement
and continuous latent refinement with Adam followed by \NN recover 81\%
and 60\%, respectively, of the PSNR gain over \NN achieved by our
reconstruction-only relaxation. The comparison uses $\lambda=0$ and omits
the relaxation's final discrete refinement (\cref{sec:negative}).

The distributional gains depend on the feature space. After bias correction,
four of eleven tokenizers improve in DINOv2 space (\cref{tab:fd}). For
GigaTok B-L with the DINO discriminator, the PSNR gain comes with a 10\%
increase in rFID. For GigaTok B-L-2D, \RX reaches 0.72 rFID. Comparisons with
published scores depend on the crop and codebook size
(\cref{sec:supp-recon}).

\noindent\textbf{Fine texture.}
Reconstruction-only search can smooth fine detail as PSNR improves
(\cref{fig:texture-main}). On 10k ImageNet validation images, a local energy
term raises median finest-scale energy from 54\% to 74\% of the original and
retains a 1.01\,dB PSNR gain over \NN (\cref{tab:sharp-fd}); a pilot without
discrete refinement in either arm shows the same trade-off
(\cref{tab:sharp-fix}).

\subsection{Generation}\label{sec:gen}

\begin{table*}[t]
\caption{\textbf{GPT-B after 100 epochs of training on different token assignments of the same images.}
  Values are means over three seeds, with SD where shown. FID(X,R) compares training images with
  their reconstructions and is measured once per token set.
  Bits are validation bits per token on each model's own token set. gFID-50k uses the training images as
  reference. Held-out gFID uses the validation images (\INh: 5,000; \INk: 50,000, every run at cfg~2.0;
  post hoc).
  $p$ against \NN (Welch, two-sided, uncorrected): \INh gFID-50k $p<0.001$ for \MAP and 0.012 for \RX, held-out
  0.018 for \MAP; \INk gFID-50k 0.003, held-out 0.057.
  Two \NN runs used cfg~2.5 for the precision and recall columns. At matched cfg~2.0 the two arms are comparable.
  The released GPT-B used ten-crop \NN tokens and 300 epochs (FID(X,R): one of the ten crops per image). We report
  its scores under our evaluator.
  Bold and underlining mark the best and second-best scores within a setting (underlining on \INh only).
  Precision, recall and the released model are unranked.}
\label{tab:gen}
\centering\small
\setlength{\tabcolsep}{5.5pt}
\begin{tabular*}{\textwidth}{@{\extracolsep{\fill}}llrrrrrrr@{}}
\toprule
 & tokens & bits $\downarrow$ & gFID-50k $\downarrow$ & held-out gFID $\downarrow$ & FID$_\infty$ $\downarrow$ &
 FID(X,R) $\downarrow$ & Prec. & Rec. \\
\midrule
\multirow{3}{*}{\INh} & \NN & \second{12.22} & 12.05 $\pm$ 0.15 & 16.57 $\pm$ 0.20 & 11.55 & 2.95 & 0.73 & 0.35 \\
 & \RX ($\lambda=0$) & 12.66 & \second{11.54 $\pm$ 0.13} & \second{16.41 $\pm$ 0.14} & \second{11.04} &
   \best{1.48} & 0.72 & 0.36 \\
 & \MAP ($\lambda=0.01$) & \best{12.09} & \best{10.90 $\pm$ 0.16} & \best{15.97 $\pm$ 0.16}
   & \best{10.40} & \second{1.51} & 0.73 & 0.36 \\
\midrule
\multirow{2}{*}{\INk} & \NN & 11.07 & 6.90 $\pm$ 0.19 & 6.79 $\pm$ 0.06 & 6.35 & 1.93 & 0.87 & 0.48 \\
 & \MAP ($\lambda=0.01$) & \best{10.89} & \best{5.77 $\pm$ 0.21} & \best{6.19 $\pm$ 0.27} &
   \best{5.20} & \best{0.99} & 0.84 & 0.53 \\
\midrule
\multicolumn{2}{@{}l}{released GPT-B, 300 epochs} & 10.97 & 5.20 & 5.86 & 4.63 & 1.88 & 0.83 & 0.55 \\
\bottomrule
\end{tabular*}
\end{table*}

\begin{table}[t]
\caption{\textbf{Where the generation gain holds} (GPT-B on \INk, epoch~100).
  Mean $\pm$ SD over three seeds per arm. Two-sided Welch $p$-values, unadjusted; $^{\dagger}$post hoc.
  Bold: the lower score where $p<0.05$. The DINOv2 rows compare both arms at the same cfg.
  \Cref{tab:supp-300} reports epoch~300.}
\label{tab:robust}
\centering\small
\setlength{\tabcolsep}{4pt}
\begin{tabular}{@{}lrrl@{}}
\toprule
\INk, GPT-B, epoch~100 & \NN & \ours & $p$ \\
\midrule
gFID-50k (registered) & 6.90 $\pm$ 0.19 & \best{5.77 $\pm$ 0.21} & 0.003 \\
OpenAI's evaluator, FID & 6.92 $\pm$ 0.21 & \best{5.73 $\pm$ 0.22} & 0.002 \\
second sampling seed & 6.84 $\pm$ 0.14 & \best{5.81 $\pm$ 0.23} & 0.005 \\
FID-10k, no guidance & 34.51 $\pm$ 1.12 & \best{29.55 $\pm$ 0.69} & 0.005 \\
held-out, cfg~2.0$^{\dagger}$ & 6.79 $\pm$ 0.06 & 6.19 $\pm$ 0.27 & 0.057 \\
FD$_\mathrm{DINOv2}$, cfg~2.0$^{\dagger}$ & 327.7 $\pm$ 7.3 & \best{299.9 $\pm$ 4.0} & 0.009 \\
FD$_\mathrm{DINOv2}$, cfg~2.5$^{\dagger}$ & 274.4 $\pm$ 4.9 & \best{257.6 $\pm$ 3.1} & 0.011 \\
\bottomrule
\end{tabular}
\end{table}

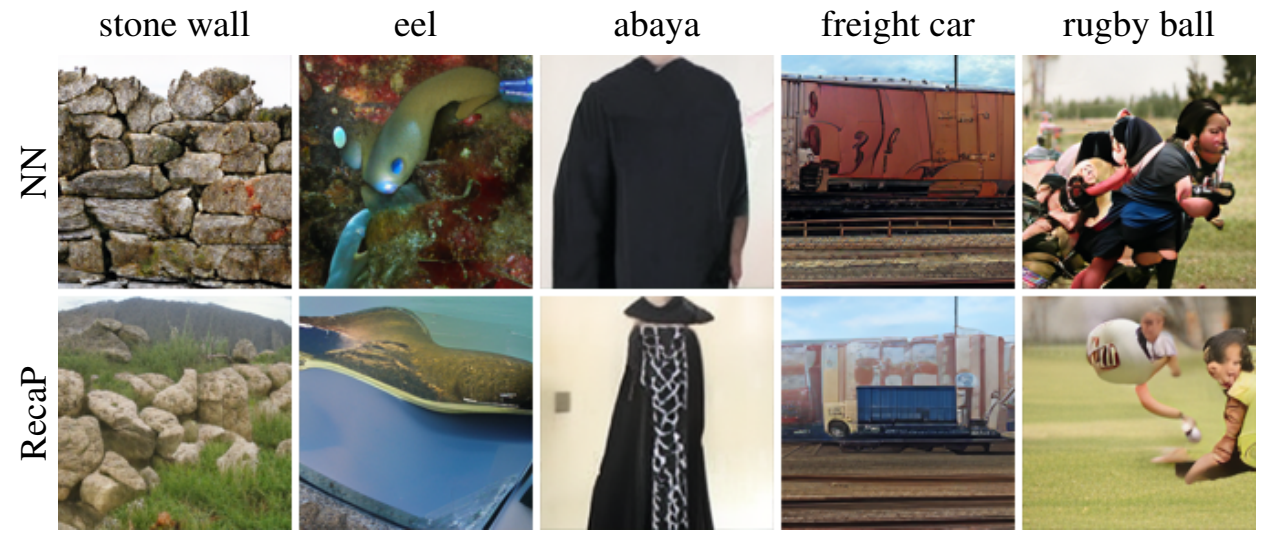
\begin{figure}[t]
\centering
\begin{tikzpicture}
  \node[inner sep=0, anchor=south west] (s) at (0,0) {\includegraphics[width=\dimexpr\linewidth-12pt\relax]{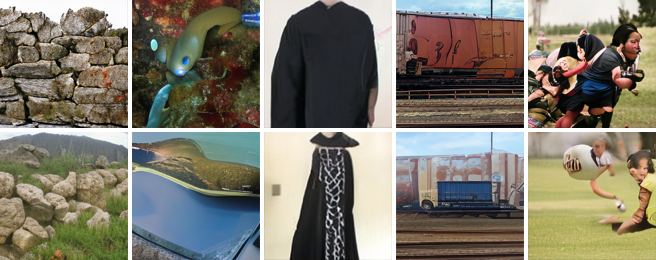}};%
  \begin{scope}[x={(s.south east)}, y={(s.north west)}]
    \foreach \n [count=\c from 0] in {stone wall, eel, abaya, freight car, rugby ball}
      \node[anchor=south, font=\scriptsize, inner sep=1.5pt, text height=1.6ex, text depth=0.4ex]
        at ({(64+132*\c)/656}, 1) {\n};
    \foreach \m/\t in {0/{\NN}, 1/{\ours}}
      \node[anchor=south, rotate=90, font=\scriptsize, inner sep=2pt] at (0, {1-(64+132*\m)/260}) {\t};
  \end{scope}
\end{tikzpicture}
\caption{\textbf{ImageNet-1k samples from GPT-B trained on \NN and \ours tokens.}
  Each row shows the first five of 50k samples at cfg~2.0 from training seed 0, after 100 epochs. The \NN
  row uses the corresponding cfg~2.0 evaluation. Each column shares a class and sampling seed.
  \Cref{fig:samples} shows further samples.}
\label{fig:samples-main}
\end{figure}

After 100 epochs with the same GPT-B training recipe, \ours lowers gFID-50k
against the training images from 6.90 to 5.77 on \INk ($-16\%$, Welch $p=0.003$)
and from 12.05 to 10.90 on \INh ($-9.5\%$, $p<0.001$; \cref{tab:gen}).
On \INk the FID-10k gap is visible by epoch~50
(\cref{fig:teaser-gen,fig:epochs}). At epoch~100 it is largest at low to
moderate guidance (up to cfg~2.0: $-13$ to $-19\%$, every \ours seed below
every \NN seed) and narrows to within seed variation at cfg~2.5--3.0
(\cref{fig:cfg}). Precision and recall are similar at matched cfg~2.0
(\cref{sec:supp-sharp}). \Cref{fig:samples-main} shows samples at this
guidance setting.

The gain is smaller against validation images. In the post hoc comparison at
epoch~100, the held-out difference is $-0.60$ FID on both datasets (\INh:
$p=0.018$; \INk at matched cfg~2.0: $p=0.057$).
At epoch~300 on \INk, every \ours seed still scores below every \NN seed in
training-reference gFID-50k (5.21 vs.\ 5.92, $p=0.079$); against validation
images at matched cfg~2.0, the two means are close (6.09 for \ours and 6.03
for \NN; \cref{tab:supp-300}). \ours and the released ten-crop GPT-B then have
similar training-reference gFID-50k (5.21 and 5.20), and the released model
has lower held-out gFID (5.86; \cref{tab:context}).

The result depends on the generator and on how samples are compared.
On \INh, \ours also lowers the Fr\'echet distance in DINOv2 feature space. On
\INk, it lowers the DINOv2 distance against training images when both arms
use the same cfg (2.0 at epochs~100 and 300, 2.5 at epoch~100; post hoc); at
the pre-registered per-run cfgs (\NN partly at 2.5, \ours at 2.0) the DINOv2
comparison is inconclusive
(\cref{tab:robust,tab:gfd}). With per-run guidance selection, GPT-L falls
within the pre-registered tie range in the one-seed-per-variant comparison.
IBQ AR-B favours \NN without guidance, and its tuned-guidance difference is
within seed variation. The masked generator shows no clear gain
(\cref{tab:prior,sec:supp-impl,sec:negative}).

\subsection{Cost}\label{sec:cost}
On \INk at cfg~2.0, GPT-B trained on \ours tokens for 100 epochs reaches a
similar training-reference FID-10k to the \NN run at epoch~300 (8.08 vs.\
8.21). Including encoding, \ours costs more for one generator. In a post hoc
estimate with three runs sharing the token set, \ours costs less per generator
at comparable FID-10k (\cref{fig:costgen}).
\section{Ablations}\label{sec:ablation}

We use GPT-B to test how the prior, additional training views and search
settings affect generation. Comparisons are pre-registered unless marked
post hoc and use three training seeds per variant unless stated otherwise.

\subsection{Reconstruction, rate and prior}\label{sec:mechanism}

\begin{table}[t]
\caption{\textbf{What carries the gain on \INh} (GPT-B, three seeds per arm, mean $\pm$ SD).
  \MAPo: prior = \INh's own \NN GPT-B (seed 0, epoch~25).
  \MAPx: each half of the images tokenized under a GPT-B trained on the other half's \NN tokens.
  KD: 0.5 cross-entropy + 0.5 distillation from the released GPT-B, temperature 1; bits: as in \cref{tab:gen};
  held-out gFID: pre-registered for KD and \MAPx, post hoc for the rest (\cref{fig:ablation}).
  Bold: best in a column; underlined: second best.}
\label{tab:mech}
\centering\small
\setlength{\tabcolsep}{2.5pt}
\begin{tabular}{@{}llrrr@{}}
\toprule
arm & tokens: prior, $\lambda$ & gFID-50k $\downarrow$ & held-out $\downarrow$ & bits $\downarrow$ \\
\midrule
\NN & none (\NN) & 12.05 $\pm$ 0.15 & 16.57 $\pm$ 0.20 & 12.22 \\
\RX & none, 0 & 11.54 $\pm$ 0.13 & 16.41 $\pm$ 0.14 & 12.66 \\
\MAP & released, 0.01 & 10.90 $\pm$ 0.16 & 15.97 $\pm$ 0.16 & 12.09 \\
\MAP & released, 0.03 & 11.36 $\pm$ 0.05 & 16.55 $\pm$ 0.06 & \second{11.35} \\
\MAPo & own, 0.01 & 11.54 $\pm$ 0.31 & 16.54 $\pm$ 0.14 & 12.13 \\
\MAPx & cross-fitted, 0.01 & 12.02 $\pm$ 0.28 & 16.90 $\pm$ 0.29 & 12.25 \\
\KDNN & none (\NN) & \second{10.00 $\pm$ 0.14} & \second{14.87 $\pm$ 0.84} & 11.43 \\
\KDMAP & released, 0.01 & \best{9.25 $\pm$ 0.21} & \best{13.81 $\pm$ 0.14} & \best{11.29} \\
\bottomrule
\end{tabular}
\end{table}

\noindent\textbf{Reconstruction-only search.}
To test what the prior adds beyond improving the decoded targets, we compare
\RX and \ours at similar reconstruction FID on \INh (1.48 and 1.51 in
\cref{tab:gen}). Adding the prior lowers training-reference gFID from
11.54 to 10.90 ($p=0.006$) at similar reconstruction FID. The held-out gFID
difference between \RX and \NN is small (0.16; post hoc $p=0.33$), and the
held-out mean is lowest for \ours among these three arms
(\cref{tab:gen,tab:mech}). \RX also uses a final discrete refinement,
which adds 0.002\,dB in reconstruction (\cref{sec:supp-impl}).

\RX raises validation NLL by 0.44 bit per token relative to \NN. Both searches
reduce the finest-scale texture energy by a similar amount, consistent with
smoothing from the shared reconstruction loss (\cref{sec:supp-sharp}).

\noindent\textbf{Prior weight.}
On \INh, $\lambda=0.01$ yields the lowest gFID among the tested weights.
At $\lambda=0.03$, the selected maps are more likely under the frozen prior,
and the student's validation NLL on its own token maps is lower.
Training-reference gFID rises from 10.90 to 11.36 ($p=0.026$). The smaller
weight of 0.003 also raises gFID (\cref{tab:mech,tab:supp-ablations,fig:cost}).

\noindent\textbf{Prior predictability.}
Within the GPT-B priors tested on \INh, lower NLL on the same validation
\NN tokens accompanies larger downstream gFID gains
(\cref{tab:prior-predictability}). Both GPT-L priors achieve lower NLL than the
released GPT-B prior and yield smaller gains. In the cross-fitted variant, each
image is encoded using a prior trained on the other half of the
dataset~\cite{dao2021knowledge}.

\begin{table}[t]
\caption{\textbf{Prior predictability and generation on \INh.}
  The NLL advantage is the \NN-trained GPT-B's NLL (12.22 bits/token)
  minus the prior's, both on the same validation \mbox{\NN} tokens. For the
  cross-fitted pair we report prior A. The last column gives the change
  in training-reference gFID-50k from \NN-trained GPT-B (three-seed means).
  We relate NLL to gFID post hoc. \Cref{tab:mech,tab:prior} give the
  individual scores.}
\label{tab:prior-predictability}
\centering\small
\setlength{\tabcolsep}{5pt}
\begin{tabular}{@{}lrr@{}}
\toprule
Prior & \shortstack[r]{NLL advantage\\(bits/token)} & \shortstack[r]{gFID-50k change\\(\%)} \\
\midrule
Cross-fitted GPT-B & 0.16 & $-0.2$ \\
Our GPT-B & 0.40 & $-4.3$ \\
Released GPT-B & 1.21 & $-9.5$ \\
Our GPT-L & 1.23 & $-7.0$ \\
Released GPT-L & 1.49 & $-6.6$ \\
\bottomrule
\end{tabular}
\end{table}

\noindent\textbf{Online distillation.}
We compare two uses of the released GPT-B. It scores candidate tokens
when we encode the training set and supplies next-token probabilities at
every student update in online distillation. Distillation improves gFID on
both token sets on \INh, where the teacher saw ten times as many images as
the students.
Adding it to \ours reduces held-out gFID by 2.15 ($p<0.001$).
On \INk, where both saw the same number of images, distilling the \NN
student improves training-reference gFID by 11.7\%, and \ours alone
still scores lower in all runs. Held-out comparisons are inconclusive
(\cref{tab:mech,sec:supp-gen}). A larger GPT-L teacher gives the \NN
student a similar gFID to \ours in a one-seed test, with training steps
2.7 times as long as without a teacher.

\subsection{Two views against ten crops}\label{sec:crops}

\begin{table}[t]
\caption{\textbf{More views per image: GPT-B on two views (centre crop and flip) or on LlamaGen's ten crops.}
  3 maps: \ours's map and two restarts from noisy logits, per view. Mean $\pm$ SD over three seeds; held-out scores
  use each run's selected cfg (\cref{sec:supp-gen}). Pre-registered; tests in the text. Bold: best in a column;
  underlined: second best. $^{*}$Ten-crop \ours on \INk would take an estimated 3,900 RTX 4090 GPU-h
  (\cref{sec:supp-cost}).}
\label{tab:crops}
\centering\footnotesize
\setlength{\tabcolsep}{1.8pt}
\begin{tabular}{@{}lcrrrr@{}}
\toprule
 & & \multicolumn{2}{c}{\INh} & \multicolumn{2}{c}{\INk} \\
\cmidrule(lr){3-4}\cmidrule(l){5-6}
tokens & views & gFID-50k $\downarrow$ & held-out $\downarrow$ & gFID-50k $\downarrow$ & held-out $\downarrow$ \\
\midrule
\NN & 2 & 12.05 $\pm$ 0.15 & 16.57 & 6.90 $\pm$ 0.19 & 7.482 \\
\ours & 2 & 10.90 $\pm$ 0.16 & 15.97 & \second{5.77 $\pm$ 0.21} & \second{6.193} \\
\ours, 3 maps & 2 & 8.81 $\pm$ 0.22 & 13.75 & \best{5.57 $\pm$ 0.06} & \best{6.186} \\
\NN & 10 & \second{8.26 $\pm$ 0.05} & \second{13.31} & 6.23 $\pm$ 0.08 & 6.436 \\
\ours & 10 & \best{7.61 $\pm$ 0.19} & \best{13.12} & \multicolumn{2}{c}{not run$^{*}$} \\
\bottomrule
\end{tabular}
\end{table}

At matched ten-crop augmentation on \INh, \ours lowers gFID by 7.9\%
relative to \NN ($p=0.022$). The held-out reduction is 1.4\%
($p=0.34$). Ten-crop \NN also outperforms two-view \ours on this subset,
so the number of views matters alongside the assignment (\cref{tab:crops}).
With two views, adding more maps per view improves gFID on \INh and has
a smaller effect on \INk. On \INk, ten crops improve \NN gFID by 9.7\% over
two-view \NN. \ours with two views and three maps per view achieves
10.6\% lower training-reference gFID than ten-crop \NN ($p=0.0005$). With one
map per view, two-view \ours still scores below ten-crop \NN in every seed
(5.77 vs.\ 6.23; $p=0.053$). The matched ten-crop comparison is available on
\INh (\cref{tab:crops}).

\noindent\textbf{Solver and loss.}\phantomsection\label{sec:solver-ablation}
On \INh, shortening the search to 35 steps cuts encoding time by 30\%
and retains a gFID gain over \NN ($p=0.016$). Omitting LPIPS-VGG
doubles the PSNR gain but worsens downstream gFID by 39\% relative to
\ours (\cref{tab:supp-ablations}).
\section{Conclusion}\label{sec:conclusion}

The decoder uses the selected codes jointly, so the effect of changing one
code can depend on the codes at other positions. A local model of the
reconstruction loss casts code assignment as a closest-vector problem with a
decoder-dependent target and metric. For distinct encoder features,
nearest-neighbour lookup solves it for every codebook only when the target is
the encoder output and the metric is isotropic within positions and separable
across them. With the target fixed, the high-resolution model gives a
relative penalty for the lookup metric that is independent of codebook
density. Decoder-aware reassignment raises PSNR for every frozen pretrained
tokenizer we tested. We also derive the finite-sample bias of FID when images
are paired with their reconstructions.

Different token maps can reconstruct the same image nearly equally well.
\ours uses this flexibility to select training targets that are more likely
under a pretrained autoregressive prior. Within the candidate sets, exact MAP
encoding minimizes expected reconstruction loss among encoders that induce the
same joint distribution of classes and token maps. On ImageNet, LlamaGen GPT-B
reaches comparable training-reference FID in fewer epochs with \ours tokens
than with \NN tokens. On \INh, adding the prior lowers generation FID at
comparable reconstruction FID.

\noindent\textbf{Limitations.}\label{sec:limitations}
Our generation experiments use ImageNet. Extending \ours to other domains may
require a prior that predicts their tokens well
(\cref{tab:prior-predictability}). Reusing the tokens across training runs can
offset the one-time encoding cost (\cref{sec:cost}). The local energy term in
\cref{sec:supp-sharp} is a starting point for adding perceptual criteria to
the search.\label{main-end}

\section*{LLM Usage Statement}
Anthropic Claude, used through Claude Code, assisted with code implementation and debugging, experiment
planning, result analysis, and manuscript drafting and revision. \mbox{OpenAI} ChatGPT assisted with editing the
manuscript. The research question and the idea of encoding as MAP inference came from one of the authors,
without the use of LLMs. The authors are responsible for the final manuscript, software, experimental
protocols and reported results.

{
    \small
    \bibliographystyle{ieeenat_fullname}
    \bibliography{refs}

@inproceedings{fifty2025rotation,
  title     = {{Restructuring Vector Quantization with the Rotation Trick}},
  author    = {Fifty, Christopher and Junkins, Ronald G. and Duan, Dennis and Iyengar, Aniketh and Liu, Jerry W. and Amid, Ehsan and Thrun, Sebastian and R{\'e}, Christopher},
  booktitle = {International Conference on Learning Representations (ICLR)},
  year      = {2025},
  note      = {arXiv:2410.06424}
}

@inproceedings{zhu2024simvq,
  title     = {{Addressing Representation Collapse in Vector Quantized Models with One Linear Layer}},
  author    = {Zhu, Yongxin and Li, Bocheng and Xin, Yifei and Xia, Zhihua and Xu, Linli},
  booktitle = {IEEE/CVF International Conference on Computer Vision (ICCV)},
  year      = {2025},
  note      = {SimVQ. arXiv:2411.02038}
}

@inproceedings{mentzer2024fsq,
  title     = {{Finite Scalar Quantization: VQ-VAE Made Simple}},
  author    = {Mentzer, Fabian and Minnen, David and Agustsson, Eirikur and Tschannen, Michael},
  booktitle = {International Conference on Learning Representations (ICLR)},
  year      = {2024},
  note      = {arXiv:2309.15505}
}

@inproceedings{rdvq2026,
  title     = {{Differentiable Vector Quantization for Rate-Distortion Optimization of Generative Image Compression}},
  author    = {Jiang, Shiyin and Long, Wei and Han, Minghao and Chen, Zhenghao and Zhu, Ce and Gu, Shuhang},
  booktitle = {IEEE/CVF Conference on Computer Vision and Pattern Recognition (CVPR)},
  year      = {2026},
  note      = {arXiv:2604.10546}
}

@inproceedings{vandenoord2017neural,
  title     = {{Neural Discrete Representation Learning}},
  author    = {van den Oord, Aaron and Vinyals, Oriol and Kavukcuoglu, Koray},
  booktitle = {Advances in Neural Information Processing Systems (NeurIPS)},
  year      = {2017},
  note      = {VQ-VAE. arXiv:1711.00937}
}

@inproceedings{jang2017gumbel,
  title     = {{Categorical Reparameterization with Gumbel-Softmax}},
  author    = {Jang, Eric and Gu, Shixiang and Poole, Ben},
  booktitle = {International Conference on Learning Representations (ICLR)},
  year      = {2017},
  note      = {arXiv:1611.01144}
}

@inproceedings{yu2024lfq,
  title     = {{Language Model Beats Diffusion -- Tokenizer is Key to Visual Generation}},
  author    = {Yu, Lijun and Lezama, Jos{\'e} and Gundavarapu, Nitesh B. and Versari, Luca and Sohn, Kihyuk and Minnen, David and others},
  booktitle = {International Conference on Learning Representations (ICLR)},
  year      = {2024},
  note      = {MAGVIT-v2, LFQ. arXiv:2310.05737}
}

@inproceedings{zhu2024vqganlc,
  title     = {{Scaling the Codebook Size of VQGAN to 100,000 with a Utilization Rate of 99\%}},
  author    = {Zhu, Lei and Wei, Fangyun and Lu, Yanye and Chen, Dong},
  booktitle = {Advances in Neural Information Processing Systems (NeurIPS)},
  year      = {2024},
  note      = {VQGAN-LC. arXiv:2406.11837}
}

@inproceedings{shi2024ibq,
  title     = {{Scalable Image Tokenization with Index Backpropagation Quantization}},
  author    = {Shi, Fengyuan and Luo, Zhuoyan and Ge, Yixiao and Yang, Yujiu and Shan, Ying and Wang, Limin},
  booktitle = {IEEE/CVF International Conference on Computer Vision (ICCV)},
  year      = {2025}
}

@article{jia2025mgvq,
  title     = {{MGVQ: Could VQ-VAE Beat VAE? A Generalizable Tokenizer with Multi-group Quantization}},
  author    = {Jia, Mingkai and Yin, Wei and Hu, Xiaotao and Guo, Jiaxin and Guo, Xiaoyang and Zhang, Qian and Long, Xiao-Xiao and Tan, Ping},
  journal   = {arXiv preprint arXiv:2507.07997},
  year      = {2025}
}

@inproceedings{yu2024titok,
  title     = {{An Image is Worth 32 Tokens for Reconstruction and Generation}},
  author    = {Yu, Qihang and Weber, Mark and Deng, Xueqing and Shen, Xiaohui and Cremers, Daniel and Chen, Liang-Chieh},
  booktitle = {Advances in Neural Information Processing Systems (NeurIPS)},
  year      = {2024},
  note      = {TiTok. arXiv:2406.07550}
}

@article{tang2026stablevq,
  title     = {{StableVQ: Practical Guidelines for Stable Vector-Quantized Tokenizer Training}},
  author    = {Tang, Bao and Guo, Jiahao and Cao, Haoxiang and Liu, Wenyu and Yu, Changqian and Gai, Kun and Wang, Xinggang},
  journal   = {arXiv preprint arXiv:2609.26774},
  year      = {2026}
}

@article{zador1982asymptotic,
  title   = {{Asymptotic Quantization Error of Continuous Signals and the Quantization Dimension}},
  author  = {Zador, Paul L.},
  journal = {IEEE Transactions on Information Theory},
  volume  = {28},
  number  = {2},
  pages   = {139--149},
  year    = {1982}
}

@article{gersho1979asymptotically,
  title   = {{Asymptotically Optimal Block Quantization}},
  author  = {Gersho, Allen},
  journal = {IEEE Transactions on Information Theory},
  volume  = {25},
  number  = {4},
  pages   = {373--380},
  year    = {1979}
}

@inproceedings{blau2018perception,
  title     = {{The Perception-Distortion Tradeoff}},
  author    = {Blau, Yochai and Michaeli, Tomer},
  booktitle = {IEEE/CVF Conference on Computer Vision and Pattern Recognition (CVPR)},
  year      = {2018},
  note      = {arXiv:1711.06077}
}

@inproceedings{heusel2017gans,
  title     = {{GANs Trained by a Two Time-Scale Update Rule Converge to a Local Nash Equilibrium}},
  author    = {Heusel, Martin and Ramsauer, Hubert and Unterthiner, Thomas and Nessler, Bernhard and Hochreiter, Sepp},
  booktitle = {Advances in Neural Information Processing Systems (NeurIPS)},
  year      = {2017},
  note      = {FID. arXiv:1706.08500}
}

@inproceedings{esser2021taming,
  title     = {{Taming Transformers for High-Resolution Image Synthesis}},
  author    = {Esser, Patrick and Rombach, Robin and Ommer, Bj{\"o}rn},
  booktitle = {IEEE/CVF Conference on Computer Vision and Pattern Recognition (CVPR)},
  year      = {2021},
  note      = {arXiv:2012.09841}
}

@inproceedings{lee2022rqvae,
  title     = {{Autoregressive Image Generation using Residual Quantization}},
  author    = {Lee, Doyup and Kim, Chiheon and Kim, Saehoon and Cho, Minsu and Han, Wook-Shin},
  booktitle = {IEEE/CVF Conference on Computer Vision and Pattern Recognition (CVPR)},
  year      = {2022},
  note      = {arXiv:2203.01941}
}

@inproceedings{chang2022maskgit,
  title     = {{MaskGIT: Masked Generative Image Transformer}},
  author    = {Chang, Huiwen and Zhang, Han and Jiang, Lu and Liu, Ce and Freeman, William T.},
  booktitle = {IEEE/CVF Conference on Computer Vision and Pattern Recognition (CVPR)},
  year      = {2022},
  note      = {arXiv:2202.04200}
}

@article{sun2024llamagen,
  title     = {{Autoregressive Model Beats Diffusion: Llama for Scalable Image Generation}},
  author    = {Sun, Peize and Jiang, Yi and Chen, Shoufa and Zhang, Shilong and Peng, Bingyue and Luo, Ping and Yuan, Zehuan},
  journal   = {arXiv preprint arXiv:2406.06525},
  year      = {2024}
}

@article{luo2024openmagvit2,
  title     = {{Open-MAGVIT2: An Open-Source Project Toward Democratizing Auto-regressive Visual Generation}},
  author    = {Luo, Zhuoyan and Shi, Fengyuan and Ge, Yixiao and Yang, Yujiu and Wang, Limin and Shan, Ying},
  journal   = {arXiv preprint arXiv:2409.04410},
  year      = {2024}
}

@inproceedings{chang2025fvq,
  title     = {{Scalable Training for Vector-Quantized Networks with 100\% Codebook Utilization}},
  author    = {Chang, Yifan and Qin, Jie and Qiao, Limeng and Wang, Xiaofeng and Zhu, Zheng and Ma, Lin and Wang, Xingang},
  booktitle = {International Conference on Learning Representations (ICLR)},
  year      = {2026},
  note      = {FVQ. arXiv:2509.10140}
}

@inproceedings{zhuang2025wetok,
  title     = {{WeTok: Powerful Discrete Tokenization for High-Fidelity Visual Reconstruction}},
  author    = {Zhuang, Shaobin and Guo, Yiwei and Wang, Fangyikang and Fu, Canmiao and Huang, Zhipeng and
               Tian, Zeyue and Li, Xiaohui and Zhang, Ying and Li, Chen and Wang, Yali},
  booktitle = {International Conference on Learning Representations (ICLR)},
  year      = {2026},
  note      = {arXiv:2508.05599}
}

@inproceedings{jayasumana2024rethinking,
  title     = {{Rethinking FID: Towards a Better Evaluation Metric for Image Generation}},
  author    = {Jayasumana, Sadeep and Ramalingam, Srikumar and Veit, Andreas and Glasner, Daniel and Chakrabarti, Ayan and Kumar, Sanjiv},
  booktitle = {IEEE/CVF Conference on Computer Vision and Pattern Recognition (CVPR)},
  year      = {2024}
}

@inproceedings{stein2023exposing,
  title     = {{Exposing Flaws of Generative Model Evaluation Metrics and Their Unfair Treatment of Diffusion Models}},
  author    = {Stein, George and Cresswell, Jesse C. and Hosseinzadeh, Rasa and Sui, Yi and Ross, Brendan Leigh and Villecroze, Valentin and Liu, Zhaoyan and Caterini, Anthony L. and Taylor, J. Eric T. and Loaiza-Ganem, Gabriel},
  booktitle = {Advances in Neural Information Processing Systems (NeurIPS)},
  year      = {2023}
}

@inproceedings{xiong2025gigatok,
  title     = {{GigaTok: Scaling Visual Tokenizers to 3 Billion Parameters for Autoregressive Image Generation}},
  author    = {Xiong, Tianwei and Liew, Jun Hao and Huang, Zilong and Feng, Jiashi and Liu, Xihui},
  booktitle = {IEEE/CVF International Conference on Computer Vision (ICCV)},
  year      = {2025},
  note      = {arXiv:2504.08736}
}

@inproceedings{parmar2022aliased,
  title     = {{On Aliased Resizing and Surprising Subtleties in GAN Evaluation}},
  author    = {Parmar, Gaurav and Zhang, Richard and Zhu, Jun-Yan},
  booktitle = {IEEE/CVF Conference on Computer Vision and Pattern Recognition (CVPR)},
  year      = {2022}
}

@inproceedings{chong2020effectively,
  title     = {{Effectively Unbiased FID and Inception Score and Where to Find Them}},
  author    = {Chong, Min Jin and Forsyth, David},
  booktitle = {IEEE/CVF Conference on Computer Vision and Pattern Recognition (CVPR)},
  year      = {2020}
}

@inproceedings{kynkaanniemi2023role,
  title     = {{The Role of ImageNet Classes in Fr\'echet Inception Distance}},
  author    = {Kynk\"a\"anniemi, Tuomas and Karras, Tero and Aittala, Miika and Aila, Timo and Lehtinen, Jaakko},
  booktitle = {International Conference on Learning Representations (ICLR)},
  year      = {2023}
}

@inproceedings{dhariwal2021diffusion,
  title     = {{Diffusion Models Beat GANs on Image Synthesis}},
  author    = {Dhariwal, Prafulla and Nichol, Alexander},
  booktitle = {Advances in Neural Information Processing Systems (NeurIPS)},
  year      = {2021}
}

@article{babai1986lovasz,
  title     = {{On Lov\'asz' Lattice Reduction and the Nearest Lattice Point Problem}},
  author    = {Babai, L\'aszl\'o},
  journal   = {Combinatorica},
  volume    = {6},
  number    = {1},
  pages     = {1--13},
  year      = {1986}
}

@article{agrell2002closest,
  title     = {{Closest Point Search in Lattices}},
  author    = {Agrell, Erik and Eriksson, Thomas and Vardy, Alexander and Zeger, Kenneth},
  journal   = {IEEE Transactions on Information Theory},
  volume    = {48},
  number    = {8},
  pages     = {2201--2214},
  year      = {2002}
}

@article{chou1989entropy,
  title     = {{Entropy-Constrained Vector Quantization}},
  author    = {Chou, Philip A. and Lookabaugh, Tom and Gray, Robert M.},
  journal   = {IEEE Transactions on Acoustics, Speech, and Signal Processing},
  volume    = {37},
  number    = {1},
  pages     = {31--42},
  year      = {1989}
}

@article{rose1998deterministic,
  title     = {{Deterministic Annealing for Clustering, Compression, Classification, Regression, and Related Optimization
               Problems}},
  author    = {Rose, Kenneth},
  journal   = {Proceedings of the IEEE},
  volume    = {86},
  number    = {11},
  pages     = {2210--2239},
  year      = {1998}
}

@article{ortega1998rate,
  title     = {{Rate-Distortion Methods for Image and Video Compression}},
  author    = {Ortega, Antonio and Ramchandran, Kannan},
  journal   = {IEEE Signal Processing Magazine},
  volume    = {15},
  number    = {6},
  pages     = {23--50},
  year      = {1998}
}

@book{jayant1984digital,
  title     = {{Digital Coding of Waveforms: Principles and Applications to Speech and Video}},
  author    = {Jayant, N. S. and Noll, Peter},
  publisher = {Prentice-Hall},
  year      = {1984}
}

@inproceedings{frantar2023gptq,
  title     = {{GPTQ: Accurate Post-Training Quantization for Generative Pre-trained Transformers}},
  author    = {Frantar, Elias and Ashkboos, Saleh and Hoefler, Torsten and Alistarh, Dan},
  booktitle = {International Conference on Learning Representations (ICLR)},
  year      = {2023}
}

@inproceedings{chen2026gptqbabai,
  title     = {{The Geometry of LLM Quantization: GPTQ as Babai's Nearest Plane Algorithm}},
  author    = {Chen, Jiale and Shabanzadeh, Yalda and Crn{\v{c}}evi{\'c}, Elvir and Hoefler, Torsten and Alistarh, Dan},
  booktitle = {International Conference on Learning Representations (ICLR)},
  year      = {2026},
  note      = {arXiv:2507.18553}
}

@article{gelbrich1990formula,
  title     = {{On a Formula for the $L^2$ Wasserstein Metric between Measures on Euclidean and Hilbert Spaces}},
  author    = {Gelbrich, Matthias},
  journal   = {Mathematische Nachrichten},
  volume    = {147},
  number    = {1},
  pages     = {185--203},
  year      = {1990}
}

@article{dowson1982frechet,
  title     = {{The Fr\'echet Distance between Multivariate Normal Distributions}},
  author    = {Dowson, D. C. and Landau, B. V.},
  journal   = {Journal of Multivariate Analysis},
  volume    = {12},
  number    = {3},
  pages     = {450--455},
  year      = {1982}
}

@inproceedings{cremer2018inference,
  title     = {{Inference Suboptimality in Variational Autoencoders}},
  author    = {Cremer, Chris and Li, Xuechen and Duvenaud, David},
  booktitle = {International Conference on Machine Learning (ICML)},
  year      = {2018}
}

@inproceedings{kim2018semi,
  title     = {{Semi-Amortized Variational Autoencoders}},
  author    = {Kim, Yoon and Wiseman, Sam and Miller, Andrew C. and Sontag, David and Rush, Alexander M.},
  booktitle = {International Conference on Machine Learning (ICML)},
  year      = {2018}
}

@inproceedings{marino2018iterative,
  title     = {{Iterative Amortized Inference}},
  author    = {Marino, Joseph and Yue, Yisong and Mandt, Stephan},
  booktitle = {International Conference on Machine Learning (ICML)},
  year      = {2018}
}

@inproceedings{yang2020improving,
  title     = {{Improving Inference for Neural Image Compression}},
  author    = {Yang, Yibo and Bamler, Robert and Mandt, Stephan},
  booktitle = {Advances in Neural Information Processing Systems (NeurIPS)},
  year      = {2020}
}

@inproceedings{campos2019content,
  title     = {{Content Adaptive Optimization for Neural Image Compression}},
  author    = {Campos, Joaquim and Meierhans, Simon and Djelouah, Abdelaziz and Schroers, Christopher},
  booktitle = {IEEE/CVF Conference on Computer Vision and Pattern Recognition Workshops (CVPRW)},
  year      = {2019}
}

@inproceedings{tian2024var,
  title     = {{Visual Autoregressive Modeling: Scalable Image Generation via Next-Scale Prediction}},
  author    = {Tian, Keyu and Jiang, Yi and Yuan, Zehuan and Peng, Bingyue and Wang, Liwei},
  booktitle = {Advances in Neural Information Processing Systems (NeurIPS)},
  year      = {2024}
}

@inproceedings{li2024mar,
  title     = {{Autoregressive Image Generation without Vector Quantization}},
  author    = {Li, Tianhong and Tian, Yonglong and Li, He and Deng, Mingyang and He, Kaiming},
  booktitle = {Advances in Neural Information Processing Systems (NeurIPS)},
  year      = {2024}
}

@inproceedings{peebles2023dit,
  title     = {{Scalable Diffusion Models with Transformers}},
  author    = {Peebles, William and Xie, Saining},
  booktitle = {IEEE/CVF International Conference on Computer Vision (ICCV)},
  year      = {2023}
}

@inproceedings{kingma2015adam,
  title     = {{Adam: A Method for Stochastic Optimization}},
  author    = {Kingma, Diederik P. and Ba, Jimmy},
  booktitle = {International Conference on Learning Representations (ICLR)},
  year      = {2015}
}

@inproceedings{laobeyer2025highly,
  title     = {{Highly Compressed Tokenizer Can Generate Without Training}},
  author    = {Lao Beyer, Lukas and Li, Tianhong and Chen, Xinlei and Karaman, Sertac and He, Kaiming},
  booktitle = {International Conference on Machine Learning (ICML)},
  year      = {2025}
}

@inproceedings{ramanujan2025worse,
  title     = {{When Worse is Better: Navigating the Compression-Generation Tradeoff in Visual Tokenization}},
  author    = {Ramanujan, Vivek and Tirumala, Kushal and Aghajanyan, Armen and Zettlemoyer, Luke and Farhadi, Ali},
  booktitle = {Advances in Neural Information Processing Systems (NeurIPS)},
  year      = {2025}
}

@inproceedings{wang2025larp,
  title     = {{{LARP}: Tokenizing Videos with a Learned Autoregressive Generative Prior}},
  author    = {Wang, Hanyu and Suri, Saksham and Ren, Yixuan and Chen, Hao and Shrivastava, Abhinav},
  booktitle = {International Conference on Learning Representations (ICLR)},
  year      = {2025}
}

@inproceedings{wu2026alitok,
  title     = {{Towards Sequence Modeling Alignment between Tokenizer and Autoregressive Model}},
  author    = {Wu, Pingyu and Zhu, Kai and Liu, Yu and Tang, Longxiang and Yang, Jian and Peng, Yansong and Zhai, Wei and
               Cao, Yang and Zha, Zheng-Jun},
  booktitle = {International Conference on Learning Representations (ICLR)},
  year      = {2026}
}

@article{lin2026gear,
  title     = {{GEAR: Guided End-to-End AutoRegression for Image Synthesis}},
  author    = {Lin, Bin and Liu, Zheyuan and Lin, Chenguo and Chen, Sixiang and Ge, Yunyang and Lin, Yunlong and
               Zhang, Jianwei and Yang, Miles and Zhong, Zhao and Bo, Liefeng and Yuan, Li},
  journal   = {arXiv preprint arXiv:2606.32039},
  year      = {2026}
}

@inproceedings{nagel2020adaround,
  title     = {{Up or Down? Adaptive Rounding for Post-Training Quantization}},
  author    = {Nagel, Markus and Amjad, Rana Ali and van Baalen, Mart and Louizos, Christos and Blankevoort, Tijmen},
  booktitle = {International Conference on Machine Learning (ICML)},
  year      = {2020}
}

@inproceedings{chee2023quip,
  title     = {{{QuIP}: 2-Bit Quantization of Large Language Models With Guarantees}},
  author    = {Chee, Jerry and Cai, Yaohui and Kuleshov, Volodymyr and De Sa, Christopher},
  booktitle = {Advances in Neural Information Processing Systems (NeurIPS)},
  year      = {2023}
}

@inproceedings{zheng2025rae,
  title     = {{Diffusion Transformers with Representation Autoencoders}},
  author    = {Zheng, Boyang and Ma, Nanye and Tong, Shengbang and Xie, Saining},
  booktitle = {International Conference on Learning Representations (ICLR)},
  year      = {2026},
  note      = {arXiv:2510.11690}
}

@inproceedings{kudo2018subword,
  title     = {{Subword Regularization: Improving Neural Network Translation Models with Multiple Subword Candidates}},
  author    = {Kudo, Taku},
  booktitle = {Annual Meeting of the Association for Computational Linguistics (ACL)},
  year      = {2018}
}

@inproceedings{meister2026langmap,
  title     = {{{LangMAP}: A Language-Adaptive Approach to Tokenization}},
  author    = {Meister, Clara and Salhan, Suchir and Szablewski, Andrzej and Lesci, Pietro and Buttery, Paula and
               Pimentel, Tiago},
  booktitle = {Conference on Empirical Methods in Natural Language Processing (EMNLP)},
  pages     = {49341--49362},
  year      = {2026},
  note      = {arXiv:2606.23566}
}

@inproceedings{agustsson2017soft,
  title     = {{Soft-to-Hard Vector Quantization for End-to-End Learning Compressible Representations}},
  author    = {Agustsson, Eirikur and Mentzer, Fabian and Tschannen, Michael and Cavigelli, Lukas and Timofte, Radu and
               Benini, Luca and Van Gool, Luc},
  booktitle = {Advances in Neural Information Processing Systems (NeurIPS)},
  year      = {2017}
}

@inproceedings{binkowski2018demystifying,
  title     = {Demystifying {MMD} {GANs}},
  author    = {Bi{\'n}kowski, Miko{\l}aj and Sutherland, Danica J. and Arbel, Michael and Gretton, Arthur},
  booktitle = {International Conference on Learning Representations (ICLR)},
  year      = {2018}
}

@inproceedings{schroeder1985celp,
  title     = {{Code-Excited Linear Prediction (CELP): High-Quality Speech at Very Low Bit Rates}},
  author    = {Schroeder, Manfred R. and Atal, Bishnu S.},
  booktitle = {IEEE International Conference on Acoustics, Speech, and Signal Processing (ICASSP)},
  pages     = {937--940},
  year      = {1985}
}

@inproceedings{liao2025vapi,
  title     = {{VA-$\pi$: Variational Policy Alignment for Pixel-Aware Autoregressive Generation}},
  author    = {Liao, Xinyao and He, Qiyuan and Xu, Kai and Qu, Xiaoye and Li, Yicong and Wei, Wei and Yao, Angela},
  booktitle = {IEEE/CVF Conference on Computer Vision and Pattern Recognition (CVPR)},
  year      = {2026},
  note      = {arXiv:2512.19680}
}

@inproceedings{kamper2021dpdp,
  title     = {{Towards Unsupervised Phone and Word Segmentation Using Self-Supervised Vector-Quantized Neural Networks}},
  author    = {Kamper, Herman and van Niekerk, Benjamin},
  booktitle = {Annual Conference of the International Speech Communication Association (Interspeech)},
  year      = {2021},
  note      = {arXiv:2012.07551}
}

@inproceedings{visser2025dpdp,
  title     = {{Spoken Language Modeling with Duration-Penalized Self-Supervised Units}},
  author    = {Visser, Nicol and Kamper, Herman},
  booktitle = {Annual Conference of the International Speech Communication Association (Interspeech)},
  year      = {2025},
  note      = {arXiv:2505.23494}
}

@article{hinton2015distilling,
  title     = {{Distilling the Knowledge in a Neural Network}},
  author    = {Hinton, Geoffrey and Vinyals, Oriol and Dean, Jeff},
  journal   = {arXiv preprint arXiv:1503.02531},
  year      = {2015},
  note      = {NIPS 2014 Deep Learning Workshop}
}

@inproceedings{kim2016sequence,
  title     = {{Sequence-Level Knowledge Distillation}},
  author    = {Kim, Yoon and Rush, Alexander M.},
  booktitle = {Conference on Empirical Methods in Natural Language Processing (EMNLP)},
  year      = {2016},
  note      = {arXiv:1606.07947}
}

@article{peruzzo2026varkd,
  title     = {{Knowledge Distillation for Visual Autoregressive Models}},
  author    = {Peruzzo, Elia and Bhowmik, Aritra and Sautiere, Guillaume and Asano, Yuki M. and Habibian, Amirhossein},
  journal   = {arXiv preprint arXiv:2606.06078},
  year      = {2026},
  note      = {VarKD}
}

@inproceedings{menon2021statistical,
  title     = {{A Statistical Perspective on Distillation}},
  author    = {Menon, Aditya Krishna and Rawat, Ankit Singh and Reddi, Sashank J. and Kim, Seungyeon and Kumar, Sanjiv},
  booktitle = {International Conference on Machine Learning (ICML)},
  year      = {2021}
}

@inproceedings{fang2026vqtransplant,
  title     = {{VQ-Transplant: Efficient VQ-Module Integration for Pre-trained Visual Tokenizers}},
  author    = {Fang, Xianghong and Yuan, Yuan and Kong, Dehan and Rudner, Tim G. J.},
  booktitle = {International Conference on Learning Representations (ICLR)},
  year      = {2026},
  note      = {arXiv:2607.19575}
}

@article{kamper2023word,
  title     = {{Word Segmentation on Discovered Phone Units with Dynamic Programming and Self-Supervised Scoring}},
  author    = {Kamper, Herman},
  journal   = {IEEE/ACM Transactions on Audio, Speech, and Language Processing},
  volume    = {31},
  pages     = {684--694},
  year      = {2023},
  note      = {arXiv:2202.11929}
}

@inproceedings{dao2021knowledge,
  title     = {{Knowledge Distillation as Semiparametric Inference}},
  author    = {Dao, Tri and Kamath, Govinda M. and Syrgkanis, Vasilis and Mackey, Lester},
  booktitle = {International Conference on Learning Representations (ICLR)},
  year      = {2021},
  note      = {arXiv:2104.09732}
}

@article{kitagawa2019convergence,
  title     = {{Convergence of a Newton Algorithm for Semi-Discrete Optimal Transport}},
  author    = {Kitagawa, Jun and M{\'e}rigot, Quentin and Thibert, Boris},
  journal   = {Journal of the European Mathematical Society},
  volume    = {21},
  number    = {9},
  pages     = {2603--2651},
  year      = {2019}
}

@inproceedings{lin2014coco,
  title     = {{Microsoft COCO: Common Objects in Context}},
  author    = {Lin, Tsung-Yi and Maire, Michael and Belongie, Serge and Hays, James and Perona, Pietro and Ramanan, Deva and Doll{\'a}r, Piotr and Zitnick, C. Lawrence},
  booktitle = {European Conference on Computer Vision (ECCV)},
  year      = {2014}
}

@inproceedings{agustsson2017ntire,
  title     = {{NTIRE 2017 Challenge on Single Image Super-Resolution: Dataset and Study}},
  author    = {Agustsson, Eirikur and Timofte, Radu},
  booktitle = {IEEE Conference on Computer Vision and Pattern Recognition Workshops (CVPRW)},
  year      = {2017}
}

@inproceedings{yu2024rar,
  title     = {{Randomized Autoregressive Visual Generation}},
  author    = {Yu, Qihang and He, Ju and Deng, Xueqing and Shen, Xiaohui and Chen, Liang-Chieh},
  booktitle = {IEEE/CVF International Conference on Computer Vision (ICCV)},
  pages     = {18431--18441},
  year      = {2025},
  note      = {arXiv:2411.00776}
}

@article{murata2024g2d2,
  title     = {{G2D2: Gradient-Guided Discrete Diffusion for Inverse Problem Solving}},
  author    = {Murata, Naoki and Lai, Chieh-Hsin and Takida, Yuhta and Uesaka, Toshimitsu and Nguyen, Bac and Ermon, Stefano and Mitsufuji, Yuki},
  journal   = {Transactions on Machine Learning Research (TMLR)},
  year      = {2025},
  note      = {arXiv:2410.14710}
}

@inproceedings{rout2025aps,
  title     = {{Test-Time Anchoring for Discrete Diffusion Posterior Sampling}},
  author    = {Rout, Litu and Lugmayr, Andreas and Jafarian, Yasamin and Varadharajan, Srivatsan and Caramanis, Constantine and Shakkottai, Sanjay and Kemelmacher-Shlizerman, Ira},
  booktitle = {International Conference on Machine Learning (ICML)},
  series    = {PMLR},
  volume    = {306},
  pages     = {105808--105848},
  year      = {2026},
  note      = {arXiv:2510.02291}
}

@article{zhang2026fdloss,
  title     = {{Beyond Token-Level Cross-Entropy: Fr{\'e}chet Distributional Post-Training for Autoregressive Image Generation}},
  author    = {Zhang, Jinhua and Lin, Yisong and Long, Wei and Gu, Shuhang},
  journal   = {arXiv preprint arXiv:2608.00562},
  year      = {2026}
}

@inproceedings{chen2026mpq,
  title     = {{MPQ: A Message-Passing View of Post-Training Quantization}},
  author    = {Chen, Yanshuo and Liu, Chenxi and Shirkavand, Reza and Huang, Heng},
  booktitle = {Advances in Neural Information Processing Systems (NeurIPS)},
  year      = {2026}
}
}

\clearpage
\maketitlesupplementary

\appendix
\appendixrefs

\noindent \Cref{sec:supp-proofs,sec:supp-bias} give the proofs and derive the finite-sample FID bias.
\Cref{sec:supp-recon,sec:supp-impl} provide additional reconstruction results and implementation details.
\Cref{sec:supp-gen} reports generation results, prior comparisons and ablations.
\Cref{sec:supp-cost,sec:negative} document the cost measurements and alternative approaches.
Captions identify ranked tables, where bold and underlining mark the best and second-best values.
References to the main paper are marked ``\maintag'' (e.g., \cref{tab:recon}); references without a marker point
to this supplement.

\section{Theory and proofs}\label{sec:supp-proofs}

We give the assumptions and proofs for the analysis of code assignment, the prior and the relaxation in
\cref{sec:theory,sec:objective,sec:relax,sec:gen-method,sec:mechanism}. \Cref{tab:supp-assumptions}
summarizes the assumptions for each result. \Cref{sec:supp-bias} derives the finite-sample bias used in
\cref{sec:exp-setup}.

\subsection{Setting and assumptions}\label{sec:supp-assumptions}

For image $x$, the encoder $E$ produces features $z=(z_1,\dots,z_n)\in\R^{nd}$ and the decoder $D$
reconstructs from codes in $\mathcal C=\{c_1,\dots,c_K\}\subset\R^d$ (\cref{sec:setup}). Released decoders
accept continuous inputs, so $\ell(q)=\ell(x,D(q))$ is defined for $q\in\R^{nd}$.
For unit-norm codebooks (LlamaGen, GigaTok, TiTok-S-128), we use tangent-space coordinates, including in
the measurements of $\Hm$. The local dimension is then $d-1$.

\paragraph{(A1) Quadratic model.}
$\ell(q)=\ell(z)+g\T(q-z)+\tfrac12(q-z)\T\Hm(q-z)$, with $g=\nabla\ell(z)$ and $\Hm\succ0$ the Gauss--Newton matrix of
$\ell$ at $z$.
The quadratic form is exact for a linear decoder under MSE. We use $\Hm+\mu I$ with $\mu>0$ when
the curvature is singular. For L1 objectives (SimVQ, SEED-Voken), $\Hm$ denotes a positive-definite
curvature model.

\paragraph{(A2) Poisson model (high resolution).}
Near the target of each token, the codes form a homogeneous Poisson process of density $\rho$ in $\R^d$.
We assume independence across tokens and from the targets, although nearby targets can share codes in a
finite codebook. The metric $\Hm$ is constant over the neighbourhood containing the nearest codes.

\paragraph{(A3) Exact minimizer.}
The encoder returns an exact minimizer of \cref{eq:map}.
Only the application of \cref{prop:transport} to our tokens uses (A3).

\paragraph{How far the assumptions hold.}
(A2) idealizes a dense, unstructured codebook.
LlamaGen ($d=8$, 7 on the sphere) is closer to it than IBQ ($d=256$), where it fails.
The local model identifies three sources of mismatch: target shift, anisotropy and coupling.
To assess the magnitude it predicts, we apply an MSE relaxation to eight LlamaGen VQ-16 validation images.
A relaxed code update changes the image with a median of 16\% of the energy predicted by (A1).
When expanded at the \NN codes, (A1) predicts a PSNR 5.5\,dB below \NN for the relaxed map.
The measured PSNR is 2.4\,dB above \NN in this experiment. The L2 + LPIPS search in
\cref{tab:recon} gains 1.27\,dB.
These checks support a qualitative reading of the local model (\cref{sec:limitations}).

\begin{table}[h]
\caption{\textbf{Results of this section and of \cref{sec:supp-bias}}, their assumptions and the evidence for them
  (descriptive; MC: Monte Carlo).  Lem.~\ref{lem:supp-nn-exact}'s conditions fail on trained encoders
  (cosine 0.30--0.54); \cref{cor:fischer}'s factors come from the measured tangent metric (secant metric: 1.05 and
  1.18); \cref{prop:transport} holds for any prior with $p_\theta>0$ and a finite objective, and applies to our tokens
  under (A3), which \cref{alg:relax} approximates.}
\label{tab:supp-assumptions}
\centering\footnotesize
\setlength{\tabcolsep}{4pt}
\begin{tabular}{@{}lll@{}}
\toprule
result & needs & evidence \\
\midrule
Lem.~\ref{lem:supp-square} (CVP) & (A1) & (A1) only qualitative \\
Lem.~\ref{lem:supp-nn-exact} (NN exact) & (A1) & fails on trained encoders \\
Prop.~\ref{prop:amgm} & (A1), (A2) & derivation checked by MC \\
Cor.~\ref{cor:fischer} & (A1), (A2) & factors from measured $\Hm$ \\
Lem.~\ref{lem:supp-ties} (near-ties) & (A2) & fits run overlap \\
Prop.~\ref{prop:transport} & $p_\theta>0$ & applies via (A3) \\
Lem.~\ref{lem:supp-temperature} & none & exact \\
Lem.~\ref{lem:supp-bures}, Cor.~\ref{cor:supp-triangle} & none & exact \\
Lem.~\ref{lem:sqrt}, Cor.~\ref{cor:supp-bures2} & none & exact to second order \\
Prop.~\ref{prop:supp-paired} & $\hat\Sigma\approx\Sigma$, large $N$ & 0.6--2.4\% off a fit \\
Prop.~\ref{prop:supp-gfid} & large $N$ & excess 2.1--4.1\% off subsets \\
\bottomrule
\end{tabular}
\end{table}

\subsection{The quadratic model and the closest-vector problem}\label{sec:supp-cvp}

\begin{lemma}[Completing the square]\label{lem:supp-square}
Under (A1), for every $q\in\R^{nd}$,
\begin{equation}
\ell(q)=\ell(z)-\tfrac12 g\T\Hm^{-1}g+\tfrac12\|q-t\|_{\Hm}^2,\qquad t=z-\Hm^{-1}g.
\label{eq:supp-square}
\end{equation}
Hence $\argmin_{q\in\mathcal C^n}\ell(q)=\argmin_{q\in\mathcal C^n}\|q-t\|_{\Hm}$: a closest-vector problem (CVP) over the
product codebook $\mathcal C^n$ in the metric $\Hm$, with target $t$.
\end{lemma}
\begin{proof}
$\tfrac12(q-z+\Hm^{-1}g)\T\Hm(q-z+\Hm^{-1}g)=\tfrac12(q-z)\T\Hm(q-z)+g\T(q-z)+\tfrac12g\T\Hm^{-1}g$.
Subtracting $\tfrac12g\T\Hm^{-1}g$ and adding $\ell(z)$ gives (A1).
\end{proof}

\noindent \Cref{eq:supp-square} gives the local objective in \cref{eq:quad}. The CVP is over the finite
product set $\mathcal C^n$, as in finite-constellation MIMO detection~\cite{agrell2002closest}.

\paragraph{Independent and sequential assignment.}
Index the tokens by $i$; $t_i$ and $\Hm_{ij}$ are the blocks of $t$ and $\Hm$.
\begin{itemize}\setlength\itemsep{1pt}
\item \emph{NN} picks $q_i=\argmin_{c\in\mathcal C}\|c-z_i\|$ for each token: target $z$, metric $I$.
\item \emph{Babai rounding.}
  For a lattice with basis $L$, Babai's rounding-off rounds the coordinates of $L^{-1}t$ one by
  one~\cite{babai1986lovasz}.
  It is exact for an orthogonal basis.
  NN is the token-wise analogue: by \cref{lem:supp-nn-exact}, for pairwise distinct $z_i$, it is exact for every
  codebook exactly when $\Hm$ is block diagonal with isotropic blocks and $t=z$.
\item \emph{Nearest plane.}
  Fix an order of the tokens.
  Let $\Hm=R\T R$ be the block Cholesky factorization, with $R$ block upper triangular and blocks $R_{ij}\in\R^{d\times d}$.
  For $e=q-t$,
  \begin{equation}
  \|q-t\|_{\Hm}^2=\sum_{i=1}^n(e_i+w_i)\T S_i(e_i+w_i),
  \label{eq:supp-plane}
  \end{equation}
  with $w_i=R_{ii}^{-1}\sum_{j>i}R_{ij}e_j$ and $S_i=R_{ii}\T R_{ii}$.
  The rule picks $q_n,q_{n-1},\dots,q_1$ in turn.
  Token $i$ takes the code nearest to $t_i-w_i$ in the metric $S_i$; $w_i$ is known, since it depends only on later tokens.
  $S_i=\Hm_{ii}-\Hm_{i,<i}\Hm_{<i,<i}^{-1}\Hm_{<i,i}$ is the Schur complement of the block of tokens before $i$ in
  $\Hm_{\le i,\le i}$: token $i$'s metric once the earlier tokens, decided later, are optimized out.
  This is Babai's nearest-plane algorithm for $\mathcal C^n$~\cite{babai1986lovasz}: successive interference cancellation
  in MIMO detection, GPTQ in weight quantization~\cite{chen2026gptqbabai} and DPCM in source
  coding~\cite{jayant1984digital}.
\end{itemize}
\Cref{eq:supp-plane} is $\|q-t\|_{\Hm}^2=\|Re\|^2$ written block row by block row.

\begin{lemma}[When NN is exact]\label{lem:supp-nn-exact}
Fix the encoder features $z$, with pairwise distinct $z_i$, and the gradient $g$.
Under (A1), NN minimizes $\ell$ over $\mathcal C^n$ for every finite codebook $\mathcal C\subset\R^d$ if and only if $g=0$
and $\Hm=\operatorname{blockdiag}(\alpha_1I_d,\dots,\alpha_nI_d)$ with $\alpha_i>0$.
\end{lemma}
\begin{proof}
If $g=0$ and $\Hm$ has that form, $\|q-z\|_{\Hm}^2=\sum_i\alpha_i\|q_i-z_i\|^2$, which NN minimizes token by token.
For the converse, let $F(e)=g\T e+\tfrac12e\T\Hm e$ be the model's change for $e=q-z$.
Fix unit vectors $a,b\in\R^d$, $\eta>0$ and a small $\varepsilon>0$.
Every codebook below contains $z_j$ for the tokens $j$ not named, so NN keeps them.
(i) Give token $i$ the candidates $z_i+\varepsilon a$ and $z_i-\varepsilon(1+\eta)a$.
NN picks the first, so $F(\varepsilon a)\le F(-\varepsilon(1+\eta)a)$ with only block $i$ nonzero.
This reads $(2+\eta)\,g_i\T a\le\tfrac12\varepsilon\big((1+\eta)^2-1\big)a\T\Hm_{ii}a$.
Letting $\varepsilon\to0$ and replacing $a$ by $-a$ gives $g_i=0$.
(ii) With $g=0$, give token $i$ the candidates $z_i+\varepsilon a$ and $z_i+\varepsilon(1+\eta)b$.
Then $a\T\Hm_{ii}a\le(1+\eta)^2\,b\T\Hm_{ii}b$.
Letting $\eta\to0$ and swapping $a$ and $b$ makes the Rayleigh quotient of $\Hm_{ii}$ constant, so $\Hm_{ii}=\alpha_iI$.
(iii) For $j\ne i$, give token $i$ the candidates $z_i+\varepsilon a$, $z_i-\varepsilon(1+\eta)a$ and token $j$ the
candidates $z_j+\varepsilon b$, $z_j-\varepsilon(1+\eta)b$.
NN picks $(z_i+\varepsilon a,z_j+\varepsilon b)$.
Against $(z_i+\varepsilon a,z_j-\varepsilon(1+\eta)b)$ this gives
$\tfrac12\alpha_j+a\T\Hm_{ij}b\le\tfrac12\alpha_j(1+\eta)^2-(1+\eta)\,a\T\Hm_{ij}b$.
Letting $\eta\to0$ gives $a\T\Hm_{ij}b\le0$ for all unit $a,b$, so $\Hm_{ij}=0$.
\end{proof}

\noindent The two conditions characterize the mismatches in \cref{sec:theory}. The target can shift away
from $z$, and the metric can be anisotropic within positions or couple different positions.

\paragraph{The target and the encoder's training.}
Let $u_{\mathrm c}=2\beta_{\mathrm c}/(nd)$ for a commitment weight $\beta_{\mathrm c}$ on the mean squared commitment term.
The straight-through gradient of $\ell(q)+\tfrac{\beta_{\mathrm c}}{nd}\sum_i\|z_i-q_i\|^2$ with respect to $z_i$ is
$g_i(q)+u_{\mathrm c}(z_i-q_i)$, where $g_i(q)=\partial\ell/\partial q_i$ at the hard codes $q$.
At a stationary point $z_i=q_i-g_i(q)/u_{\mathrm c}$, and for every code $c$
\begin{equation*}
\|z_i-c\|^2-\|z_i-q_i\|^2=\tfrac2{u_{\mathrm c}}\big(g_i(q)\T(c-q_i)+\tfrac{u_{\mathrm c}}2\|c-q_i\|^2\big).
\end{equation*}
So NN would switch token $i$ to $c$ exactly when the quadratic model around $q$ with Hessian $u_{\mathrm c}I$ predicts a decrease.
(Expand $\|(q_i-c)-g_i/u_{\mathrm c}\|^2-\|g_i/u_{\mathrm c}\|^2$.)
Trained encoders do not sit at this stationary point: on our own ImageNet tokenizers ($K=1024$ and 16,384, MSE and
LPIPS objectives), $\cos(z_i-q_i,-g_i(q))$ is 0.30--0.54.
A GAN term changes the straight-through gradient used to train the encoder.
The assignment loss $\ell$ omits this term, which can lead to a mismatch even
when changing one token. In our $K=1024$ tokenizers, with architecture, data
and search budget fixed, the PSNR gain from relaxation is 2.6 times larger
after GAN training (1.07--1.15 vs.\ 0.43\,dB). Single-token search gains 0.4\,dB with GAN training and 0.02\,dB without it.

\subsection{Metric mismatch: proof of \texorpdfstring{\cref{prop:amgm}}{Proposition \getrefnumber{prop:amgm}}}\label{sec:supp-amgm}

\noindent To isolate the effect of the metric in \cref{sec:theory}, we fix the target and assume the
high-resolution regime~\cite{zador1982asymptotic,gersho1979asymptotically}: near each target the codes form
a homogeneous Poisson process of density $\rho$ in $\R^d$, independent of the targets and across tokens, and $\Hm$ is
constant over the region that holds the nearest codes.
Index the tokens by $i$; $z_i$, $q_i$, $t_i$ and $\Hm_{ii}$ are the blocks of $z$, $q$, $t$ and $\Hm$ of
\cref{eq:quad}.
\begin{proposition}[Metric mismatch]\label{prop:amgm}
Assigning token $i$ to the code nearest to its target $t_i$ in a metric $A\succ0$ gives, with $e=q_i-t_i$,
\begin{equation*}
\E\big[e\T\Hm_{ii}e\big]=\tfrac{\Gamma(1+2/d)}{d}(\rho V_d)^{-2/d}\det(A)^{1/d}\operatorname{tr}(A^{-1}\Hm_{ii}),
\end{equation*}
twice token $i$'s expected excess in \cref{eq:quad} when the other tokens sit at their targets; $V_d$ is the volume of
the unit ball.
It is minimal for $A\propto\Hm_{ii}$, and NN's metric ($A=I$) exceeds that minimum by the ratio
$\mathrm{AM}(\Hm_{ii})/\mathrm{GM}(\Hm_{ii})$ of the block's eigenvalues.
\end{proposition}
\noindent The proof below whitens the metric, and a Monte Carlo check of the ratio agrees within its noise.
Over coupled tokens $\mathrm{AM}(\Hm)/\mathrm{GM}(\Hm)$ bounds NN's error relative to nearest-plane decoding
(successive rounding in the metric $\Hm$) and splits into a per-token and a coupling factor, both $\ge1$
(\cref{cor:fischer}).
Under the same high-resolution assumptions, \cref{lem:supp-ties} gives the expected number of additional
codes within a factor $1+\varepsilon$ of the best squared distance: $(1+\varepsilon)^{d/2}-1=0.46$ for $d=8$
and $1+\varepsilon=1.1$ (0.40 for unit-norm codes).

\Cref{prop:amgm} concerns one token.
Its metric is a fixed $d\times d$ matrix $B\succ0$: the block $\Hm_{ii}$, or $S_i$ in the nearest-plane rule.
The code is chosen as the nearest to the token's target $t_i$, so the proposition isolates the metric and holds the
target fixed.
The error is $e=q_i-t_i$.
For $B=\Hm_{ii}$, $e\T Be$ is twice the excess of the model over its continuous minimum when the other tokens sit at
their targets (\cref{lem:supp-square}).
Let $V_d=\pi^{d/2}/\Gamma(1+d/2)$ be the volume of the unit ball.

\smallskip\noindent\textbf{\Cref{prop:amgm}, precise form.}
\emph{Let the codes form a homogeneous Poisson process $\Pi$ of density $\rho$ in $\R^d$, let $A,B\succ0$, and let
$c_A=\argmin_{c\in\Pi}(c-t_i)\T A(c-t_i)$ and $e=c_A-t_i$.
Then}
\begin{equation}
\E\big[e\T Be\big]=\frac{\Gamma(1+2/d)}{d}\,(\rho V_d)^{-2/d}\,\det(A)^{1/d}\operatorname{tr}(A^{-1}B).
\label{eq:supp-amgm}
\end{equation}
\emph{Over $A$ this is minimal if and only if $A\propto B$.
NN ($A=I$) exceeds the minimum by the factor $\mathrm{AM}(B)/\mathrm{GM}(B)=\tfrac1d\operatorname{tr}B/\det(B)^{1/d}\ge1$.}

\begin{proof}
\emph{Whitening.}
Let $u=A^{1/2}(c-t_i)$.
The map $c\mapsto u$ is affine and bijective, with Jacobian determinant $\det(A)^{1/2}$.
By the mapping theorem, the codes in $u$ form a homogeneous Poisson process of density $\rho'=\rho\det(A)^{-1/2}$.
The code nearest in the metric $A$ is the point $u^\star$ nearest to the origin in the Euclidean norm.

\emph{Distance.}
$P(\|u^\star\|>s)$ is the probability of no point in the ball of radius $s$: $\exp(-\rho'V_ds^d)$.
Hence
\[
\E\|u^\star\|^2=\int_0^\infty e^{-\rho'V_dv^{d/2}}\,dv=\Gamma(1+\tfrac2d)\,(\rho'V_d)^{-2/d},
\]
by the substitution $w=\rho'V_dv^{d/2}$, which turns the integral into $\tfrac2d(\rho'V_d)^{-2/d}\Gamma(\tfrac2d)$.

\emph{Isotropy.}
The process is invariant in law under rotations about the origin, so $u^\star$ is too and
$\E[u^\star u^{\star\top}]=\tfrac1d\E\|u^\star\|^2\,I$.

\emph{Excess.}
$e=A^{-1/2}u^\star$, so $\E[e\T Be]=\E[u^{\star\top}A^{-1/2}BA^{-1/2}u^\star]=\tfrac1d\E\|u^\star\|^2\operatorname{tr}(A^{-1}B)$.
Substituting $(\rho'V_d)^{-2/d}=(\rho V_d)^{-2/d}\det(A)^{1/d}$ gives \cref{eq:supp-amgm}.

\emph{Minimum.}
Let $\mu_1,\dots,\mu_d>0$ be the eigenvalues of $A^{-1}B$, which is similar to $A^{-1/2}BA^{-1/2}\succ0$.
Since $\prod_k\mu_k=\det B/\det A$,
\begin{align*}
\det(A)^{1/d}\operatorname{tr}(A^{-1}B)&=\det(B)^{1/d}\,\frac{\sum_k\mu_k}{(\prod_k\mu_k)^{1/d}}\\
&\ge d\det(B)^{1/d}
\end{align*}
by the AM--GM inequality, with equality if and only if all $\mu_k$ are equal, that is $A\propto B$.
At $A=I$ the left side is $\operatorname{tr}B$, which is $\mathrm{AM}(B)/\mathrm{GM}(B)$ times the minimum.
\end{proof}

\noindent A Monte Carlo check (4,000 targets per metric, the two metrics drawn independently) gives ratios 1.008, 1.223
and 1.611 for $d=2$ and 1.044, 1.129 and 1.305 for $d=4$ (eigenvalue spreads 2, 4 and 8), against 1.061, 1.250, 1.591,
1.034, 1.138 and 1.326 predicted.
For the largest deviation ($d=2$, spread 2), a paired rerun with 40,000 targets gives $1.060\pm0.001$.

\paragraph{Dependence on codebook size.}
Both sides of the ratio scale as $\rho^{-2/d}$, so the ratio does not depend on $K$.
A larger codebook shrinks the quantization excess of NN and of the metric-matched rule alike.
Write $\ell=\ell_0+\ell_q$, with $\ell_0$ the part that no assignment removes and $\ell_q$ NN's quantization excess.
If $\ell$ is the MSE, a rule with ratio $r$ gains $10\log_{10}\big((\ell_0+\ell_q)/(\ell_0+\ell_q/r)\big)$\,dB, which falls only as $\ell_q$ falls.
SimVQ gains 1.22\,dB at $K=2^{10}$ and 1.04\,dB at $2^{18}$; IBQ gains 1.17 and 1.09\,dB at $2^{14}$ and $2^{18}$.
The trend is consistent with the prediction; the comparison spans different released models, and IBQ ($d=256$)
lies outside the regime of (A2).

\subsection{Per-token and coupling factors}\label{sec:supp-fischer}

For $M\succ0$ of size $m$ let $\mathrm{AM}(M)=\tfrac1m\operatorname{tr}M$ and $\mathrm{GM}(M)=\det(M)^{1/m}$.
Let $\mathrm{GM}_b=(\prod_i\det\Hm_{ii})^{1/(nd)}$, the GM of $\operatorname{blockdiag}(\Hm_{11},\dots,\Hm_{nn})$,
and $\chi=\tfrac{\Gamma(1+2/d)}{d}(\rho V_d)^{-2/d}$.

\begin{corollary}[Per-token and coupling factors]\label{cor:fischer}
(i) \emph{Algebra.}
\[
\frac{\mathrm{AM}(\Hm)}{\mathrm{GM}(\Hm)}=\frac{\mathrm{AM}(\Hm)}{\mathrm{GM}_b}\cdot\frac{\mathrm{GM}_b}{\mathrm{GM}(\Hm)}.
\]
The first factor is ${\ge}1$ by the AM--GM inequality, with equality iff $\operatorname{blockdiag}(\Hm_{ii})\propto I$.
The second is ${\ge}1$ by Fischer's inequality $\det\Hm\le\prod_i\det\Hm_{ii}$, with equality iff $\Hm$ is block
diagonal.

\noindent(ii) \emph{Operational factors.}
Under (A1) and (A2), with every rule aimed at the target $t$, the expected values of $e\T\Hm e$ (twice the excesses,
\cref{prop:amgm}) are
\begin{align*}
E_{\mathrm{NN}}&=\chi\operatorname{tr}\Hm,\\
E_{\mathrm{tok}}&=\chi\,d\textstyle\sum_i\det(\Hm_{ii})^{1/d},\\
E_{\mathrm{plane}}&=\chi\,d\textstyle\sum_i\det(S_i)^{1/d}
\end{align*}
for NN, for the best per-token metric ($A_i\propto\Hm_{ii}$) and for the nearest-plane rule.
The per-token factor $r_{\mathrm{tok}}=E_{\mathrm{NN}}/E_{\mathrm{tok}}$ and the coupling factor
$r_{\mathrm{coup}}=E_{\mathrm{tok}}/E_{\mathrm{plane}}$ are ${\ge}1$, and
\[
r_{\mathrm{tok}}\,r_{\mathrm{coup}}=\frac{E_{\mathrm{NN}}}{E_{\mathrm{plane}}}\le\frac{\mathrm{AM}(\Hm)}{\mathrm{GM}(\Hm)},
\]
with equality iff all $\det S_i$ are equal.
If all $\det\Hm_{ii}$ are equal, $r_{\mathrm{tok}}=\mathrm{AM}(\Hm)/\mathrm{GM}_b$.
If all $\det S_i$ are equal as well, $r_{\mathrm{coup}}=\mathrm{GM}_b/\mathrm{GM}(\Hm)$.
\end{corollary}

\begin{proof}
(i) $\operatorname{tr}\Hm=\sum_i\operatorname{tr}\Hm_{ii}$, so the first factor is AM over GM of the eigenvalues of
$\operatorname{blockdiag}(\Hm_{ii})$.
For the second, $\det\Hm=\det(R)^2=\prod_i\det S_i$ by the block Cholesky factorization.
Each Schur complement satisfies $\Hm_{ii}-S_i=\Hm_{i,<i}\Hm_{<i,<i}^{-1}\Hm_{<i,i}\succeq0$, so $0\prec S_i\preceq\Hm_{ii}$
and $\det S_i\le\det\Hm_{ii}$.
The product gives Fischer's inequality.
Equality forces $S_i=\Hm_{ii}$, that is $\Hm_{i,<i}=0$, for every $i$.

(ii) A per-token rule chooses each token on its own.
By (A2) the errors $e_i$ are independent, and zero-mean by the isotropy in the proof of \cref{prop:amgm}.
So the cross terms vanish and $\E[e\T\Hm e]=\sum_i\E[e_i\T\Hm_{ii}e_i]$.
\Cref{eq:supp-amgm} with $B=\Hm_{ii}$ and $A=I$ or $A\propto\Hm_{ii}$ gives $E_{\mathrm{NN}}$ and $E_{\mathrm{tok}}$.
For the nearest-plane rule use \cref{eq:supp-plane}.
When token $i$ is chosen, $w_i$ is fixed and independent of the codes near $t_i$.
\Cref{eq:supp-amgm} with $A=B=S_i$ and target $t_i-w_i$ gives $\E[(e_i+w_i)\T S_i(e_i+w_i)]=\chi d\det(S_i)^{1/d}$.
Now $r_{\mathrm{tok}}\ge1$ follows from $\operatorname{tr}\Hm_{ii}\ge d\det(\Hm_{ii})^{1/d}$ for each token, and
$r_{\mathrm{coup}}\ge1$ from $\det S_i\le\det\Hm_{ii}$.
By the AM--GM inequality over tokens, $\sum_i\det(S_i)^{1/d}\ge n\big(\prod_i\det S_i\big)^{1/(nd)}=n\,\mathrm{GM}(\Hm)$, with
equality iff all $\det S_i$ are equal.
This bounds $E_{\mathrm{NN}}/E_{\mathrm{plane}}=\operatorname{tr}\Hm/\big(d\sum_i\det(S_i)^{1/d}\big)$ by
$\mathrm{AM}(\Hm)/\mathrm{GM}(\Hm)$.
With equal determinants the sums are $n$ times one term, which gives the last two identities.
\end{proof}

\noindent We measure the operational factors in part (ii) to quantify the roles of anisotropy and coupling.

\paragraph{Measured factors.}
On LlamaGen VQ-16 (MSE metric at the NN codes, 8 images), $r_{\mathrm{tok}}=1.23$, $r_{\mathrm{coup}}=1.96$ and
$\mathrm{AM}(\Hm)/\mathrm{GM}(\Hm)=6.84$ (means).
The product $r_{\mathrm{tok}}r_{\mathrm{coup}}$ is below the AM/GM bound on every image. In
\cref{cor:fischer}~(ii), the gap to the bound comes from variation in $\det S_i$ across positions.
At the codebook's scale the secant metric gives 1.05, 1.18 and 1.85 instead.
With an exact quadratic model (a linear decoder whose tokens overlap, so that they couple) the nearest-plane rule
closes 67\% (isotropic) and 85\% (anisotropic) of the gap between NN and the exact optimum at $K=16$; without overlap
the per-token rule is already exact.
Both factors approach their predicted values from below as $K$ grows (post hoc for the first of three seeds).
On two CIFAR tokenizers ($K=256$ and 1024), where $\Hm$ can be formed per image, the relaxation's gain splits into
the target (28\% and 39\%), the per-token metric (4\% and 11\%), coupling (39\% and 34\%) and a part beyond the
quadratic model (30\% and 17\%; post hoc, including the choice of damping; rounded).

\subsection{Codes and token maps with similar reconstruction}\label{sec:supp-flat}

\begin{lemma}[Near-ties]\label{lem:supp-ties}
Under (A2), consider one token assigned in its own metric $B\succ0$.
Let $r^2$ be the smallest value of $(c-t_i)\T B(c-t_i)$ over the codes, and let $N_\varepsilon$ count the other codes with
$(c-t_i)\T B(c-t_i)\le(1+\varepsilon)r^2$.
Then $N_\varepsilon$ is geometric: for $m=0,1,\dots$,
\[
P(N_\varepsilon=m)=(1+\varepsilon)^{-d/2}\big(1-(1+\varepsilon)^{-d/2}\big)^m,
\]
so $\E N_\varepsilon=(1+\varepsilon)^{d/2}-1$ and $P(N_\varepsilon\ge1)=1-(1+\varepsilon)^{-d/2}$.
Neither depends on $\rho$ or $B$.
\end{lemma}
\begin{proof}
Whiten with $B^{1/2}$ as in the proof of \cref{prop:amgm}.
The codes become a Poisson process of density $\rho'$ around the target, and $r$ is the distance to the nearest point.
Let $W=\rho'V_dr^d$; $P(W>w)=e^{-w}$, so $W$ is exponential with mean 1.
Given $r$, the points outside the closed ball of radius $r$ form a Poisson process of density $\rho'$, independent of the
ball's contents.
The shell $r<\|u\|\le r\sqrt{1+\varepsilon}$ has volume $V_dr^d\big((1+\varepsilon)^{d/2}-1\big)$.
So, given $W$, $N_\varepsilon$ is Poisson with mean $\gamma W$, where $\gamma=(1+\varepsilon)^{d/2}-1$.
Mixing over $W$,
\[
P(N_\varepsilon=m)=\int_0^\infty e^{-w}e^{-\gamma w}\frac{(\gamma w)^m}{m!}\,dw=\frac{\gamma^m}{(1+\gamma)^{m+1}},
\]
a geometric law with mean $\gamma$ and $P(N_\varepsilon=0)=1/(1+\gamma)=(1+\varepsilon)^{-d/2}$.
\end{proof}

\paragraph{Expected number of nearby alternatives.}
For $d=8$ and $\varepsilon=0.1$, $\E N_\varepsilon=1.1^4-1\approx0.46$.
For $\varepsilon=0.25$ it is $1.25^4-1\approx1.44$.
A further code within $(1+\varepsilon)r^2$ raises this token's excess $e\T Be$ by at most the factor $1+\varepsilon$.
So at $d=8$ a token has on average about half a further code within 10\% of its best excess.

\paragraph{Scope.}
The lemma counts alternatives for one token in its matched metric. For unit-norm codes the local dimension
is $d-1$, and the mean is $(1+\varepsilon)^{(d-1)/2}-1$. Coupling can affect how these alternatives combine into
full maps. Empirically, runs from perturbed starts share 55\% of their tokens for \RX and 58\% for \MAP at equal
PSNR. These overlaps describe full maps, whereas the lemma describes individual positions.
Runs with the prior also return different maps. In an exploratory check without a prior, a majority vote
over six runs reduces the rate by only 0.05 bit per token relative to one run. This supports the role of
prior probability in selecting lower-rate maps.

\subsection{Prior and training distribution: proof of \texorpdfstring{\cref{prop:transport}}{Proposition \getrefnumber{prop:transport}}}\label{sec:supp-transport}

\noindent We characterize the reconstruction loss of encoders that induce the token distribution in \cref{sec:gen-method}.
Let $\psi_y(k)=\frac{\lambda}{n}\big(-\log_2p_\theta(k\mid y)\big)$, the rate term of \cref{eq:map}, and let
$f_p(x,y)$ be an exact minimizer of $\ell(x,k)+\psi_y(k)$ over $\mathcal K(x)$, with $\ell(x,k)=\ell(x,D(q(k)))$.
Let $Q_p$ be the joint distribution of $(y,f_p(x,y))$ over the data.
\begin{proposition}[The prior selects the training distribution]\label{prop:transport}
Every encoder, deterministic or stochastic, that outputs token maps in $\mathcal K(x)$ and whose pairs of class and
token map follow $Q_p$ has expected loss at least $\E\,\ell(x,f_p(x,y))$.
\end{proposition}
\noindent For an encoder $\xi(k\mid x,y)$ supported on $\mathcal K(x)$ with output $Q_p$,
$\E_\xi[\ell]=\E_\xi[\ell+\psi]-\E_{Q_p}[\psi]\ge\E[\min_{k\in\mathcal K(x)}(\ell+\psi_y(k))]-\E_{Q_p}[\psi]
=\E\,\ell(x,f_p(x,y))$.
The prior and the MAP tie-breaking rule determine the output law. Exact MAP minimizes reconstruction
loss among encoders with that law. We state the equality condition below.

\paragraph{Likelihood interpretation.}
With $\psi_y$ as above, \cref{eq:map} minimizes
$\ell(x,D(q(k)))+\psi_y(k)$ over $k\in\mathcal K(x)$.
Multiplying by $\beta=n\ln2/\lambda$ turns it into maximizing $\ln p_\theta(k\mid y)-\beta\,\ell(x,D(q(k)))$.
At $\lambda=0.01$ and $n=256$, $\beta\approx1.8\times10^4$.
If $p(x\mid k)=e^{-\beta\ell(x,D(q(k)))}/Z$ with $Z$ independent of $k$, this is the MAP estimate of $k$.
$Z$ is independent of $k$ for MSE, where $p(x\mid k)$ is Gaussian; for LPIPS it is an assumption.
Since $p(x\mid y)=\sum_kp_\theta(k\mid y)p(x\mid k)$, every $k$ gives
$\ln p(x\mid y)\ge\ln p_\theta(k\mid y)+\ln p(x\mid k)$, and $k^\star$ gives the tightest such bound among the
candidates.
No result below uses this reading.

\paragraph{Setting.}
Let $(x,y)\sim P$.
For $k\in[K]^n$ let $c(x,k)=\ell(x,D(q(k)))$ if $k\in\mathcal K(x)$ and $+\infty$ otherwise.
An \emph{encoder} is a Markov kernel $\xi(k\mid x,y)$; a deterministic encoder is a measurable map.
Its \emph{output law} is the law of $(y,k)$.
Let $f^\star(x,y)$ be a measurable selection of $\argmin_k\,[c(x,k)+\psi_y(k)]$, for instance the smallest index among the
minimizers, and let $Q^\star$ be the law of $(y,f^\star(x,y))$.
These are $f_p$ and $Q_p$ of \cref{prop:transport}.
Assume $p_\theta(k\mid y)>0$ for all $k$ and $\E\big[c(x,f^\star)+\psi_y(f^\star)\big]<\infty$.

\smallskip\noindent\textbf{\Cref{prop:transport}, precise form.}
\emph{Every encoder $\xi$ whose output law is $Q^\star$ satisfies}
\[
\E_\xi\big[c(x,k)\big]\ge\E\big[c(x,f^\star(x,y))\big],
\]
\emph{with equality if and only if $\xi(\cdot\mid x,y)$ is supported on $\argmin_k[c(x,k)+\psi_y(k)]$ for $P$-almost every
$(x,y)$.}

\begin{proof}
If $\E_\xi[c]=\infty$ there is nothing to prove.
Otherwise write $\E_\xi[c]=\E_\xi[c+\psi]-\E_\xi[\psi]$.
For the first term,
$\E_\xi[c+\psi]=\E_{(x,y)}\sum_k\xi(k\mid x,y)\big(c(x,k)+\psi_y(k)\big)\ge\E\min_k\big(c(x,k)+\psi_y(k)\big)$.
Its right side is $\E[c(x,f^\star)+\psi_y(f^\star)]$, and equality holds iff $\xi$ puts no mass outside the minimizers,
almost surely.
The second term depends on $\xi$ only through the law of $(y,k)$, which is $Q^\star$, the law of $(y,f^\star)$.
So $\E_\xi[\psi]=\E[\psi_y(f^\star)]<\infty$.
Subtracting gives the claim.
\end{proof}

\paragraph{Reading.}
\begin{itemize}\setlength\itemsep{1pt}
\item \emph{Joint law.}
  The prior is class-conditional, so the encoder must match the law of $(y,k)$, the training distribution of a
  class-conditional generator.
  Matching the law of $k$ alone does not suffice, because $\psi_y$ depends on $y$.
\item \emph{Ties.}
  If the minimizer is not unique on a set of positive probability, each selection $f^\star$ (or a randomized tie-break)
  has its own $Q^\star$, and the statement holds for each.
  If ties have probability zero, $Q^\star$ is unique.
\item \emph{Transport.}
  This is the elementary half of semi-discrete optimal transport: $f^\star$ is an optimal map from $P$ to its own image
  $Q^\star$ for the cost $c$, with the prices $\psi$ as dual variables.
\item \emph{Scope.}
  The statement says that the prior and $\lambda$ fix a token law and that exact MAP attains it at the least loss.
  Which prior gives a good training distribution is measured in \cref{sec:mechanism}.
\item \emph{Approximate minimization.}
  On planted reconstruction targets ($\lambda=0$), \cref{alg:relax} falls short of the optimal gain by
  2.5\% for LlamaGen and 3.8\% for IBQ-$2^{14}$. It recovers the full map on 57\% and 75\% of images,
  respectively.
  For $\lambda=0.01$, where no planted test exists, eight steps of a rate-aware polish (over the 16 candidates, with
  the exact bits in the proposals, kept only if the exact objective falls) leave the rate unchanged ($+0.001$ bit per
  token) and lower rFID-2k by $0.07\pm0.04$: this local search finds no rate saving near the relaxation's
  output.
  Two runs share 58\% of their tokens.
  We train on the maps returned by the relaxation. Their distribution depends weakly on the batch through
  the median gap $\bar\delta$ in \cref{alg:relax}.
\end{itemize}

\paragraph{Converse for price-based encoders.}
A converse would need every law $Q$ of $(y,k)$ that some encoder with finite expected cost outputs to be the output
law of some price-based encoder $f_\phi(x,y)=\argmin_k[c(x,k)+\phi_y(k)]$; \cref{prop:transport} would then make every
stochastic encoder dominated by a deterministic one with the same output law.
In semi-discrete optimal transport such prices exist when, for each class, $x$ has a density and the tie sets
$\{x:c(x,k)-c(x,k')=\text{const}\}$ are null for $k\ne k'$, a twist-type condition~\cite{kitagawa2019convergence}, for
finite continuous costs; with the restriction to $\mathcal K(x)$, where $c=+\infty$ outside, this is unproven.
On a finite training set $x$ has no density, so this concerns the population.
For the squared L2 cost the tie sets are hyperplanes.
For L2 + LPIPS restricted to the candidates $\mathcal K(x)$ we have not checked the condition.
The prices $\phi$ also need not come from any autoregressive prior.

\subsection{The relaxation: mean-field rate and temperature}\label{sec:supp-relax}

We use the indexing of \cref{alg:relax}. At position $t$, the candidates $k_{t1},\dots,k_{tm}$ have
codes $c_{tj}$, logits $s_t\in\R^m$ and weights $\pi_t=\softmax(s_t/\tau)$.
The soft code is $\tilde q_t(\pi)=\sum_j\pi_{tj}c_{tj}$, renormalized for unit-norm codebooks.

\paragraph{Mean-field rate.}
Let $e(k)$ be the prior's input embedding of token $k$.
The mean-field input at position $t$ is $\bar e_t(\pi)=\sum_j\pi_{tj}e(k_{tj})$.
Feeding $\bar e_1,\dots,\bar e_n$ after the class token, the causal prior outputs $p_\theta(\cdot\mid\bar e_{<t},y)$ at
position $t$.
The mean-field bits are
\begin{equation}
B(\pi)=\sum_{t=1}^n\sum_{j=1}^m\pi_{tj}\big(-\log_2p_\theta(k_{tj}\mid\bar e_{<t}(\pi),y)\big),
\label{eq:supp-meanfield}
\end{equation}
and \cref{alg:relax} optimizes $\tilde J(\pi)=\ell\big(x,D(\tilde q(\pi))\big)+\tfrac\lambda nB(\pi)$ with Adam on the logits.
\begin{itemize}\setlength\itemsep{1pt}
\item At a vertex, where each $\pi_t$ is one-hot at $k_t$, $\bar e_t=e(k_t)$ and $B=-\log_2p_\theta(k\mid y)$.
  So $\tilde J$ equals the objective of \cref{eq:map} on the vertices; it extends it to the product of simplices.
\item $B$ is differentiable in $\pi$, through the weights $\pi_{tj}$ and through the inputs $\bar e_{<t}$.
\item Off the vertices, $B(\pi)$ is not the expected code length $\E_{k\sim\pi}[-\log_2p_\theta(k\mid y)]$ of the product
  law $\pi$: the prior sees the average embedding, not an average over contexts.
  It is a surrogate that is exact at the vertices.
  At $\lambda=0.01$, a rate-aware polish after the relaxation does not lower the rate (\cref{sec:supp-transport}).
\end{itemize}

\begin{lemma}[What the temperature changes]\label{lem:supp-temperature}
Let $\Delta^\circ$ be the open simplex in $\R^m$ and $u=s/\tau$ the natural logits.
\begin{enumerate}[label=(\roman*),leftmargin=*,itemsep=1pt]
\item For every $\tau>0$, $s\mapsto\softmax(s/\tau)$ maps $\R^{n\times m}$ onto $(\Delta^\circ)^n$; two logit arrays give
  the same $\pi$ iff they differ by a constant in each row.
  So $\inf_s\tilde J(\softmax(s/\tau))=\inf_{(\Delta^\circ)^n}\tilde J$ for every $\tau$.
  The stationary points of $s\mapsto\tilde J(\softmax(s/\tau))$ are $\tau$ times those of $u\mapsto\tilde J(\softmax(u))$.
  For a given $s$, the rounding $\argmax_js_{tj}=\argmax_j\pi_{tj}$ is the same for every $\tau$.
\item At a fixed $\tau$, $\nabla_s\tilde J(\softmax(s/\tau))=\tau^{-1}\nabla_u\tilde J(\softmax(u))$.
  Gradient descent on $s$ with step $\eta$ is gradient descent on $u$ with step $\eta/\tau^2$.
  Adam on $s$ with learning rate $\eta$ and constant $\epsilon$ is Adam on $u$ with learning rate $\eta/\tau$ and constant
  $\tau\epsilon$.
\item Under the schedule $\tau_i=0.01^{i/(S-1)}$, the temperature changes between steps:
  \begin{align*}
  u_{i+1}&=\frac{\tau_i}{\tau_{i+1}}\Big(u_i+\frac{\Delta s_i}{\tau_i}\Big),\\
  \frac{\tau_i}{\tau_{i+1}}&=100^{1/(S-1)}>1.
  \end{align*}
  Besides the step $\Delta s_i$, each step multiplies the natural logits by $100^{1/(S-1)}$.
  This moves every $\pi_t$ towards the vertex $\argmax_js_{tj}$, independently of $\tilde J$.
  Adam's moment estimates also mix gradients scaled by different factors $1/\tau_i$.
\end{enumerate}
\end{lemma}
\begin{proof}
(i) For $\pi_t\in\Delta^\circ$, $s_t=\tau\log\pi_t+a\mathbf 1$ maps to $\pi_t$ for every $a\in\R$, and
$\softmax(v)=\softmax(v')$ iff $v-v'\propto\mathbf 1$.
The rest follows because $s=\tau u$ is a linear bijection, and $\argmax$ is invariant under positive scaling.
(ii) The chain rule gives the gradient.
Adam's first and second moments are linear and quadratic in the gradients.
With every gradient multiplied by $\tau^{-1}$, its update on $s$ becomes $-\eta\,\hat m_u/(\sqrt{\hat v_u}+\tau\epsilon)$,
and $\Delta u=\Delta s/\tau$.
(iii) $u_{i+1}=s_{i+1}/\tau_{i+1}=(s_i+\Delta s_i)/\tau_{i+1}$.
\end{proof}

\noindent At fixed temperature, the relaxed objective has the same infimum, and stationary points correspond
under the change of variables $u=s/\tau$. In natural logits, Adam's effective learning rate is scaled by
$1/\tau$. It rises from 0.3 to 30 over the schedule. The extra scaling in part (iii) also sharpens the
mixtures towards a vertex.
These changes affect the optimization trajectory and can change the returned map.
Stopping the annealing at $\tau\approx0.04$ keeps 99.4\% of the PSNR gain.
Unlike deterministic annealing~\cite{rose1998deterministic}, $\tilde J$ has no entropy term, so no continuation path is
followed.
$\tilde J$ need not attain its infimum at a vertex; the final $\argmax$ is a rounding step.
On a linear toy without token overlap, where mixtures decode well, the relaxation ends as much as 1.31\,dB (of the
quantization part of the loss) short of the exact optimum at $K=256$.
On released decoders and planted targets ($\lambda=0$) it falls short of the optimal gain by 2.5--3.8\%
(\cref{sec:supp-transport}).

\subsection{Splitting gFID by the triangle inequality}\label{sec:supp-triangle}

For positive semidefinite $\Sigma_1,\Sigma_2$ let
$d_B(\Sigma_1,\Sigma_2)^2=\operatorname{tr}\Sigma_1+\operatorname{tr}\Sigma_2-2\operatorname{tr}\big(\Sigma_1^{1/2}\Sigma_2\Sigma_1^{1/2}\big)^{1/2}$.
FID between Gaussian fits $(\mu_1,\Sigma_1)$ and $(\mu_2,\Sigma_2)$ is $\|\mu_1-\mu_2\|^2+d_B(\Sigma_1,\Sigma_2)^2$, the squared
2-Wasserstein distance between $\mathcal N(\mu_1,\Sigma_1)$ and
$\mathcal N(\mu_2,\Sigma_2)$~\cite{dowson1982frechet,gelbrich1990formula}.

\begin{lemma}[Bures--Wasserstein metric]\label{lem:supp-bures}
$d_B(\Sigma_1,\Sigma_2)=\min_{U\in O(p)}\|\Sigma_1^{1/2}-\Sigma_2^{1/2}U\|_F$ ($p$ the feature dimension), and $d_B$ is a metric on positive
semidefinite matrices.
Hence $\sqrt{\mathrm{FID}}$ is a metric on Gaussian fits.
\end{lemma}
\begin{proof}
Let $A_k=\Sigma_k^{1/2}$.
$\|A_1-A_2U\|_F^2=\operatorname{tr}\Sigma_1+\operatorname{tr}\Sigma_2-2\operatorname{tr}(A_1A_2U)$.
By von Neumann's trace inequality, $\max_U\operatorname{tr}(A_1A_2U)$ is the sum of the singular values of $A_1A_2$, attained
at the polar factor.
These are the square roots of the eigenvalues of $A_1A_2^2A_1=\Sigma_1^{1/2}\Sigma_2\Sigma_1^{1/2}$, which gives the first
claim.
Symmetry: $\|A_1-A_2U\|_F=\|A_2-A_1U\T\|_F$.
If $d_B=0$ then $A_1=A_2U$ and $\Sigma_1=A_1A_1\T=A_2UU\T A_2=\Sigma_2$.
Triangle: take minimizers $U$ for $(\Sigma_1,\Sigma_2)$ and $V$ for $(\Sigma_2,\Sigma_3)$.
Then
$d_B(\Sigma_1,\Sigma_3)\le\|A_1-A_3VU\|_F\le\|A_1-A_2U\|_F+\|(A_2-A_3V)U\|_F=d_B(\Sigma_1,\Sigma_2)+d_B(\Sigma_2,\Sigma_3)$.
For $\sqrt{\mathrm{FID}}$, apply Minkowski's inequality in $\R^2$ to the pairs $(\|\mu_1-\mu_2\|,d_B(\Sigma_1,\Sigma_2))$.
\end{proof}

\begin{corollary}[Reconstruction and modelling terms]\label{cor:supp-triangle}
Let $X$, $R$ and $G$ be the Inception features of the training images, of the decoded training tokens and of the samples,
and $C=\mathrm{FID}(X,R)$, $M=\mathrm{FID}(R,G)$, $\mathrm{gFID}=\mathrm{FID}(X,G)$.
Then $(\sqrt C-\sqrt M)^2\le\mathrm{gFID}\le(\sqrt C+\sqrt M)^2$.
Equivalently, for $C,M>0$, $\mathrm{gFID}=C+M+2\omega\sqrt{CM}$ for a unique $\omega\in[-1,1]$.
The bounds hold exactly for the plug-in values when the three FIDs share the moments of $X$, $R$ and $G$.
\end{corollary}
\begin{proof}
\Cref{lem:supp-bures} applied to the three Gaussian fits gives the triangle and reverse triangle inequalities for
$\sqrt{\mathrm{FID}}$.
Each plug-in FID is the distance between Gaussians with the sample moments, so the lemma applies to them directly.
\end{proof}

\paragraph{Interpreting the FID terms.}
Reducing $C$ or $M$ reduces the upper bound. The exact expression also depends on $\omega$, so the measured
gFID can respond differently.
On \INh (descriptive; three seeds per arm; $R$: all 128,135 decoded training images, view 0), $\omega$ is $-0.02$ to $-0.01$ for \NN,
0.11--0.13 for \RX, 0.09--0.11 for \MAP and 0.10--0.14 for \MAPo.
So the cross term is small relative to gFID, but its change between arms is comparable to the gains: it moves from
$-0.16$ (\NN) to $+0.71$ (\MAP).
Against \NN, \MAP lowers $C$ from 2.953 to 1.512 ($-49\%$) and $M$ from 9.256 to 8.677 ($-6.3\%$); \RX lowers $C$ as
much but $M$ by 1\% only, and KD lowers $M$ by 11\% (\NN tokens) and 22\% (\MAP tokens).
So under \MAP $C+M$ falls by 2.02 but gFID by 1.15.
On \INk at cfg~2.0, $\sqrt{\mathrm{gFID}}=2.612\le1.391+2.122$ for \NN (seed 1) and $2.435\le0.994+2.036$ for \MAP (seed 0).
\NN's $M$ is 6.63 for seed 0 at cfg~2.5 against 4.50 for seed 1 at cfg~2.0; seed and cfg are confounded here.

\section{Bias of paired FID and of gFID}\label{sec:supp-bias}

We use the FID and Bures--Wasserstein definitions of \cref{sec:supp-triangle}. The plug-in estimate
$\mathrm{FID}_N$ uses sample means and unbiased sample covariances from $N$ feature vectors per side, with
$N>p$ ($p=2048$ for Inception and 1024 for DINOv2). The population value is $\mathrm{FID}_\infty$.
The sampling errors have order $N^{-1/2}$. A second-order expansion gives the leading bias, of order $1/N$.
For rFID we account for pairing between each image and its reconstruction. For gFID we compare two designs
for sampling class labels.

\begin{lemma}[Square-root trace to second order]\label{lem:sqrt}
Let $M_0=U\operatorname{diag}(q_1^2,\dots,q_p^2)U\T$ with all $q_k>0$, and let $P$ be symmetric with entries $P_{kl}$ in the
basis $U$.
Then
\begin{align*}
\operatorname{tr}(M_0+P)^{1/2}={}&\operatorname{tr}M_0^{1/2}+\sum_k\frac{P_{kk}}{2q_k}\\
&-\frac14\sum_{k,l}\frac{P_{kl}^2}{q_kq_l(q_k+q_l)}+O(\|P\|^3).
\end{align*}
\end{lemma}
\begin{proof}
The square root is analytic near $M_0\succ0$.
Write $(M_0+P)^{1/2}=M_0^{1/2}+L+Q+O(\|P\|^3)$ with $L$ linear and $Q$ quadratic in $P$.
Squaring and matching orders gives $M_0^{1/2}L+LM_0^{1/2}=P$ and $M_0^{1/2}Q+QM_0^{1/2}=-L^2$.
In the basis $U$ these read $L_{kl}=P_{kl}/(q_k+q_l)$ and $Q_{kk}=-(L^2)_{kk}/(2q_k)$.
So $\operatorname{tr}L=\sum_kP_{kk}/(2q_k)$ and $\operatorname{tr}Q=-\sum_{k,l}P_{kl}^2/\big(2q_k(q_k+q_l)^2\big)$.
Symmetrizing the last sum in $k$ and $l$ gives the stated form.
\end{proof}

\begin{corollary}[Bures distance to second order]\label{cor:supp-bures2}
For $\Sigma\succ0$ with eigenpairs $(\nu_i,v_i)$ and symmetric $\Delta$ with $\Delta_{ij}=v_i\T\Delta v_j$,
\begin{equation}
d_B(\Sigma,\Sigma+\Delta)^2=\frac12\sum_{i,j}\frac{\Delta_{ij}^2}{\nu_i+\nu_j}+O(\|\Delta\|^3).
\label{eq:bures2}
\end{equation}
\end{corollary}
\begin{proof}
$\Sigma^{1/2}(\Sigma+\Delta)\Sigma^{1/2}=\Sigma^2+P$ with $P=\Sigma^{1/2}\Delta\Sigma^{1/2}$, so \cref{lem:sqrt} applies
with $q_i=\nu_i$ and $P_{ij}=\sqrt{\nu_i\nu_j}\,\Delta_{ij}$.
It gives $\operatorname{tr}(\Sigma^2+P)^{1/2}=\operatorname{tr}\Sigma+\tfrac12\operatorname{tr}\Delta
-\tfrac14\sum_{i,j}\Delta_{ij}^2/(\nu_i+\nu_j)+O(\|\Delta\|^3)$.
In $d_B^2=2\operatorname{tr}\Sigma+\operatorname{tr}\Delta-2\operatorname{tr}(\Sigma^2+P)^{1/2}$ the linear terms cancel.
In one dimension this is $(\sigma-\sqrt{\sigma^2+\delta})^2\approx\delta^2/(4\sigma^2)$.
\end{proof}

\noindent The quadratic form in \cref{eq:bures2} is $\tfrac12\langle\Delta,\mathcal L_\Sigma^{-1}(\Delta)\rangle$ with
$\mathcal L_\Sigma(Z)=\Sigma Z+Z\Sigma$.
It depends smoothly on $\Sigma\succ0$, with no perturbation of eigenvectors, even for repeated eigenvalues.

\begin{proposition}[Paired rFID]\label{prop:supp-paired}
Let the pairs $(r_n,\hat r_n)$, $n=1,\dots,N$, be i.i.d.\ with finite sixth moments, means $\mu=\E r$ and
$\hat\mu=\E\hat r$, and let $N>p$.
Let $\Sigma=\operatorname{Cov}r\succ0$ have eigenpairs $(\nu_i,v_i)$, and $\hat\Sigma=\operatorname{Cov}\hat r$.
Assume that the third-order remainders of the expansion in the proof are uniformly integrable, as in the delta method.
With $R_{ni}=v_i\T(r_n-\mu)$, $\hat R_{ni}=v_i\T(\hat r_n-\hat\mu)$ and
$V_{ij}=\operatorname{Var}_n(\hat R_{ni}\hat R_{nj}-R_{ni}R_{nj})$,
\begin{align}
\E\,\mathrm{FID}_N&=\mathrm{FID}_\infty+\tfrac bN+O\big(\tfrac{\|\hat\Sigma-\Sigma\|}N\big)+O(N^{-3/2}),\label{eq:bias}\\
b&=\operatorname{tr}\operatorname{Cov}_n(\hat r_n-r_n)+\frac12\sum_{i,j}\frac{V_{ij}}{\nu_i+\nu_j}.\nonumber
\end{align}
\end{proposition}
\begin{proof}
\emph{Mean term.}
$\bar{\hat r}-\bar r$ is the mean of $N$ i.i.d.\ differences, so
$\E\|\bar{\hat r}-\bar r\|^2=\|\hat\mu-\mu\|^2+\operatorname{tr}\operatorname{Cov}_n(\hat r_n-r_n)/N$ exactly.

\emph{Covariance term.}
Write $S=\Sigma+X$ and $\hat S=\hat\Sigma+Y$ for the sample covariances; $\E X=\E Y=0$.
Let $D_0=\hat\Sigma-\Sigma$, $E=Y-X$, $\Delta=\hat S-S=D_0+E$ and
$\mathcal Q_\Sigma(\Delta)=\tfrac12\langle\Delta,\mathcal L_\Sigma^{-1}(\Delta)\rangle$.
By \cref{cor:supp-bures2} at $S$ and at $\Sigma$,
$d_B(S,\hat S)^2=\mathcal Q_S(\Delta)+O(\|\Delta\|^3)$ and $d_B(\Sigma,\hat\Sigma)^2=\mathcal Q_\Sigma(D_0)+O(\|D_0\|^3)$.
Replacing $\mathcal L_S^{-1}$ by $\mathcal L_\Sigma^{-1}$ changes the form by $O(\|X\|\,\|\Delta\|^2)$.
In expectation, this change and the difference of the two third-order remainders are
$O(\|D_0\|/N)+O(N^{-3/2})$, because their parts linear in $X$ and $E$ have mean zero; as in the delta method, this
uses the sixth moments and assumes that the remainders are uniformly integrable.
Since $\E E=0$, $\E\,\mathcal Q_\Sigma(D_0+E)=\mathcal Q_\Sigma(D_0)+\tfrac12\sum_{i,j}\operatorname{Var}(E_{ij})/(\nu_i+\nu_j)$.
To first order $E_{ij}=\tfrac1N\sum_n\big(\hat R_{ni}\hat R_{nj}-R_{ni}R_{nj}-(\hat\Sigma-\Sigma)_{ij}\big)$, a mean of $N$
i.i.d.\ zero-mean terms.
Centring at the sample means changes its variance by $O(N^{-2})$.
So $\operatorname{Var}(E_{ij})=V_{ij}/N+O(N^{-2})$, and the two terms add up to $b/N$.
\end{proof}

\noindent The remainder $O(\|\hat\Sigma-\Sigma\|/N)$ is small when the reconstruction and original
feature covariances are close. We estimate all terms of $b$ from the same $N$ pairs in one pass and use
$\mathrm{FID}_\infty\approx\mathrm{FID}_N-b/N$.

\paragraph{Effect of pairing.}
$V_{ij}$ is the variance of a difference of two products on the same image.
If $\hat r_n=r_n$, both terms of $b$ vanish, and $b$ grows with the displacement $\hat r_n-r_n$.
On our arms $b$ is 430--524 times $\E_n\|\hat r_n-r_n\|^2$, so at $N=50$k the bias is about 1\% of the per-image distance.
For one-dimensional Gaussian pairs with standard deviations $\sigma,\hat\sigma$ and correlation $\varrho$, the covariance
term $(s-\hat s)^2$ exceeds $(\sigma-\hat\sigma)^2$ by $\sigma\hat\sigma(1-\varrho^2)/N+O(N^{-2})$.
This holds for any $\sigma,\hat\sigma$, and agrees with \cref{prop:supp-paired} when $\sigma=\hat\sigma$.
A Monte Carlo check at $N=100$ with $\hat\sigma=2\sigma$ and $\varrho=0$ gives $N$ times the excess as $1.988\pm0.033$, against
2.000 to first order (2.015 exactly for normal samples).
With an independent reference set, $X$ and $Y$ are independent.
The mean term becomes $\operatorname{tr}(\Sigma+\hat\Sigma)/N$, about $350/N$ for ImageNet validation features and about
ten times the paired value.
$V_{ij}$ becomes $\operatorname{Var}(\hat R_i\hat R_j)+\operatorname{Var}(R_iR_j)$, which does not shrink with fidelity.

\paragraph{Band blocks.}
The same second-order form splits the covariance term among pairs of eigenvalue-rank bands $(\mathcal I,\mathcal J)$, as
$\tfrac12\sum_{i\in\mathcal I,j\in\mathcal J}(\hat S-S)_{ij}^2/(\nu_i+\nu_j)$.
The entries $i=j$ are the part that commutes with $\Sigma$, the others a rotation.
Subtracting $V_{ij}/\big(2N(\nu_i+\nu_j)\big)$ from each entry removes its bias.
Debiased, GigaTok B-L + DINO's rotation part under our assignment roughly doubles in every band pair (second-order sum
$0.085\to0.165$), whereas IBQ's and Open-MAGVIT2's blocks fall by 5--15\% (\cref{sec:supp-recon}, ``GigaTok with a DINO discriminator'').

\paragraph{Check against subsets.}
On eleven 50k arms we also fit $\mathrm{FID}_N=\mathrm{FID}_\infty+a/N$ by least squares to strided subsets of
$N=50\text{k}/k$, $k\in\{1,2,3,4,5,6,8,10\}$, each the mean over up to four offsets and scored against the real features
of the same images~\cite{chong2020effectively}.
The closed form exceeds the fitted $a$ by 0.6--2.4\%.
The two estimates of $\mathrm{FID}_\infty$ differ by 0.008--0.022, and the change between a tokenizer's two arms by at
most 0.009.
At $N=50$k the bias of the \NN reconstructions, by the $1/N$ fit, is 0.38--0.47 on the three SEED-Voken models and
GigaTok B-L + DINO,
33--71\% of their clean-fid rFID (the closed form of \cref{prop:supp-paired} gives 33--71\% as well); for GigaTok
B-L + DINO the fit gives $\mathrm{FID}_\infty=0.166$ and a bias of
0.384 (0.550 in all, against the measured 0.544).
\Cref{tab:recon} gives the fit in the cells it marks ($^\S$) and the closed form otherwise.
TiTok's 10k evaluations use the closed form, since at $N\le10$k the residuals of the $1/N$ fit reach 0.10;
\cref{fig:cost} uses the closed form too.

\paragraph{Generation FID.}
We treat the training set moments $(\mu_r,\Sigma_r)$ as a fixed reference. The randomness comes from
$N$ generated samples $g_n$ with labels from $C$ classes.
Let $(\mu_g,\Sigma_g)$ be the generator's moments under uniform labels, $S_w$ the mean within-class covariance and $S_b$
the covariance of the class means, so $\Sigma_g=S_w+S_b$.
Let $\Sigma_r^{1/2}\Sigma_g\Sigma_r^{1/2}=U\operatorname{diag}(q^2)U\T$ with $q_k>0$, and $h=U\T\Sigma_r^{1/2}(g-\mu_g)$.

\begin{proposition}[gFID bias and the label design]\label{prop:supp-gfid}
Under the conditions of \cref{prop:supp-paired} for the samples' features (finite sixth moments, $N>p$, uniformly
integrable remainders), to order $1/N$, $\E\,\mathrm{FID}_N=\mathrm{FID}_\infty+a/N$.
With i.i.d.\ uniform labels,
\[
a_{\mathrm{iid}}=\operatorname{tr}\Sigma_g+\frac12\sum_{k,l}\frac{\operatorname{Var}(h_kh_l)}{q_kq_l(q_k+q_l)}.
\]
With $N/C$ samples per class,
\[
a_{\mathrm{bal}}=\operatorname{tr}S_w+\zeta+\frac12\sum_{k,l}\frac{\E_c\operatorname{Var}(h_kh_l\mid c)}{q_kq_l(q_k+q_l)},
\]
where $\zeta=\operatorname{tr}S_b-\sum_k(U\T\Sigma_r^{1/2}S_b\Sigma_r^{1/2}U)_{kk}/q_k$, which is zero when
$\Sigma_g=\Sigma_r$.
The label term is
\[
a_{\mathrm{iid}}-a_{\mathrm{bal}}=\operatorname{tr}S_b-\zeta+\frac12\sum_{k,l}\frac{\operatorname{Var}_c\E(h_kh_l\mid c)}
{q_kq_l(q_k+q_l)}.
\]
\end{proposition}
\begin{proof}
$\mathrm{FID}_N=\|\bar g-\mu_r\|^2+\operatorname{tr}\Sigma_r+\operatorname{tr}S_g-2\operatorname{tr}(\Sigma_r^{1/2}S_g\Sigma_r^{1/2})^{1/2}$,
where $S_g=\Sigma_g+Y$ is the sample covariance.
Apply \cref{lem:sqrt} with $M_0=\Sigma_r^{1/2}\Sigma_g\Sigma_r^{1/2}$ and $P=\Sigma_r^{1/2}Y\Sigma_r^{1/2}$.
It needs no assumption on how close $\Sigma_g$ is to $\Sigma_r$.
To first order, $P_{kl}$ is the sample mean of $h_{nk}h_{nl}$ minus its expectation.

\emph{I.i.d.\ labels.}
The samples are i.i.d., so $\E\|\bar g-\mu_r\|^2=\|\mu_g-\mu_r\|^2+\operatorname{tr}\Sigma_g/N$, $\E Y=0$, and
$\operatorname{Var}(P_{kl})=\operatorname{Var}(h_kh_l)/N$.
The quadratic term of the lemma, times $-2$, gives the sum.

\emph{Balanced labels.}
The sample mean has covariance $S_w/N$.
Since $\sum_n\E[(g_n-\bar g)(g_n-\bar g)\T]=N\Sigma_g-S_w$, the sample covariance has
$\E S_g=\Sigma_g+S_b/(N-1)$.
This mean enters through the linear terms: $\operatorname{tr}\E Y$ minus twice
$\sum_k\E P_{kk}/(2q_k)$, which is $\zeta/(N-1)$.
The fluctuations of the products are within classes only, so
$\operatorname{Var}(P_{kl})=\E_c\operatorname{Var}(h_kh_l\mid c)/N+O(N^{-2})$.

\emph{Label term.}
By the law of total variance,
$\operatorname{Var}(h_kh_l)=\E_c\operatorname{Var}(h_kh_l\mid c)+\operatorname{Var}_c\E(h_kh_l\mid c)$, and subtracting the within-class part gives the label term.
\end{proof}

\noindent For $\Sigma_g=\Sigma_r$ the i.i.d.\ form is the one-sided version of \cref{eq:bias}:
$\operatorname{tr}\Sigma_r+\tfrac12\sum_{i,j}\operatorname{Var}(G_iG_j)/(\nu_i+\nu_j)$, with $G$ the sample's
centred coordinates in $\Sigma_r$'s eigenbasis.
The label term is the multinomial noise of the class frequencies in the mean ($\operatorname{tr}S_b$) and in the second
moments (the last sum), corrected by $\zeta$, the first-order effect of balancing on the sample covariance.
Unlike paired rFID, the gFID bias depends only on how the samples spread, not on the tokens' fidelity.
On \INh the biases of the \NN, \RX and \MAP arms differ by 0.003, so the gaps between arms are the same at
$N=\infty$.

\paragraph{Empirical bias estimates.}
For the released LlamaGen GPT-B at cfg~2.0 (50k samples, 1000 classes), $a_{\mathrm{iid}}=34{,}131$ and
$a_{\mathrm{bal}}=28{,}652$.
The label term is 0.110 at 50k (0.101 for the real validation features against their own statistics), 99\% of it in the
second moments; $\operatorname{tr}S_b=71$ comes from the mean.
At cfg~2.0 the samples' total variance is 13\% below the real images', and their between-class share is 46\%,
against 39\%.
Both forms predict the excess of the mean FID of disjoint random and balanced subsets of the 50k samples ($N=25$k,
10k, 5k) over the FID-50k within 2.1--4.1\%, always slightly below.
At 10k the label term, times the finite-population factor $(50\text{k}-10\text{k})/(50\text{k}-1)=0.80$ of subsets
drawn without replacement from the balanced 50k, predicts the random-minus-balanced gap as 0.438 against 0.458
observed.
$\mathrm{FID}_\infty=5.203-28{,}652/50\text{k}=4.63$, the value in \cref{tab:gen}.
The 5.31 for i.i.d.\ labels in \cref{sec:supp-gen} is $5.203$ plus the predicted label term, not a run with i.i.d.\ labels.
With \INh's 100 classes the label term is only 0.032--0.033 at 50k: it depends on the class set.

\section{Reconstruction: further results}\label{sec:supp-recon}
We report paired reconstruction results for individual tokenizers, test whether the gains extend to other
datasets and feature spaces, and measure how reassignment affects fine texture.

\paragraph{Published reconstruction scores.}
Comparisons with published scores depend on the crop and codebook size; our paired comparisons use the same images
and pipeline for \NN and \RX.
For FVQ, our crop gives an \NN rFID of 1.15, compared with the published
0.88. An independent run reports 1.07~\cite{tang2026stablevq}.
On GigaTok B-L-2D, \RX reaches 0.72. For $16\times16$ token maps with
codebooks of at most 18 bits per token, GigaTok and FVQ report 0.86 and
0.88, respectively~\cite{xiong2025gigatok,chang2025fvq}.
WeTok reports 0.61 with a larger implicit codebook ($2^{32}$)~\cite{zhuang2025wetok}.
These comparisons use Inception features.

\begin{figure}[t]
\centering
\begin{tikzpicture}
  \node[inner sep=0, anchor=south west] (g) at (0,0) {\includegraphics[width=\linewidth]{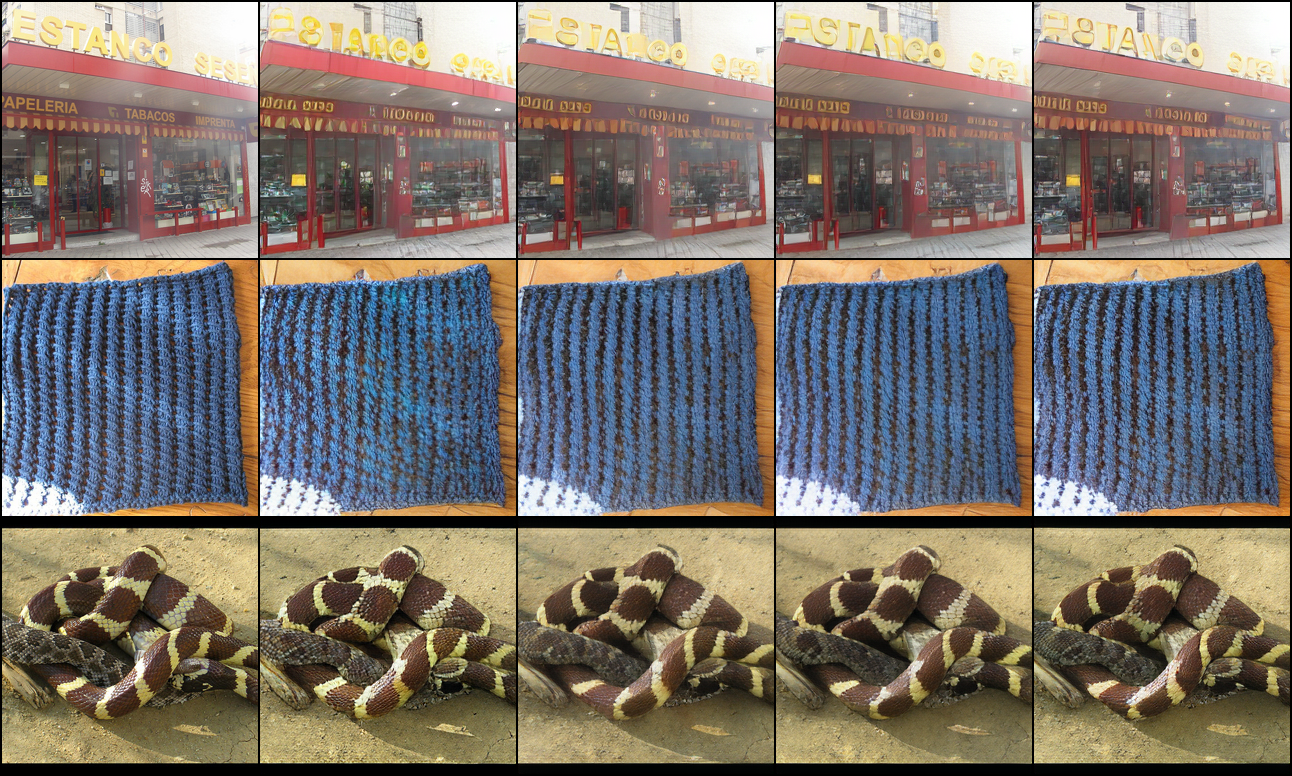}};
  \begin{scope}[x={(g.south east)}, y={(g.north west)}]
    \foreach \i/\t in {0/original, 1/{\NN}, 2/{\RXplot}, 3/{\ours}}
      \node[anchor=south, font=\scriptsize, inner sep=1.5pt, text height=1.6ex, text depth=0.4ex] at ({(130+258*\i)/1292},1) {\mbox{\t}};
    \node[anchor=south, font=\scriptsize, inner sep=1.5pt, align=center, text depth=0.4ex] at ({(130+258*4)/1292},1)
      {\mbox{\RXplot}\\+\,energy};
    \foreach \r/\a/\b/\c/\d in {0/20.84/22.19/22.18/21.80, 1/16.59/17.96/17.95/17.54, 2/18.77/20.26/20.17/19.84}
      \foreach \i/\v in {1/\a, 2/\b, 3/\c, 4/\d}
        \node[anchor=south west, font=\scriptsize, text=white, fill=black, fill opacity=0.6, text opacity=1, inner sep=1pt]
          at ({(2+258*\i)/1292+0.004}, {(518-258*\r)/776+0.008}) {\v\,dB};
  \end{scope}
\end{tikzpicture}
\caption{\textbf{Reconstructions of LlamaGen VQ-16 that differ only in the tokens:} the released \NN, \RX
  ($\lambda=0$, as in \cref{tab:recon}), \ours ($\lambda=0.01$) and \RX with a local energy term ($\gamma=0.02$,
  \cref{sec:supp-sharp}); numbers: PSNR.
  \NN's detail is sharp but partly wrong (the letters of the sign, the knit); \RX and \ours keep the structure and smooth fine texture, which
  the energy term partly restores.
  The examples have large \lpipsA gains, to make the gain clear. \cref{fig:recon-random} shows a large random
  sample.}
\label{fig:grid}
\end{figure}

\begin{figure}[t]
\centering
\begin{tikzpicture}
\begin{axis}[
  xbar, width=0.80\columnwidth, height=6.2cm,
  bar width=5pt, xmin=-56, xmax=18,
  xlabel={rFID change, \RXplot{} vs.\ \NN (\%)},
  ytick={1,...,12}, y dir=reverse, ymin=0.4, ymax=12.6,
  yticklabels={LlamaGen VQ-16, FVQ-$2^{18}$, SimVQ-$2^{10}$, SimVQ-$2^{18}$, IBQ-$2^{14}$, Open-MAGVIT2-L,
    TiTok-SL-256, GigaTok B-L-2D, IBQ-$2^{18}$, TiTok-S-128, GigaTok B-L, GigaTok B-L + DINO},
  yticklabel style={font=\scriptsize}, xticklabel style={font=\scriptsize}, xlabel style={font=\small},
  point meta=x, nodes near coords, nodes near coords align={horizontal},
  every node near coord/.append style={font=\scriptsize},
  axis x line*=bottom, axis y line*=left, axis line style={black!60, line width=0.5pt},
  tick style={black!60, line width=0.4pt}]
\draw[black!60, line width=0.5pt] (axis cs:0,0.4) -- (axis cs:0,12.6);
\addplot[fill=mapcol, draw=none, bar shift=0pt] coordinates {(-42,1) (-30,2) (-29,3) (-28,4) (-21,5) (-19,6)
  (-18,7) (-17,8) (-16,9) (-12,10) (-11,11)};
\addplot[fill=gray!60, draw=none, bar shift=0pt] coordinates {(10,12)};
\end{axis}
\end{tikzpicture}
\caption{\textbf{rFID change from NN to \RX} on the twelve tokenizers of \cref{tab:recon}, sorted (blue: a fall).
  rFID falls by 11--42\% on eleven tokenizers and rises by 10\% on GigaTok B-L with a DINO discriminator.}
\label{fig:rfid}
\end{figure}
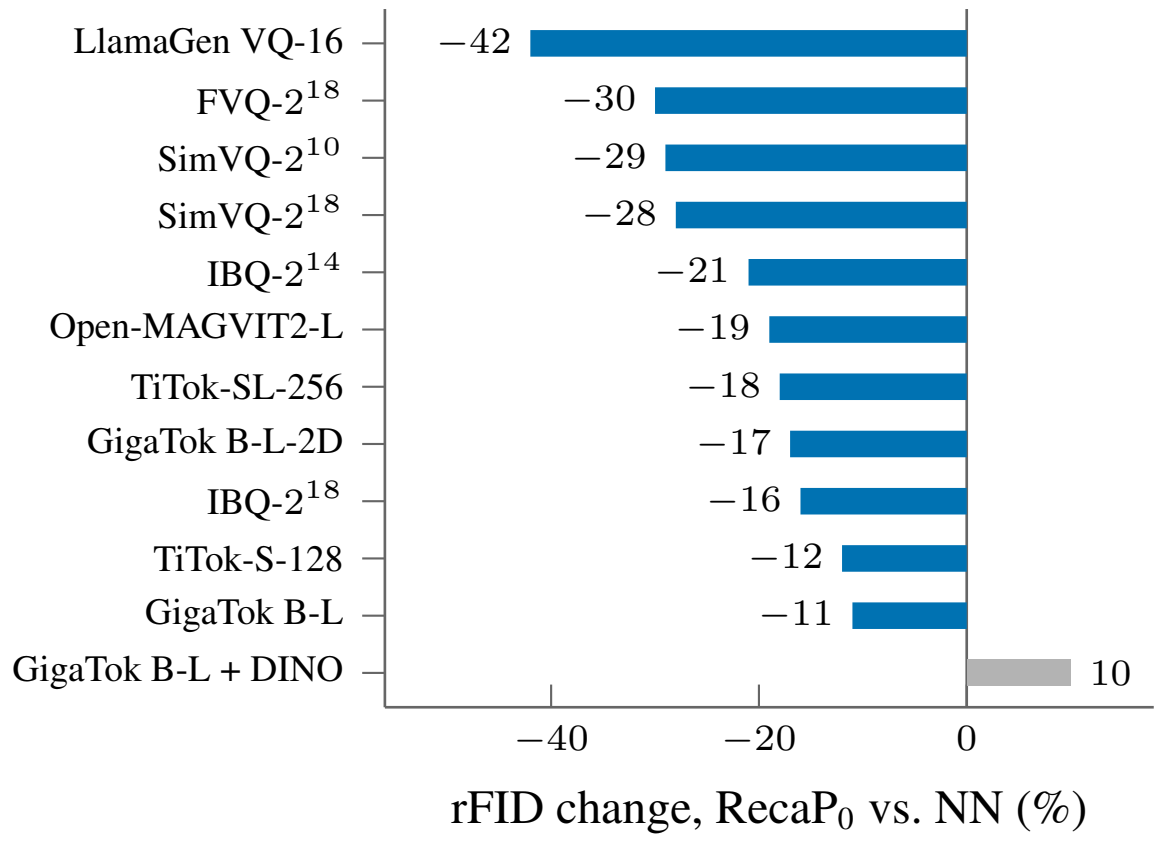

For the twelve tokenizers of \cref{tab:recon}, \cref{tab:se} gives the paired differences, \RX $-$ \NN on the same
images.
PSNR and \lpipsA: the mean of the per-image differences with its SE and 95\% interval (\lpipsA relative to NN's
mean), and the share of images that improve.
t-tests and Wilcoxon tests give $p<10^{-5}$ for every PSNR and \lpipsA difference, except the t-test on
IBQ-$2^{18}$'s \lpipsA, whose mean does not move ($p=0.50$; Wilcoxon $p<10^{-5}$, as 55.5\% of the images improve).
rFID: the difference with a delete-one jackknife SE over ten folds of whole 1000-image shards (SimVQ's runs saved
2000-image shards, each cut into its two halves), on the authors' Inception for SimVQ and the SEED-Voken models and on
clean-fid's otherwise, as in \cref{tab:recon}.
KID: the unbiased MMD$^2$ with a cubic polynomial kernel on clean-fid's Inception
features~\cite{binkowski2018demystifying}, averaged over the disjoint 1000-image shards, with the SE over the shards.
It is unbiased at every $N$ and agrees with rFID in sign on all twelve.
\Cref{fig:rfid} sorts the rFID changes.

\begin{table*}[t]
\caption{\textbf{Paired statistics of \cref{tab:recon}}: \RX $-$ \NN on the same images; $\pm$: SE, ($z$): the
  difference over its SE, images better: the share whose PSNR rises, [ ]: 95\% interval (definitions in
  \cref{sec:supp-recon}).
  Bold: the largest improvement in a column; underlined: the second largest (not marked where five rows tie).}
\label{tab:se}
\centering\footnotesize
\setlength{\tabcolsep}{4pt}
\begin{tabular}{@{}lrrrrrr@{}}
\toprule
tokenizer & $N$ & $\Delta$PSNR (dB) & images better & $\Delta$\lpipsA [95\% CI] & $\Delta$rFID $\pm$ SE ($z$) &
$\Delta$KID$\times10^3$ ($z$) \\
\midrule
LlamaGen VQ-16 & 50k & +1.269 $\pm$ 0.002 & 99.99\% & $-$5.37\% [$-$5.44, $-$5.29] & \second{$-$0.951 $\pm$ 0.091 ($-$10.4)} & \second{$-$0.954 ($-$13.4)} \\
SimVQ-$2^{10}$ & 50k & +1.222 $\pm$ 0.002 & 99.99\% & \second{$-$9.15\% [$-$9.28, $-$9.03]} & \best{$-$1.072 $\pm$ 0.077 ($-$14.0)} & \best{$-$1.303 ($-$16.6)} \\
SimVQ-$2^{18}$ & 50k & +1.042 $\pm$ 0.002 & \best{99.996\%} & \best{$-$13.11\% [$-$13.23, $-$12.98]} & $-$0.566 $\pm$ 0.039 ($-$14.3) & $-$0.684 ($-$16.7) \\
Open-MAGVIT2-L & 50k & +1.060 $\pm$ 0.001 & 99.99\% & $-$1.06\% [$-$1.14, $-$0.98] & $-$0.237 $\pm$ 0.033 ($-$7.2) & $-$0.308 ($-$11.0) \\
IBQ-$2^{14}$ & 50k & +1.166 $\pm$ 0.002 & 99.86\% & $-$0.20\% [$-$0.29, $-$0.12] & $-$0.296 $\pm$ 0.046 ($-$6.4) & $-$0.348 ($-$10.5) \\
IBQ-$2^{18}$ & 50k & +1.090 $\pm$ 0.001 & 99.99\% & $-$0.03\% [$-$0.11, +0.06] & $-$0.159 $\pm$ 0.027 ($-$5.8) & $-$0.203 ($-$10.0) \\
FVQ-$2^{18}$ & 50k & +1.136 $\pm$ 0.002 & 99.99\% & $-$3.50\% [$-$3.58, $-$3.43] & $-$0.343 $\pm$ 0.036 ($-$9.6) & $-$0.350 ($-$13.3) \\
TiTok-S-128 & 10k & \best{+1.858 $\pm$ 0.008} & 99.74\% & $-$8.73\% [$-$8.90, $-$8.57] & $-$0.505 $\pm$ 0.071 ($-$7.2) & $-$0.324 ($-$4.7) \\
TiTok-SL-256 & 10k & +0.900 $\pm$ 0.003 & 99.85\% & $-$1.69\% [$-$1.84, $-$1.55] & $-$0.508 $\pm$ 0.047 ($-$10.9) & $-$0.207 ($-$9.0) \\
GigaTok B-L-2D & 50k & +1.254 $\pm$ 0.002 & 99.98\% & $-$3.46\% [$-$3.54, $-$3.39] & $-$0.150 $\pm$ 0.011 ($-$14.1) & $-$0.093 ($-$16.4) \\
GigaTok B-L & 50k & \second{+1.288 $\pm$ 0.002} & 99.98\% & $-$5.77\% [$-$5.84, $-$5.70] & $-$0.092 $\pm$ 0.017 ($-$5.4) & $-$0.062 ($-$8.7) \\
GigaTok B-L + DINO & 50k & +1.284 $\pm$ 0.002 & 99.97\% & $-$5.02\% [$-$5.10, $-$4.95] & +0.057 $\pm$ 0.008 (+7.1) & +0.034 (+6.3) \\
\bottomrule
\end{tabular}
\end{table*}

\paragraph{Reconstruction in DINOv2 space.}
After bias correction, four of eleven tokenizers improve in DINOv2 space (\cref{tab:fd}). We decode the
registered token maps with each run's original code and preprocessing; the registered check requires every decoded
image to reproduce the run's PSNR within 0.01\,dB. IBQ-$2^{14}$ meets it only on its run's GPU model (largest
difference 0.0091\,dB) and, since its first scores had been seen, is reported descriptively, outside the registered
tests.

LlamaGen and SimVQ, decoded on another GPU model, pass the same per-image check (largest difference 0.0067\,dB).
LlamaGen's FD$_\mathrm{DINOv2}$ falls by 2.5\%, below the predicted 4--12\%.

We take every fifth image from each shard (10 per class) and use TiTok's first 10k images. Following
FD$_\mathrm{DINOv2}$~\cite{stein2023exposing}, we extract the DINOv2 ViT-L/14 CLS token after the final
norm at 224\,px, upsampling SimVQ's 128\,px images. \Cref{tab:fd} reports delete-one jackknife standard
errors over ten folds and FD$_\infty$ with the bias correction from \cref{sec:supp-bias}.

\begin{table*}[t]
\caption{\textbf{FD$_\mathrm{DINOv2}$ of the reconstructions}, 10k images per tokenizer, \RX $-$ \NN on the same
  images; ($z$): the difference over its SE.
  rFID change: the registered run's clean-fid rFID on the same tokenizer (50k images; TiTok 10k); \cref{tab:recon}
  uses the authors' Inception for SimVQ and the SEED-Voken models, so five rows differ from it.
  $^\dagger$Reported descriptively, outside the registered tests (decoded on its run's GPU model).
  Bold: the largest fall in a relative-change column; underlined: the second largest.}
\label{tab:fd}
\centering\footnotesize
\setlength{\tabcolsep}{5pt}
\begin{tabular}{@{}lrrrrr@{}}
\toprule
tokenizer & FD$_\mathrm{DINOv2}$ \NN $\to$ \RX & $\Delta$ $\pm$ SE ($z$) & change & FD$_\infty$ \NN $\to$ \RX ($z$) &
rFID change \\
\midrule
SimVQ-$2^{18}$ (128\,px) & 96.83 $\to$ 85.23 & $-$11.61 $\pm$ 0.51 ($-$22.7) & \best{$-$12.0\%} & 68.75 $\to$ 59.87 ($-$18.7) & $-$26\% \\
SimVQ-$2^{10}$ (128\,px) & 182.2 $\to$ 160.5 & $-$21.8 $\pm$ 0.8 ($-$26.5) & \second{$-$11.9\%} & 141.0 $\to$ 122.8 ($-$23.5) & $-$26\% \\
Open-MAGVIT2-L & 43.07 $\to$ 38.12 & $-$4.96 $\pm$ 0.27 ($-$18.2) & $-$11.5\% & 23.05 $\to$ 20.17 ($-$14.8) & $-$22\% \\
TiTok-S-128 & 201.5 $\to$ 190.2 & $-$11.3 $\pm$ 1.2 ($-$9.1) & $-$5.6\% & 164.3 $\to$ 156.4 ($-$6.9) & $-$12\% \\
FVQ-$2^{18}$ & 44.40 $\to$ 42.30 & $-$2.09 $\pm$ 0.34 ($-$6.1) & $-$4.7\% & 25.42 $\to$ 24.85 ($-$1.9) & \second{$-$30\%} \\
IBQ-$2^{18}$ & 33.57 $\to$ 32.12 & $-$1.45 $\pm$ 0.35 ($-$4.2) & $-$4.3\% & 15.95 $\to$ 15.88 ($-$0.2) & $-$19\% \\
LlamaGen VQ-16 & 108.9 $\to$ 106.2 & $-$2.75 $\pm$ 0.51 ($-$5.4) & $-$2.5\% & 81.45 $\to$ 81.51 (+0.1) & \best{$-$42\%} \\
IBQ-$2^{14}$$^\dagger$ & 45.32 $\to$ 42.68 & $-$2.63 $\pm$ 0.33 ($-$7.9) & $-$5.8\% & 23.80 $\to$ 23.06 ($-$2.6) & $-$23\% \\
TiTok-SL-256 & 141.9 $\to$ 142.3 & +0.4 $\pm$ 1.2 (+0.3) & +0.3\% & 119.1 $\to$ 122.4 (+2.9) & $-$18\% \\
GigaTok B-L-2D & 25.34 $\to$ 25.56 & +0.22 $\pm$ 0.23 (+1.0) & +0.9\% & 10.60 $\to$ 11.90 (+6.0) & $-$17\% \\
GigaTok B-L & 24.24 $\to$ 25.68 & +1.43 $\pm$ 0.20 (+7.2) & +5.9\% & 10.06 $\to$ 12.42 (+14.0) & $-$11\% \\
GigaTok B-L + DINO & 21.04 $\to$ 23.56 & +2.52 $\pm$ 0.18 (+14.1) & +12.0\% & 7.23 $\to$ 10.38 (+18.4) & +10\% \\
\bottomrule
\end{tabular}
\end{table*}

\paragraph{GigaTok with a DINO discriminator.}
For GigaTok B-L + DINO, rFID rises by 10\% while PSNR improves by 1.28\,dB (\cref{sec:recon}).
To examine this difference, we compare mean Inception features along the top ten principal directions
of the real features. For the six other tokenizers in this analysis (SEED-Voken, TiTok and GigaTok
B-L models), search removes 40--54\% of the shift from \NN to the originals. GigaTok B-L + DINO
starts with a smaller shift (0.018 of FID, compared with 0.057 for GigaTok B-L and 0.16--0.56 for
SEED-Voken and TiTok). GigaTok B-L uses fewer training steps (800k vs.\ 1.5M) and no DINO discriminator.
Its rFID improves by 11\%.
\paragraph{COCO and DIV2K (pre-registered).}
We compare \RX with \NN on 5,000 COCO val2017 images~\cite{lin2014coco} and 100 DIV2K
validation images~\cite{agustsson2017ntire}, using four tokenizers and their released code
(\cref{tab:recon,tab:coco}). PSNR improves on at least 99.8\% of COCO images and every DIV2K image.
The gains are 96--107\% of the corresponding ImageNet gains on COCO and 88--92\% on DIV2K.
\lpipsA improves in all four COCO runs and in three DIV2K runs, with a 3.3\% increase for IBQ.
On COCO, rFID-5k decreases in all four runs. After bias correction, FID$_\infty$ decreases for
LlamaGen and GigaTok, shows no clear change for IBQ and increases for TiTok, which has the largest
per-image gain. The per-image PSNR improvement transfers across these datasets, while the
distribution-level result varies by tokenizer.

\begin{table*}[t]
\caption{\textbf{Reconstruction-only reassignment on COCO and DIV2K (pre-registered).}
  We compare \RX ($\lambda=0$, as in \cref{tab:recon}) with \NN on the same images and using the released
  tokenizer code. The column \emph{of ImageNet's} gives each COCO PSNR gain as a fraction of the corresponding ImageNet gain.
  Each rFID-5k difference exceeds 4.9 jackknife standard errors.
  For FID$_\infty$, we report the relative change of clean-fid's value at $N\to\infty$ and the difference
  divided by its standard error ($z$).
  Bold and underlining mark the largest and second-largest improvements, including ties. rFID-5k is unranked.}
\label{tab:coco}
\centering\footnotesize
\setlength{\tabcolsep}{3.3pt}
\begin{tabular}{@{}lrrrrrrr@{}}
\toprule
 & \multicolumn{5}{c}{COCO val2017 (5,000 images)} & \multicolumn{2}{c}{DIV2K (100 images)} \\
\cmidrule(lr){2-6}\cmidrule(l){7-8}
tokenizer & $\Delta$PSNR (dB) & of ImageNet's & $\Delta$\lpipsA & rFID-5k \NN $\to$ \RX & FID$_\infty$ ($z$) &
  $\Delta$PSNR (dB) & $\Delta$\lpipsA \\
\midrule
LlamaGen VQ-16 & \second{+1.23} & \second{0.97} & \second{$-$6.0\%} & 8.22 $\to$ 6.90 & \second{$-$32.5\% ($-$6.4)} & +1.12 & $-$2.9\% \\
IBQ-$2^{14}$ & +1.12 & 0.96 & $-$1.7\% & 7.30 $\to$ 6.75 & $-$6.8\% ($-$1.0) & +1.08 & +3.3\% \\
TiTok-S-128 & \best{+1.99} & \best{1.07} & \best{$-$10.4\%} & 9.31 $\to$ 8.88 & +25.8\% (+4.5) & \best{+1.68} & \best{$-$6.0\%} \\
GigaTok B-L-2D & +1.21 & \second{0.97} & $-$4.9\% & 6.16 $\to$ 5.57 & \best{$-$36.1\% ($-$4.5)} & \second{+1.15} & \second{$-$3.4\%} \\
\bottomrule
\end{tabular}
\end{table*}

\paragraph{Per-image changes (descriptive).}
With $\lambda=0.01$, \ours gives lower PSNR than \NN on 0.044\% of the 50,000 \INk validation images.
The drop exceeds 0.1\,dB on 0.018\% of images, and \lpipsA rises on 21.4\%. The largest PSNR drops occur in
fine regular textures that are smoothed by the search (\cref{sec:supp-sharp}).

\subsection{Sharpness}\label{sec:supp-sharp}
\begin{figure*}[p]
\centering
\begin{tikzpicture}
  \node[inner sep=0, anchor=south west] (r) at (0,0) {\includegraphics[width=\textwidth]{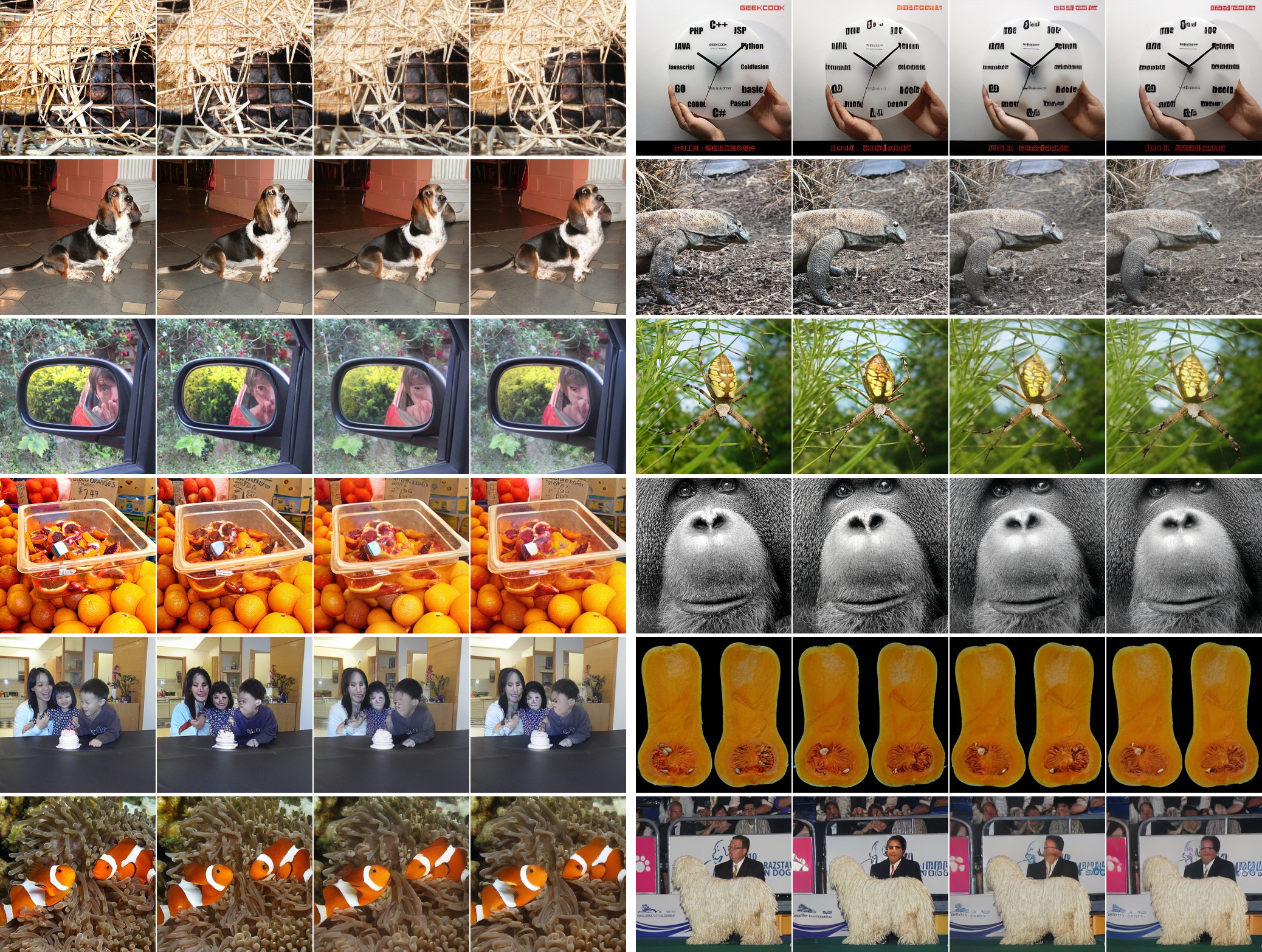}};
  \begin{scope}[x={(r.south east)}, y={(r.north west)}]
    \foreach \h in {0,1}
      \foreach \i/\t in {0/original, 1/{\NN}, 2/{\RXplot}, 3/{\ours}}
        \node[anchor=south, font=\footnotesize, inner sep=2pt, text height=1.6ex, text depth=0.4ex] at ({(\h*1046+128+258*\i)/2076}, 1) {\mbox{\t}};
  \end{scope}
\end{tikzpicture}
\caption{\textbf{Twelve random validation images} (seed 0, from the 10k of the rate experiment;
  LlamaGen VQ-16).
  Each image: the original, \NN, \RX ($\lambda=0$) and \ours ($\lambda=0.01$). \RX gains 0.8--1.9\,dB of PSNR over
  \NN on these images.}
\label{fig:recon-random}
\end{figure*}

The reconstructed images in \cref{fig:grid,fig:recon-random} appear smoother despite better PSNR, LPIPS
and rFID. We quantify this observation post hoc on 500 randomly chosen images from the 10k-image rate
experiment (\cref{sec:supp-cost}). We measure luminance energy relative to the original in two bands: the mean
squared discrete Laplacian for the finest scale and a difference of Gaussians ($\sigma=1$ and 2 pixels)
for the middle band. \Cref{tab:sharp} reports medians.
\NN's decodings keep 90\% of the finest-scale energy and those of \RX and \ours 54\%, less than \NN's on 99\% of the images; in the middle band
the shares are 99\% and 71--72\%.
The prior at $\lambda=0.01$ changes neither; at $\lambda=0.1$ the finest scale rises to 65\%, at a loss of 1\,dB.
Without the GAN term the search picks the codes whose decoding is closest to the image under L2 and LPIPS.
Where the codebook cannot reproduce a texture exactly, the closest decoding is a smoother one: the
distortion--perception trade-off~\cite{blau2018perception}.
\NN's decodings keep the decoder's sharp texture, but shapes and faces change (\cref{fig:teaser,fig:grid}).
Inception statistics favour our reconstructions (\cref{tab:recon}). Below we test a local energy term and, on
GigaTok, the tokenizer's own discriminator as perception terms in the search.

Generators trained on our tokens inherit the smoothing.
We measure the same energy on 128-px tiles of the first 64 samples of each final (\INh: \NN, \RX and \ours at cfg
2.5; \INk: \NN, \ours and the released GPT-B at cfg~2.0, with \NN's cfg-2.0 companion finals for seeds 0 and 2),
against real validation images of the same classes put through the same pipeline (post hoc).
On \INk (cfg~2.0) the samples of the GPT-B trained on \ours tokens keep 56\% of the real images' energy,
those trained on \NN tokens 90\% and the released GPT-B's 87\%; paired by class and position, \ours's sample is the
smoother one in 77\% of the pairs.
On \INh (cfg~2.5) the shares are 59\% (\ours), 56\% (\RX) and 95\% (\NN).
At cfg~1.0, 2.0 and 3.0 on \INh and cfg~2.0 on \INk at epoch~100, and at every cfg of \INk's 1.0--3.0 grid at
epoch~300, \ours's samples keep 44--62\% and \NN's 70--98\% (paired ratio 0.60--0.72; at epoch~300 on \INk, cfg~2.0: 53\% against 89\%), so longer
training does not remove the smoothing (post hoc).
gFID nevertheless falls (\cref{tab:gen}), and at a matched cfg~2.0 the \INk arms' precision and recall are level
(0.84 and 0.53 for both, rounded; no test).

\paragraph{Matching local texture energy.}
Fine-scale textures correlate weakly with the original under either assignment: median 0.10 for \NN
and 0.16 for \RX (post hoc, 500 random images from the 50k run in \cref{tab:recon}). Both band errors
exceed the original's energy in that band. Under L2, smoothing can reduce the penalty for spatially
misaligned texture. We therefore test a term that matches local texture energy.
In a pre-registered pilot on 200 validation images we add to $\ell$ the term $\gamma$ times the mean, over the token
patches, of the squared difference between the log finest-scale energy of the reconstruction and of the image
(\cref{tab:sharp-fix}).
With $\gamma=0.1$ the reconstructions keep 87\% of the original's finest-scale energy (\RX: 54\%) at 63\% of \RX's
PSNR gain, with the lowest \lpipsA of all variants.
A price per token moved off its nearest code and a larger LPIPS weight restore far less, and so does GigaTok's own
discriminator in the search (post hoc, 400 images: finest-scale energy 0.57 $\to$ 0.62 at $-0.24$\,dB on GigaTok
B-L + DINO).
On 10k validation images (pre-registered, \cref{tab:sharp-fd}) the term keeps both the texture and the
distribution-level gain.
With $\gamma=0.02$ and 0.1 the reconstructions keep 74\% and 89\% of the finest-scale energy at $+1.01$ and
$+0.79$\,dB; their clean-fid FD is at or below \RX's ($-2.2\%$, $z=-2.8$; $-1.1\%$, $z=-1.4$), and FD$_\mathrm{DINOv2}$
falls 7.3\% and 9.4\% below \RX's ($z\approx-16$), where \RX gains only 2.5\% over \NN.
For LlamaGen the texture term improves both distribution-level scores at $\gamma=0.02$ and FD$_\mathrm{DINOv2}$ at both
weights. It gives up
a fifth ($\gamma=0.02$) to two fifths ($\gamma=0.1$) of the PSNR gain.
A pre-registered generation test on \INh (GPT-B, three seeds per arm) trains on tokens from \ours's objective plus the
local energy term ($\gamma=0.02$): gFID-50k is 11.26 $\pm$ 0.03, 0.79 below \NN's (Holm~$p=0.021$) and 0.36 above
\ours's (a tie, $p=0.055$), and held-out it ties both.
Its samples keep 72\% of the real images' finest-scale energy (\ours 59\%, \NN 95\%) and have more of it than \ours's
in 58\% of matched pairs, below the registered threshold of 65\%.
In DINOv2 space they tie \ours ($-0.5\%$, $p=0.63$; \cref{tab:gfd}): the term's Inception gFID increase over \ours ($+3.3\%$)
has no DINOv2 counterpart.

\paragraph{Smoothing and the rate term (pre-registered).}
Both searches reduce the finest-scale energy by a similar amount despite different rates, consistent with
smoothing from the shared reconstruction loss.
On \INh's 5,000 validation images \RX's tokens decode as smoothly as \ours's (53\% and 54\% of the original's
finest-scale energy, \NN 92\%), yet their surprise beyond the released GPT-B's expectation ($\kappa$,
\cref{tab:kappa}) is 0.52 bit per token, against \NN's 0.22 and \ours's $-0.03$; the local energy term ($\gamma=0.02$)
restores 74\% at $\kappa=-0.01$.
Images on which \ours removes more surprise lose slightly less fine energy (partial Spearman $-0.10$ given the
original's energy; within images, over token positions, $-0.02$; both predicted small and positive).

\begin{table}[t]
\caption{\textbf{Sharpness of LlamaGen VQ-16's reconstructions} on 500 random validation images (post hoc;
  descriptive): the median energy
  of the finest scale (Laplacian) and of a middle band (difference of Gaussians), relative to the original image's;
  mean PSNR, \lpipsA and bits per token under the released GPT-B.
  \RX is \cref{tab:recon}'s run (with polish); the $\lambda$ rows are the rate
  experiment's (no polish).}
\label{tab:sharp}
\centering\footnotesize
\setlength{\tabcolsep}{3.5pt}
\begin{tabular}{@{}lrrrrr@{}}
\toprule
tokens & finest & middle & PSNR & \lpipsA & bits \\
\midrule
\NN & 0.90 & 0.99 & 20.86 & 0.137 & 10.96 \\
\RX ($\lambda=0$) & 0.54 & 0.71 & 22.12 & 0.130 & 11.50 \\
$\lambda=0.003$ & 0.54 & 0.72 & 22.12 & 0.129 & 11.22 \\
\ours ($\lambda=0.01$) & 0.54 & 0.72 & 22.07 & 0.130 & 10.72 \\
$\lambda=0.03$ & 0.56 & 0.74 & 21.82 & 0.134 & 9.91 \\
$\lambda=0.1$ & 0.65 & 0.82 & 21.09 & 0.145 & 9.18 \\
\bottomrule
\end{tabular}
\end{table}

\begin{table}[t]
\caption{\textbf{Objectives against the smoothing} on 200 LlamaGen validation images (pre-registered pilot; $\lambda=0$, no
  polish): PSNR gain over \NN, mean \lpipsA, and the median finest-scale and middle-band energy relative to the
  original (other images than \cref{tab:sharp}, hence \NN's 0.93).
  Bold: the best search objective in a column; underlined: the second best (both where tied); \NN is not ranked.}
\label{tab:sharp-fix}
\centering\scriptsize
\setlength{\tabcolsep}{2.4pt}
\begin{tabular}{@{}lrrrr@{}}
\toprule
objective & $\Delta$PSNR $\uparrow$ & \lpipsA $\downarrow$ & finest $\uparrow$ & middle $\uparrow$ \\
\midrule
\NN & n/a & 0.138 & 0.93 & 0.97 \\
\RX's objective (no polish) & \best{+1.31} & 0.130 & 0.54 & 0.71 \\
+ price per moved token, $\mu=0.1$ & +0.98 & 0.129 & 0.65 & 0.79 \\
LPIPS weight 3 (not 1) & +0.86 & 0.128 & 0.63 & 0.77 \\
+ local energy, $\gamma=0.02$ & \second{+1.07} & \second{0.126} & 0.74 & 0.80 \\
+ local energy, $\gamma=0.1$ & +0.83 & \best{0.125} & \second{0.87} & \second{0.85} \\
+ local energy, $\gamma=0.5$ & +0.52 & \second{0.126} & \best{0.96} & \best{0.90} \\
\bottomrule
\end{tabular}
\end{table}

\begin{table}[t]
\caption{\textbf{The local energy term} on 10k LlamaGen validation images (pre-registered; its rows: $\lambda=0$, no polish;
  \RX: \cref{tab:recon}'s run, with polish): PSNR gain over \NN, mean \lpipsA, median finest-scale energy relative to the original, and
  the FD of clean-fid Inception features and FD$_\mathrm{DINOv2}$ against the originals of the same images; energy:
  the local energy term.
  Bold: the best search in a column; underlined: the second best; \NN is not ranked.}
\label{tab:sharp-fd}
\centering\scriptsize
\setlength{\tabcolsep}{2.4pt}
\begin{tabular}{@{}lrrrrr@{}}
\toprule
tokens & $\Delta$PSNR $\uparrow$ & \lpipsA $\downarrow$ & finest $\uparrow$ & FD $\downarrow$ & FD$_\mathrm{DINOv2}$ $\downarrow$ \\
\midrule
\NN & n/a & 0.138 & 0.92 & 4.32 & 108.9 \\
\RX & \best{+1.27} & 0.131 & 0.54 & 3.07 & 106.2 \\
+ energy, $\gamma=0.02$ & \second{+1.01} & \second{0.126} & \second{0.74} & \best{3.00} & \second{98.5} \\
+ energy, $\gamma=0.1$ & +0.79 & \best{0.125} & \best{0.89} & \second{3.04} & \best{96.2} \\
\bottomrule
\end{tabular}
\end{table}

\subsection{Qualitative comparisons across tokenizers}\label{sec:supp-every-tok}
\begin{figure*}[p]
\centering
\begin{tikzpicture}
  \node[inner sep=0, anchor=south west] (t) at (0,0)
    {\includegraphics[width=0.94\textwidth]{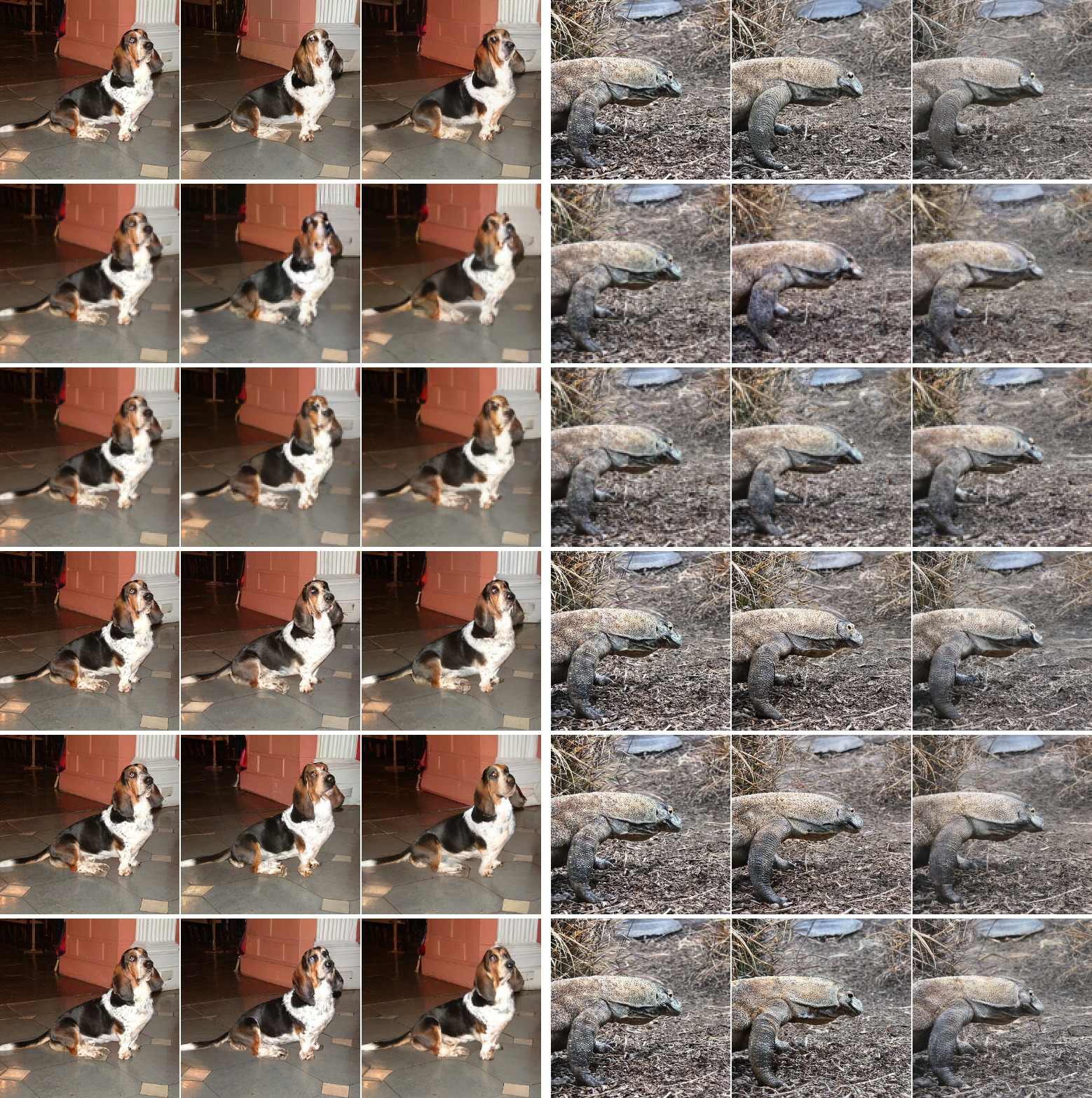}};
  \begin{scope}[x={(t.south east)}, y={(t.north west)}]
    \node[anchor=south west, font=\scriptsize, text=white, fill=black, fill opacity=0.6, text opacity=1, inner sep=1pt] at (0.1669, 0.8425) {21.89\,dB};
    \node[anchor=south west, font=\scriptsize, text=white, fill=black, fill opacity=0.6, text opacity=1, inner sep=1pt] at (0.3325, 0.8425) {22.94\,dB};
    \node[anchor=south west, font=\scriptsize, text=white, fill=black, fill opacity=0.6, text opacity=1, inner sep=1pt] at (0.6714, 0.8425) {13.94\,dB};
    \node[anchor=south west, font=\scriptsize, text=white, fill=black, fill opacity=0.6, text opacity=1, inner sep=1pt] at (0.8370, 0.8425) {15.16\,dB};
    \node[anchor=south, rotate=90, font=\scriptsize, inner sep=3pt] at (0, 0.9183) {LlamaGen VQ-16};
    \node[anchor=south west, font=\scriptsize, text=white, fill=black, fill opacity=0.6, text opacity=1, inner sep=1pt] at (0.1669, 0.6752) {24.00\,dB};
    \node[anchor=south west, font=\scriptsize, text=white, fill=black, fill opacity=0.6, text opacity=1, inner sep=1pt] at (0.3325, 0.6752) {24.79\,dB};
    \node[anchor=south west, font=\scriptsize, text=white, fill=black, fill opacity=0.6, text opacity=1, inner sep=1pt] at (0.6714, 0.6752) {18.24\,dB};
    \node[anchor=south west, font=\scriptsize, text=white, fill=black, fill opacity=0.6, text opacity=1, inner sep=1pt] at (0.8370, 0.6752) {19.12\,dB};
    \node[anchor=south, rotate=90, font=\scriptsize, inner sep=3pt] at (0, 0.7510) {SimVQ-$2^{10}$};
    \node[anchor=south west, font=\scriptsize, text=white, fill=black, fill opacity=0.6, text opacity=1, inner sep=1pt] at (0.1669, 0.5079) {26.88\,dB};
    \node[anchor=south west, font=\scriptsize, text=white, fill=black, fill opacity=0.6, text opacity=1, inner sep=1pt] at (0.3325, 0.5079) {27.67\,dB};
    \node[anchor=south west, font=\scriptsize, text=white, fill=black, fill opacity=0.6, text opacity=1, inner sep=1pt] at (0.6714, 0.5079) {19.40\,dB};
    \node[anchor=south west, font=\scriptsize, text=white, fill=black, fill opacity=0.6, text opacity=1, inner sep=1pt] at (0.8370, 0.5079) {19.93\,dB};
    \node[anchor=south, rotate=90, font=\scriptsize, inner sep=3pt] at (0, 0.5837) {SimVQ-$2^{18}$};
    \node[anchor=south west, font=\scriptsize, text=white, fill=black, fill opacity=0.6, text opacity=1, inner sep=1pt] at (0.1669, 0.3406) {22.47\,dB};
    \node[anchor=south west, font=\scriptsize, text=white, fill=black, fill opacity=0.6, text opacity=1, inner sep=1pt] at (0.3325, 0.3406) {23.52\,dB};
    \node[anchor=south west, font=\scriptsize, text=white, fill=black, fill opacity=0.6, text opacity=1, inner sep=1pt] at (0.6714, 0.3406) {13.79\,dB};
    \node[anchor=south west, font=\scriptsize, text=white, fill=black, fill opacity=0.6, text opacity=1, inner sep=1pt] at (0.8370, 0.3406) {14.80\,dB};
    \node[anchor=south, rotate=90, font=\scriptsize, inner sep=3pt] at (0, 0.4163) {Open-MAGVIT2-L};
    \node[anchor=south west, font=\scriptsize, text=white, fill=black, fill opacity=0.6, text opacity=1, inner sep=1pt] at (0.1669, 0.1733) {22.26\,dB};
    \node[anchor=south west, font=\scriptsize, text=white, fill=black, fill opacity=0.6, text opacity=1, inner sep=1pt] at (0.3325, 0.1733) {23.25\,dB};
    \node[anchor=south west, font=\scriptsize, text=white, fill=black, fill opacity=0.6, text opacity=1, inner sep=1pt] at (0.6714, 0.1733) {13.64\,dB};
    \node[anchor=south west, font=\scriptsize, text=white, fill=black, fill opacity=0.6, text opacity=1, inner sep=1pt] at (0.8370, 0.1733) {14.67\,dB};
    \node[anchor=south, rotate=90, font=\scriptsize, inner sep=3pt] at (0, 0.2490) {IBQ-$2^{14}$};
    \node[anchor=south west, font=\scriptsize, text=white, fill=black, fill opacity=0.6, text opacity=1, inner sep=1pt] at (0.1669, 0.0060) {23.18\,dB};
    \node[anchor=south west, font=\scriptsize, text=white, fill=black, fill opacity=0.6, text opacity=1, inner sep=1pt] at (0.3325, 0.0060) {23.92\,dB};
    \node[anchor=south west, font=\scriptsize, text=white, fill=black, fill opacity=0.6, text opacity=1, inner sep=1pt] at (0.6714, 0.0060) {13.79\,dB};
    \node[anchor=south west, font=\scriptsize, text=white, fill=black, fill opacity=0.6, text opacity=1, inner sep=1pt] at (0.8370, 0.0060) {14.87\,dB};
    \node[anchor=south, rotate=90, font=\scriptsize, inner sep=3pt] at (0, 0.0817) {IBQ-$2^{18}$};
    \node[anchor=south, font=\scriptsize, inner sep=2pt, text height=1.6ex, text depth=0.4ex] at (0.0822, 1) {original};
    \node[anchor=south, font=\scriptsize, inner sep=2pt, text height=1.6ex, text depth=0.4ex] at (0.2478, 1) {{\NN}};
    \node[anchor=south, font=\scriptsize, inner sep=2pt, text height=1.6ex, text depth=0.4ex] at (0.4134, 1) {{\RXplot}};
    \node[anchor=south, font=\scriptsize, inner sep=2pt, text height=1.6ex, text depth=0.4ex] at (0.5866, 1) {original};
    \node[anchor=south, font=\scriptsize, inner sep=2pt, text height=1.6ex, text depth=0.4ex] at (0.7522, 1) {{\NN}};
    \node[anchor=south, font=\scriptsize, inner sep=2pt, text height=1.6ex, text depth=0.4ex] at (0.9178, 1) {{\RXplot}};
  \end{scope}
\end{tikzpicture}
\caption{\textbf{Two validation images through the first six tokenizers} of \cref{tab:recon} (8170 and 2425, from
  \cref{fig:recon-random}): per image the original in the tokenizer's own preprocessing, \NN and \RX
  ($\lambda=0$, the runs of \cref{tab:recon}). Numbers: PSNR. SimVQ works at 128\,px and is shown with its own
  pixels.}
\label{fig:recon-tok-a}
\end{figure*}

\begin{figure*}[p]
\centering
\begin{tikzpicture}
  \node[inner sep=0, anchor=south west] (t) at (0,0)
    {\includegraphics[width=0.94\textwidth]{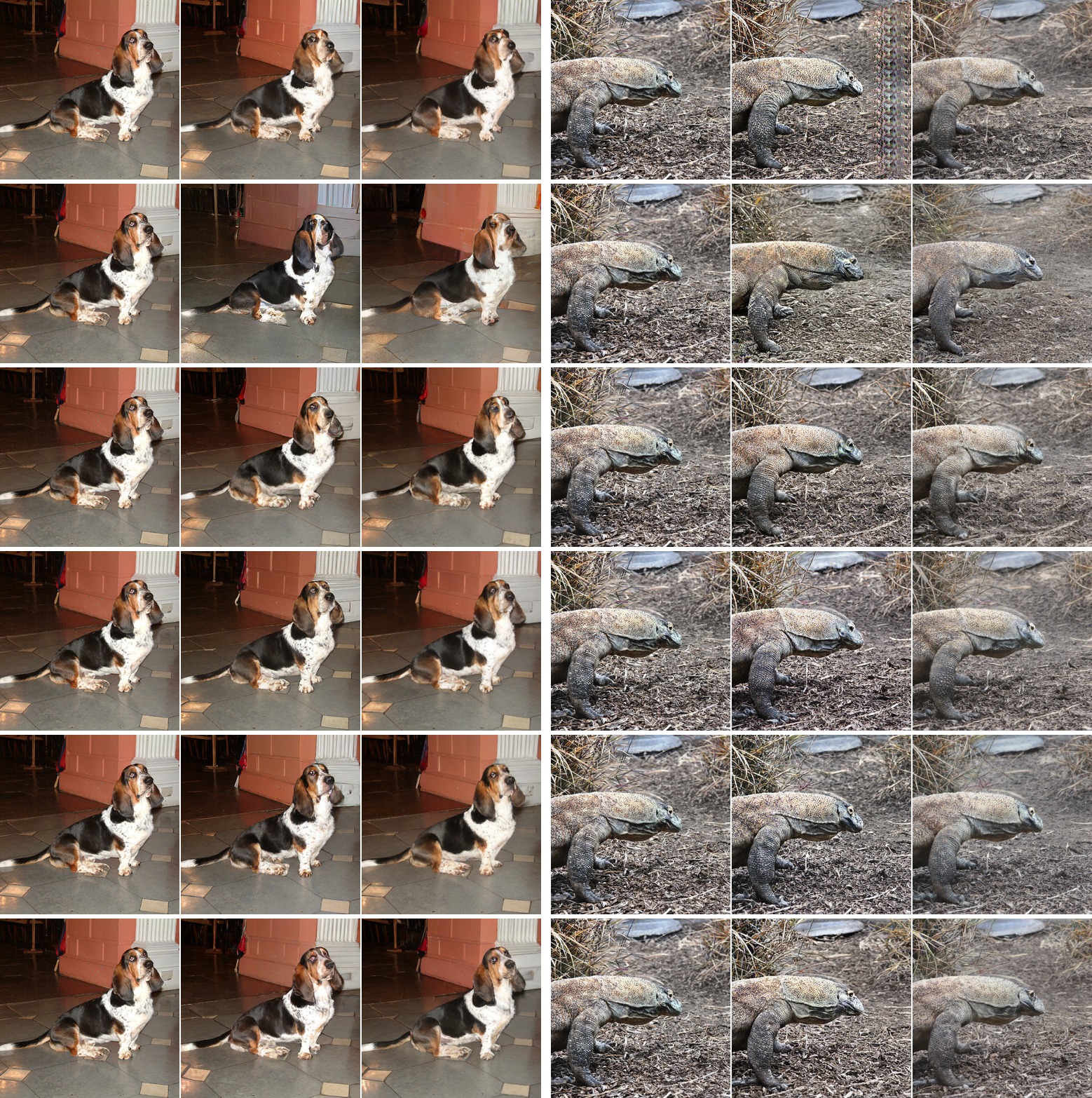}};
  \begin{scope}[x={(t.south east)}, y={(t.north west)}]
    \node[anchor=south west, font=\scriptsize, text=white, fill=black, fill opacity=0.6, text opacity=1, inner sep=1pt] at (0.1669, 0.8425) {22.95\,dB};
    \node[anchor=south west, font=\scriptsize, text=white, fill=black, fill opacity=0.6, text opacity=1, inner sep=1pt] at (0.3325, 0.8425) {23.94\,dB};
    \node[anchor=south west, font=\scriptsize, text=white, fill=black, fill opacity=0.6, text opacity=1, inner sep=1pt] at (0.6714, 0.8425) {13.46\,dB};
    \node[anchor=south west, font=\scriptsize, text=white, fill=black, fill opacity=0.6, text opacity=1, inner sep=1pt] at (0.8370, 0.8425) {15.27\,dB};
    \node[anchor=south, rotate=90, font=\scriptsize, inner sep=3pt] at (0, 0.9183) {FVQ-$2^{18}$};
    \node[anchor=south west, font=\scriptsize, text=white, fill=black, fill opacity=0.6, text opacity=1, inner sep=1pt] at (0.1669, 0.6752) {18.02\,dB};
    \node[anchor=south west, font=\scriptsize, text=white, fill=black, fill opacity=0.6, text opacity=1, inner sep=1pt] at (0.3325, 0.6752) {20.22\,dB};
    \node[anchor=south west, font=\scriptsize, text=white, fill=black, fill opacity=0.6, text opacity=1, inner sep=1pt] at (0.6714, 0.6752) {12.88\,dB};
    \node[anchor=south west, font=\scriptsize, text=white, fill=black, fill opacity=0.6, text opacity=1, inner sep=1pt] at (0.8370, 0.6752) {14.19\,dB};
    \node[anchor=south, rotate=90, font=\scriptsize, inner sep=3pt] at (0, 0.7510) {TiTok-S-128};
    \node[anchor=south west, font=\scriptsize, text=white, fill=black, fill opacity=0.6, text opacity=1, inner sep=1pt] at (0.1669, 0.5079) {22.50\,dB};
    \node[anchor=south west, font=\scriptsize, text=white, fill=black, fill opacity=0.6, text opacity=1, inner sep=1pt] at (0.3325, 0.5079) {23.04\,dB};
    \node[anchor=south west, font=\scriptsize, text=white, fill=black, fill opacity=0.6, text opacity=1, inner sep=1pt] at (0.6714, 0.5079) {13.76\,dB};
    \node[anchor=south west, font=\scriptsize, text=white, fill=black, fill opacity=0.6, text opacity=1, inner sep=1pt] at (0.8370, 0.5079) {15.02\,dB};
    \node[anchor=south, rotate=90, font=\scriptsize, inner sep=3pt] at (0, 0.5837) {TiTok-SL-256};
    \node[anchor=south west, font=\scriptsize, text=white, fill=black, fill opacity=0.6, text opacity=1, inner sep=1pt] at (0.1669, 0.3406) {22.16\,dB};
    \node[anchor=south west, font=\scriptsize, text=white, fill=black, fill opacity=0.6, text opacity=1, inner sep=1pt] at (0.3325, 0.3406) {23.72\,dB};
    \node[anchor=south west, font=\scriptsize, text=white, fill=black, fill opacity=0.6, text opacity=1, inner sep=1pt] at (0.6714, 0.3406) {13.66\,dB};
    \node[anchor=south west, font=\scriptsize, text=white, fill=black, fill opacity=0.6, text opacity=1, inner sep=1pt] at (0.8370, 0.3406) {15.17\,dB};
    \node[anchor=south, rotate=90, font=\scriptsize, inner sep=3pt] at (0, 0.4163) {GigaTok B-L-2D};
    \node[anchor=south west, font=\scriptsize, text=white, fill=black, fill opacity=0.6, text opacity=1, inner sep=1pt] at (0.1669, 0.1733) {22.12\,dB};
    \node[anchor=south west, font=\scriptsize, text=white, fill=black, fill opacity=0.6, text opacity=1, inner sep=1pt] at (0.3325, 0.1733) {23.57\,dB};
    \node[anchor=south west, font=\scriptsize, text=white, fill=black, fill opacity=0.6, text opacity=1, inner sep=1pt] at (0.6714, 0.1733) {13.55\,dB};
    \node[anchor=south west, font=\scriptsize, text=white, fill=black, fill opacity=0.6, text opacity=1, inner sep=1pt] at (0.8370, 0.1733) {15.20\,dB};
    \node[anchor=south, rotate=90, font=\scriptsize, inner sep=3pt] at (0, 0.2490) {GigaTok B-L};
    \node[anchor=south west, font=\scriptsize, text=white, fill=black, fill opacity=0.6, text opacity=1, inner sep=1pt] at (0.1669, 0.0060) {22.08\,dB};
    \node[anchor=south west, font=\scriptsize, text=white, fill=black, fill opacity=0.6, text opacity=1, inner sep=1pt] at (0.3325, 0.0060) {23.36\,dB};
    \node[anchor=south west, font=\scriptsize, text=white, fill=black, fill opacity=0.6, text opacity=1, inner sep=1pt] at (0.6714, 0.0060) {13.44\,dB};
    \node[anchor=south west, font=\scriptsize, text=white, fill=black, fill opacity=0.6, text opacity=1, inner sep=1pt] at (0.8370, 0.0060) {15.08\,dB};
    \node[anchor=south, rotate=90, font=\scriptsize, inner sep=3pt] at (0, 0.0817) {GigaTok B-L + DINO};
    \node[anchor=south, font=\scriptsize, inner sep=2pt, text height=1.6ex, text depth=0.4ex] at (0.0822, 1) {original};
    \node[anchor=south, font=\scriptsize, inner sep=2pt, text height=1.6ex, text depth=0.4ex] at (0.2478, 1) {{\NN}};
    \node[anchor=south, font=\scriptsize, inner sep=2pt, text height=1.6ex, text depth=0.4ex] at (0.4134, 1) {{\RXplot}};
    \node[anchor=south, font=\scriptsize, inner sep=2pt, text height=1.6ex, text depth=0.4ex] at (0.5866, 1) {original};
    \node[anchor=south, font=\scriptsize, inner sep=2pt, text height=1.6ex, text depth=0.4ex] at (0.7522, 1) {{\NN}};
    \node[anchor=south, font=\scriptsize, inner sep=2pt, text height=1.6ex, text depth=0.4ex] at (0.9178, 1) {{\RXplot}};
  \end{scope}
\end{tikzpicture}
\caption{\textbf{The other six tokenizers} of \cref{tab:recon}, as in \cref{fig:recon-tok-a}.}
\label{fig:recon-tok-b}
\end{figure*}

\Cref{fig:recon-tok-a,fig:recon-tok-b} show all twelve tokenizers in \cref{tab:recon} on the first
two images from \cref{fig:recon-random} covered by every run. TiTok uses the first 10k validation
images. We decode the stored maps with each tokenizer's code and preprocessing and reproduce PSNR
within 0.001\,dB. The examples show the same trade-off as \cref{fig:grid}. \NN retains fine texture
but can change shapes (TiTok-S-128 changes the dog's head and FVQ adds texture beside the lizard).
\RX preserves structure more faithfully and smooths fine texture.

\subsection{Where the tokens change}\label{sec:supp-token-map}
\begin{figure*}[p]
\centering
\begin{tikzpicture}
  \node[inner sep=0, anchor=south west] (t) at (0,0) {\includegraphics[width=\textwidth]{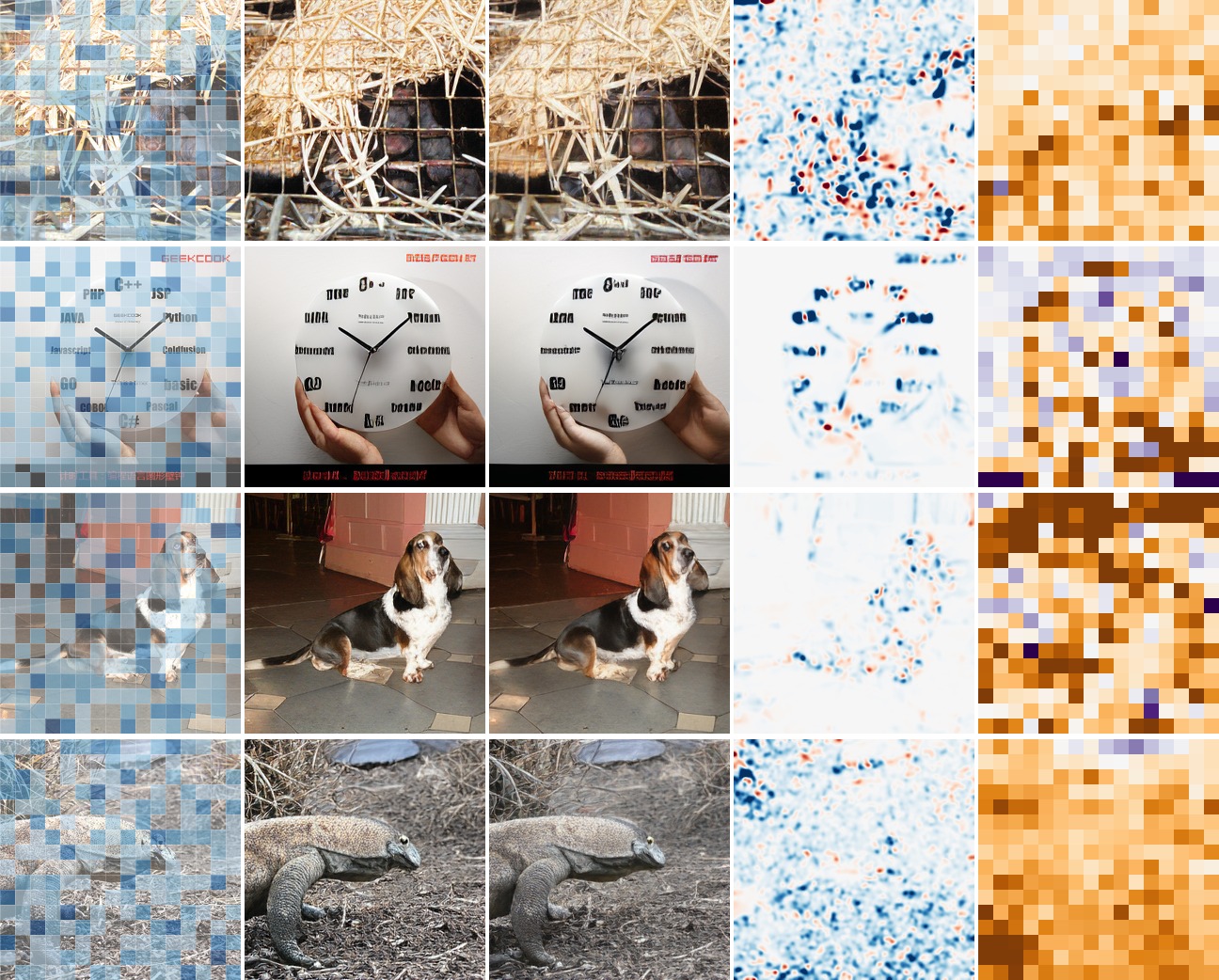}};
  \begin{scope}[x={(t.south east)}, y={(t.north west)}]
    \node[anchor=south, font=\scriptsize, inner sep=2pt, text height=1.6ex, text depth=0.4ex] at (0.0988, 1) {changed tokens};
    \node[anchor=south, font=\scriptsize, inner sep=2pt, text height=1.6ex, text depth=0.4ex] at (0.2994, 1) {{\NN}};
    \node[anchor=south, font=\scriptsize, inner sep=2pt, text height=1.6ex, text depth=0.4ex] at (0.5000, 1) {{\RXplot}};
    \node[anchor=south, font=\scriptsize, inner sep=2pt, text height=1.6ex, text depth=0.4ex] at (0.7006, 1) {error change};
    \node[anchor=south, font=\scriptsize, inner sep=2pt, text height=1.6ex, text depth=0.4ex] at (0.9012, 1) {energy change};
    \node[anchor=north, inner sep=0] (b0) at (0.0988, -0.012) {\includegraphics[width=0.1817\textwidth, height=0.18cm]{cbar_rank.png}};
    \node[anchor=north west, font=\tiny, inner sep=1pt] at (b0.south west) {2nd nearest};
    \node[anchor=north east, font=\tiny, inner sep=1pt] at (b0.south east) {16th nearest};
    \node[anchor=north, inner sep=0] (b3) at (0.7006, -0.012) {\includegraphics[width=0.1817\textwidth, height=0.18cm]{cbar_rdbu.png}};
    \node[anchor=north west, font=\tiny, inner sep=1pt] at (b3.south west) {\mbox{\RXplot} farther};
    \node[anchor=north east, font=\tiny, inner sep=1pt] at (b3.south east) {\mbox{\RXplot} closer};
    \node[anchor=north, inner sep=0] (b4) at (0.9012, -0.012) {\includegraphics[width=0.1817\textwidth, height=0.18cm]{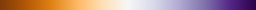}};
    \node[anchor=north west, font=\tiny, inner sep=1pt] at (b4.south west) {$-100\%$ (smoother)};
    \node[anchor=north east, font=\tiny, inner sep=1pt] at (b4.south east) {$+100\%$};
  \end{scope}
\end{tikzpicture}
\caption{\textbf{Where \RX changes the tokens} (LlamaGen VQ-16; four of the random images of \cref{fig:recon-random}).
  Columns: the changed tokens over the image (shade: the rank of the chosen code among the token's 16 nearest, a
  farther code from the polish shaded as the 16th; clear: \NN's code kept), \NN, \RX, the change of the squared error (blue: \RX closer to the image; smoothed with a
  2-pixel Gaussian) and the change of each token patch's finest-scale energy, relative to the original patch's
  (orange: smoother).}
\label{fig:token-map}
\end{figure*}

\Cref{fig:token-map} shows where search changes the codes for LlamaGen VQ-16 on four of the images
in \cref{fig:recon-random}. Across 504 validation images (500 random and the four shown), \RX changes
64\% of the tokens and \ours changes 63\%. The second-nearest code is selected at 17\% of positions.
At 10\%, the chosen code is ranked ninth to sixteenth, or farther away after \RX's refinement.

Pixel error decreases in patches at both changed and unchanged positions, consistent with the decoder
mixing neighbouring tokens. Patches at changed positions account for 68\% of the total reduction, and the
error decreases in 76\% of them. Median finest-scale energy also falls in patches at changed and
unchanged positions, by 30\% and 25\%, respectively.
In \cref{fig:token-map-main} the four close-ups are the 64\,px windows with the most changed tokens, and the
squared-error change is smoothed with a 2-pixel Gaussian.

\clearpage
\section{Implementation details}\label{sec:supp-impl}
We use the released code and freeze all tokenizer weights. We load EMA weights where the checkpoint
provides them, including SimVQ and SEED-Voken. LlamaGen, TiTok, GigaTok B-L-2D and GigaTok B-L + DINO
have no EMA weights. For FVQ we load the model weights.
\Cref{tab:impl} lists what differs between tokenizers. Open-MAGVIT2 and IBQ use
the SEED-Voken code base, and FVQ uses LlamaGen's crop.

\begin{table*}[t]
\caption{\textbf{Per-tokenizer settings of \cref{tab:recon}} (descriptive). LPIPS is LPIPS-VGG with weight 1. \emph{authors'}: their
  pytorch-fid Inception and code. Times are GPU seconds per image, including the NN pass and all metrics, on a GH200 (LlamaGen, SimVQ) or an
  RTX 4090 (the others; 8 of the 50 shards of FVQ and GigaTok B-L-2D on other GPUs).}
\label{tab:impl}
\centering\small
\setlength{\tabcolsep}{4pt}
\begin{tabular}{@{}>{\raggedright\arraybackslash}p{2.8cm}>{\raggedright\arraybackslash}p{3.6cm}>{\raggedright\arraybackslash}p{2.0cm}>{\raggedright\arraybackslash}p{3.2cm}>{\raggedright\arraybackslash}p{1.3cm}>{\raggedright\arraybackslash}p{1.0cm}>{\raggedright\arraybackslash}p{1.5cm}@{}}
\toprule
tokenizer & preprocessing & $\ell$ & codes, candidates & rFID & batch & s/image \\
\midrule
LlamaGen VQ-16 & ADM resize, centre crop 256 & L2 + LPIPS & unit norm; 16 nearest & clean-fid & 32 & 0.55 \\
SimVQ-$2^{10}$ / $2^{18}$ & resize, centre crop 128 & L1 + LPIPS & projected codes; 16 nearest & authors' & 64 / 32 & 0.13 / 0.12 \\
Open-MAGVIT2-L & shorter side 256 (bilinear), centre crop & L1 + LPIPS & LFQ signs; 16 nearest & authors' & 8 & 1.71 \\
IBQ-$2^{14}$ / $2^{18}$ & as Open-MAGVIT2 & L1 + LPIPS & 16 largest $\langle z,c\rangle$ & authors' & 8 & 1.98 / 2.05 \\
TiTok-S-128 / SL-256 & bicubic resize, centre crop 256 & L2 + LPIPS & unit norm / Euclidean; 16 nearest & clean-fid & 8 & 0.97 / 1.73 \\
GigaTok B-L / + DINO & LlamaGen's centre crop 256 & L2 + LPIPS & unit norm; 16 nearest & clean-fid & 8 & 2.12 / 2.48 \\
FVQ / GigaTok B-L-2D & LlamaGen's centre crop 256 & L2 + LPIPS & Euclidean / unit norm; 16 nearest & clean-fid & 8 & 1.60 / 2.51 \\
\bottomrule
\end{tabular}
\end{table*}

\begin{algorithm}[t]
\caption{\ours: annealed softmax relaxation}\label{alg:relax}
\DontPrintSemicolon
\SetKwInput{KwSet}{Settings}
\KwIn{image $x$, class $y$; frozen $E,\mathcal C,D$ and prior $p_\theta$}
\KwSet{weight $\lambda$, $m=16$ candidates, $S=50$ steps}
\KwOut{token map $k^\star$}
$z\gets E(x)$\;
\ForEach{token $t$}{
  $k_{t1},\dots,k_{tm}\gets$ the $m$ codes nearest to $z_t$;\ \ $c_{tj}\gets c_{k_{tj}}$\;
  $\delta_{tj}\gets\|z_t-c_{tj}\|^2$\;
}
$\bar\delta\gets\operatorname{median}_t\,(\delta_{t2}-\delta_{t1})$\Comment*[r]{over the batch's tokens}
$s_{tj}\gets-(\delta_{tj}-\delta_{t1})/\bar\delta$\Comment*[r]{argmax at NN}
\For{$i=0,\dots,S-1$}{
  $\tau\gets0.01^{\,i/(S-1)}$;\ \ $\pi_t\gets\softmax(s_t/\tau)$\Comment*[r]{$\tau$: $1\to0.01$}
  $\tilde q_t\gets\sum_j\pi_{tj}c_{tj}$\Comment*[r]{soft code}
  $\bar e_t\gets\sum_j\pi_{tj}\,e(k_{tj})$\Comment*[r]{mean-field prior input}
  $b\gets\sum_t\E_{\pi_t}\big[-\log_2p_\theta(k_t\mid\bar e_{<t},y)\big]$\Comment*[r]{bits}
  $J\gets\ell\big(x,D(\tilde q)\big)+\frac{\lambda}{n}\,b$\;
  $s\gets\textsc{Adam}(s,\nabla_sJ)$\Comment*[r]{learning rate 0.3}
}
\Return $k^\star_t=k_{t\,j_t}$ with $j_t=\argmax_js_{tj}$\;
\end{algorithm}

\paragraph{Assignment.}
\Cref{alg:relax} implements the search in \cref{sec:relax}. It uses the 16 nearest codes per
position, the 16 largest logits $\langle z_t,c\rangle$ for IBQ, or the 16 nearest sign patterns
in Open-MAGVIT2's LFQ grid. We initialize the logits to
$-(\delta_{tj}-\delta_{t1})/\bar\delta$, where $\bar\delta$ is the median of
$\delta_{t2}-\delta_{t1}$ across all tokens in the batch. For IBQ,
$\delta_{tj}=-\langle z_t,c_{tj}\rangle$ for the 16 largest logits.

The batch size affects $\bar\delta$, so changing the batch size can slightly change the selected tokens.
For example, the ten-crop \INh runs used one to three images per batch (10 to 30 crops), depending
on the GPU model.
Adam updates the logits for 50 steps with $\tau=0.01^{\,i/49}$. We use a learning rate of 0.3
and PyTorch's defaults $\beta_1=0.9$, $\beta_2=0.999$, and $\epsilon=10^{-8}$.
The soft code $\sum_j\pi_{tj}c_{tj}$ is renormalized for unit-norm codebooks.
The objective sums per-image losses over the batch. Each loss uses mean squared or absolute pixel
error plus LPIPS-VGG with weight 1. We replace TiTok's stage-2 ConvNeXt term with LPIPS-VGG. We use
taming's LPIPS network for LlamaGen and SEED-Voken and the \texttt{lpips} package for SimVQ. We omit
the GAN terms, SEED-Voken's entropy terms and GigaTok's DINOv2 feature distillation. The final
argmax selects one code per position.

The polish runs 8 steps.
Each step takes one of four interleaved colour classes ($2\times2$ on the token grid, the position modulo 4 on the 1D
sequences of TiTok and GigaTok B-L).
For every token of the class it proposes, over the whole codebook, the code $c$ that minimizes the damped first-order
model $g_t\T(c-q_t)+\frac\mu2\|c-q_t\|^2$, with $q$ the hard codes and $g$ the gradient there.
We accept the proposed updates for an image only if its discrete reconstruction loss decreases.
The damping $\mu$ halves after a kept step or when nothing is proposed, and doubles after a rejected one, within
$[u_{\mathrm c}/1024,1024u_{\mathrm c}]$, starting at $u_{\mathrm c}$, where
$u_{\mathrm c}=2\beta_{\mathrm c}/(nd)$ and the commitment weight $\beta_{\mathrm c}=0.25$.
SimVQ-$2^{18}$ runs without polish. In reconstruction alone ($\lambda=0$),
discrete refinement adds 0.002\,dB of PSNR. In a pilot
on 16 LlamaGen images, 50 steps with 16 candidates recovered 97\% of the
reconstruction gain from 150 steps with 64 candidates.
Everything in \cref{tab:recon} runs in float32, with PyTorch's defaults, which allow TF32 in convolutions on recent
GPUs.

\paragraph{Evaluation.}
We average per-image PSNR on $[0,1]$. For SEED-Voken models we use their evaluator on uint8 images.
\lpipsA uses the AlexNet version of the \texttt{lpips} package. For rFID, the reference is the same
set of original images. LlamaGen, FVQ, TiTok and GigaTok use clean-fid features on uint8 images.
SimVQ and SEED-Voken use the authors' pytorch-fid Inception. The \NN runs reproduce published rFID
within 7\% except for FVQ (1.15 vs.\ 0.88 under a different crop) and TiTok's 10k runs
(\cref{sec:recon}). We compute FID$_\infty$ as in \cref{sec:supp-bias}.
Evaluation uses 50k ImageNet-1k validation images, with the first 10k in ImageFolder order
(classes 0--199, 50 per class) for TiTok. Each tokenizer has one paired run.
Every rFID difference in \cref{tab:recon} exceeds five jackknife standard
errors, and KID agrees in sign. SimVQ uses torchvision for resizing and cropping. The
authors use albumentations, which can use a slightly different resizing filter, and the \NN run
reproduces their rFID.

\paragraph{\ours and the priors.}
The rate term is $\lambda$ times the mean-field bits per token under the class-conditional generator released for the
tokenizer: LlamaGen's GPT-B (\texttt{c2i\_B\_256.pt}, 111M) for VQ-16 and SEED-Voken's IBQ AR-B
(\texttt{AR\_256\_B.ckpt}) for IBQ-$2^{14}$.
Before training the downstream generators, we selected $\lambda=0.01$ on
500 validation images using reconstruction and rate ($+1.22$\,dB PSNR and
$-0.24$ bit per token relative to \NN in the LlamaGen pilot).
We condition the prior on the image class and turn off dropout. Under teacher forcing, the prior
uses the expected embeddings $\sum_j\pi_{uj}e(k_{uj})$ for positions $u<t$ when scoring position
$t$. \ours omits discrete refinement and runs in float32, except on \INk, where the decoder, LPIPS and
GPT-B run in bfloat16 during the relaxation (features and metrics use float32).

\paragraph{Cost and rate (\cref{fig:cost}).}
The experiment uses 10k validation images, every fifth (10 per class), with NN on the same images.
The cost group varies the steps (5, 10, 20, 50, each with 16 candidates and 8 polish steps), the candidates (4, 8) and
the polish (none).
The rate group varies $\lambda\in\{0.003,0.01,0.03,0.1\}$, with 50 steps, 16 candidates and no polish; its
$\lambda=0$ point is the cost group's run without polish.
Time is the CUDA-synchronized GPU time of the assignment alone (relaxation and polish), at batch 8 for LlamaGen and 6
for IBQ (4 and 2 on the RTX 3080s).
For LlamaGen on an RTX 5090, 50 steps take 0.99\,s per image: the polish is 0.195\,s of it, and the prior adds about
0.16\,s.
FID$_\infty$ uses the closed form, and the tests take the standard error of each difference to NN from five
disjoint folds of the images.

\paragraph{Generation (\INh, \INk).}
\INh uses classes $0,10,\dots,990$ (128,135 training and 5,000 validation images) and \INk all of ImageNet-1k (1,281,167
and 50,000).
Each training image gives two views, its ADM centre crop and the crop's horizontal flip, tokenized separately; each
item draws one of them at random.
The validation tokens come from the crop alone.
LlamaGen's ten crops (\cref{tab:crops}) are the corners and centre of a 281\,px centre crop and their horizontal flips;
training draws one per image and epoch.
\RX is the reconstruction setting (with polish); \ours uses $\lambda=0.01$ and no polish.
The released GPT-B used as the default prior and distillation teacher trained
for 300 epochs on ten crops from the full ImageNet training set.

The generator is LlamaGen's GPT-B: 12 layers, 12 heads, width 768, 111M parameters, 16,384 codes, 256 tokens and one
class token.
The 1000-class table is kept for \INh's 100 classes, with class dropout 0.1 for guidance, dropout 0.1 and token dropout
0.1.
It trains with LlamaGen's \texttt{train\_c2i.py}, unchanged except for the dataset: batch 256, AdamW at a constant
$10^{-4}$, $(\beta_1,\beta_2)=(0.9,0.95)$, weight decay 0.05 on tensors of two or more dimensions, gradient clipping at 1, bf16
autocast with TF32, no EMA, \texttt{torch.compile}, 100 epochs.
\INh runs 50,000 steps per run at the same global batch.
The runs of \cref{tab:gen} on \INh: seeds 0 and 1 of \NN, \RX and \ours on one GH200 each, seed 2 on other GPU models,
alone or under DDP.
\INk runs 500,400 steps per run at a global batch of 256.
\Cref{tab:gen}'s \INk runs use DDP: seed 0 on four RTX 5090s with 64 images per GPU (12.7 steps/s), seed 1 on four RTX
4090s (about 3.85 steps/s; 37--39\,h per 100 epochs with checkpoints and restarts), seed 2 on two GPUs with 128 per
GPU (two RTX 4090s, then two RTX 5090s).
The other students on both datasets (other priors, KD, three maps, ten crops) trained on various GPU models, mostly on
one GPU each.

Sampling uses LlamaGen's \texttt{generate()} at temperature 1, top-$k$ 0, top-$p$ 1 and a constant cfg scale, with the
GPT in bfloat16.
Labels are class-balanced, the same sequence for every run.
The 10k sweeps (sampling seed 0) cover cfg~1.0, 1.5, 2.0, 2.5 and 3.0 at epochs~100 and 50; the 50k final (sampling
seed 1) runs at the sweep's best cfg.
The training seeds of the runs are 0, 1 and 2.
gFID uses clean-fid features against the training images' centre crops, with every image (references, decoded token
sets, samples) quantized to uint8 once.
Precision and recall use $k=3$ on 10k against 10k. Additional evaluations use
FD$_{\mathrm{DINOv2}}$~\cite{stein2023exposing},
CMMD~\cite{jayasumana2024rethinking} and OpenAI's FID
evaluator~\cite{dhariwal2021diffusion} (\cref{sec:supp-gen}).
Validation bits per token are computed on the validation tokens of each variant.
FID$_\infty$ uses the class-balanced correction (\cref{sec:supp-bias}). Held-out gFID compares each
final's stored features with the validation images (5,000 for \INh and 50,000 for \INk).

\paragraph{Priors and distillation of \cref{sec:mechanism}.}
\MAPo tokenizes \INh with $\lambda=0.01$ under \INh's own \NN GPT-B (seed 0 at epoch~25) instead of the released one.
\MAPx splits the \INh training images into two halves. We tokenize each half with a GPT-B
trained only on the other half's \NN tokens.
KD trains GPT-B with the loss $0.5\,\mathrm{CE}(\text{tokens},\text{student})+0.5\,\mathrm{CE}(\text{teacher},
\text{student})$: the second term is the cross-entropy from the released GPT-B's next-token distribution, at
temperature 1, with the teacher teacher-forced on the same prefix and class (the null class where the student's class
is dropped) and its dropout off.
On \INk the released GPT-L also serves as teacher, with the same loss, for one seed per token set on two GPUs per
run.
It makes a training step 2.7 times as long as without a teacher (2.77 against 7.5 steps per second on one GH200; 4.5
with the released GPT-B as teacher), and a run takes about 53\,GH200-h.

\paragraph{GPT-L.}
The GPT-L generators (LlamaGen's GPT-L, 343M) train on \INk's token maps with the recipe above for 100 epochs, seed 0,
two GH200s per run, and are evaluated as \INk's GPT-Bs.
As a prior, the released GPT-L (\texttt{c2i\_L\_256.pt}, 300 epochs, ten crops) or our \NN-trained GPT-L at epoch~100
replaces the released GPT-B in \cref{eq:map}; everything else is \ours's.
For one seed per variant, we use a pre-registered tie range from the \INk GPT-B seeds: $\pm9.1\%$
for gFID-50k at the best cfg and $\pm8.7\%$ for held-out FID at cfg~2.0. We credit a difference
when it falls outside this range.

\paragraph{IBQ AR-B.}
The tokenizer is IBQ-$2^{14}$ (EMA).
The \NN variant uses IBQ's released assignment. \ours combines L1, LPIPS-VGG and $0.01\times$
the mean-field bits per token under the released AR-B. The pre-registered rule for the cost
experiment sets $\lambda$. We use 50 steps, 16 candidates, float32 and no discrete refinement.
Both arms get the two fixed views of \INh's images, with SEED-Voken's centre crop.
The generator is their AR-B (16 layers, 16 heads, width 1024, AdaLN; 343M parameters by our count), trained from
scratch per arm.
It runs on one GPU with a port of their recipe.
The optimizer is AdamW with $(\beta_1,\beta_2)=(0.9,0.95)$, peak learning rate $3\times10^{-4}$ and weight decay 0.05 on
every parameter group, as their schedule sets it.
The learning rate warms up over 6 epochs from $0.005\times$, stays constant for 5\% of the rest, then falls linearly to
$0.1\times$.
Training uses fp16 with loss scaling, clipping at 1, and residual, token-embedding and class-label dropout 0.1.
The global batch is 768 by gradient accumulation, with one random view per image per epoch, for 100 epochs (16,600
steps).
Sampling ports their sampler: temperature 1.1, top-$k$ 0, top-$p$ 1, constant cfg, float32 without TF32.
The sweep covers cfg~1.0, 1.5, 2.0, 2.25, 2.5 and 3.0; gFID is computed against the 128,135 training images with
SEED-Voken's centre crop.

IBQ AR-B favours \NN without guidance, and its tuned-guidance difference is within seed variation.
Every run's best cfg is 3.0.
gFID-50k there is 24.78 / 24.41 / 23.54 for \NN and 23.77 / 23.46 / 21.96 for \ours ($-$4.9\%, Welch $p=0.17$;
FID$_\infty$ $-$5.0\%).
The seed ranges overlap.
Paired by training seed (post hoc), \ours is lower by 1.01, 0.95 and 1.58 (paired $p=0.028$).
Without guidance (cfg~1.0, 10k samples) \ours is worse, 81.50 against 79.50 (Holm~$p=0.008$), unlike on LlamaGen.
In both arms the validation NLL rises by 1.3 bits from epoch~25 to 100; for context, the released AR-B, trained on
all 1000 classes for 300 epochs, reaches a 10k FID of 6.74 at cfg~2.25 on the same classes.
On a grid extended to cfg~5.0 (added before any \ours result) every run's best cfg moves to 3.5--4.0, and the 10k
comparison gives \ours 24.30 against \NN 25.09 ($-$3.2\%, $p=0.22$; on the registered grid $-$4.8\%, $p=0.21$).
With 50k samples at each run's tuned cfg (registered before they ran) \ours is 4.5\% lower:
22.15 against 23.20 (FID$_\infty$ $-$4.6\%; Holm~$p=0.29$).
At epoch~50, where both arms overfit less and both score better, \ours is lower on both grids
(10k: 21.91 against 23.18 up to cfg~3.0, Holm~$p=0.29$; 19.22 against 20.05 up to cfg~5.0, $p=0.26$); four of the six
runs are best at the extended grid's edge, cfg~5.0, so their minima may lie beyond it.

\paragraph{Masked generator.}
The masked generator is LlamaGen's GPT-B network with bidirectional attention, a [MASK] token (one more embedding
row; 110.9M parameters) and, at each image position, logits for the token at that position.
It trains with MaskGIT's objective~\cite{chang2022maskgit}: per image $r\sim U[0,1)$ and $\lceil256\cos(\pi r/2)\rceil$
positions masked at random, with the cross-entropy on the masked positions only and no label smoothing.
We use the \INh GPT-B training recipe, token files and budget (100 epochs, three seeds per variant).
The sampler implements MaskGIT's parallel decoding in $T=8$ steps. At each step, it draws tokens
at masked positions at temperature 1. It remasks the least confident draws according to the cosine
schedule, ranking by $\log p$ with Gumbel noise scaled by the choice temperature and $(1-t/T)$.
Guidance uses LlamaGen's constant cfg and null class.
Each final runs at the best point of a 10k sweep over cfg~1.0--5.0 and choice temperatures 1, 4.5 and 9, with 18
added where 9 was best; every final lies at cfg~2.0 or 2.5 and choice temperature 9 or 18.

\paragraph{Compute.}
\Cref{tab:compute} gives GPU times from the job records for the runs listed there. We report \INk GPT-B training throughput above and software versions in \cref{tab:env}.
Each cluster ran our jobs in one container image; seven of the 200 token shards of \ours on \INh's ten crops ran
in a second image, on V100s.
The workstation ran pilots, analyses of stored features, some registered experiments in full (the simpler searches of
\cref{sec:negative}, the surprise measure $\kappa$ and the nearest-neighbour test) and part of the token shards of
others, among them all of \NN's on \INh's ten crops.

\begin{table}[t]
\caption{\textbf{GPU time totalled from the job records} (job hours for the cost experiment; descriptive).  Not included: the other generator trainings, the generation evaluations, and the tokenizations with the
  GPT-L priors, $\lambda=0.03$, 35 steps, $\lambda=0.003$, L2 alone and the energy term.}
\label{tab:compute}
\centering\footnotesize
\setlength{\tabcolsep}{3pt}
\begin{tabular}{@{}>{\raggedright\arraybackslash}p{4.3cm}>{\raggedright\arraybackslash}p{1.45cm}r@{}}
\toprule
experiment & GPUs & GPU-h \\
\midrule
\Cref{tab:recon}: LlamaGen, SimVQ-$2^{10}$, $2^{18}$, 150k encodings with NN, metrics & GH200 & 11.1 \\
\Cref{tab:recon}: Open-MAGVIT2-L / IBQ-$2^{18}$ / IBQ-$2^{14}$, 50k each & RTX 4090 & 23.7 / 28.5 / 27.5 \\
TiTok-S-128 / SL-256, 10k each & RTX 4090 & 2.7 / 4.8 \\
GigaTok B-L / B-L + DINO / with its discriminator in $\ell$, 50k each & RTX 4090 & 29.4 / 34.4 / 30.8 \\
FVQ / GigaTok B-L-2D, 50k each & several & 22.3 / 34.8 \\
Cost and rate, LlamaGen: cost / rate group, 10k each & RTX 5090, A100 & 14.5 / 10.8 \\
Cost and rate, IBQ: rate group, 10k & five models & 25.4 \\
\INh tokenization, train (two views) + val: \NN / \RX / \ours & GH200, RTX 4090 & 0.5 / 79.5 / 97.2 \\
\INk tokenization, train (two views) + val: \NN / \ours & RTX 4090 & 11.4 / 800.2 \\
\INh tokenization: \MAPo / \MAPx (train + val) & several & 153.3 / 83.6 \\
IBQ AR-B tokenization: \ours, train & several & 126.4 \\
Cost and rate, IBQ: cost group, 10k & several & 27.2 \\
Estimators in LlamaGen's recipe (\cref{sec:negative}), training and evaluation & GH200 & 598 \\
\INh tokenization: three maps per view (two restarts, train) & GH200, RTX 4090 & 118.0 \\
\INk tokenization: three maps per view (two restarts, train) & several & 939.5 \\
\NN on LlamaGen's ten crops, train: \INh / \INk & RTX 3080 / several & 3.3 / 15.7 \\
\ours on LlamaGen's ten crops, \INh, train & seven models & 289.8 \\
Amortized encoder (\cref{sec:negative}), training & GH200 & 7.3 \\
GPT-L generators on \INk, two runs of 100 epochs & GH200 & 111.6 \\
GPT-B students with a GPT-L teacher, \INk, two runs & GH200, H100 & 107.3 \\
\bottomrule
\end{tabular}
\vspace{6pt}
\caption{\textbf{Software of the three environments} (each cluster's job container; A100, H100 and V100: cluster A's
  DGX nodes; descriptive).
  $^*$Seven token shards of \ours on \INh's ten crops, in a second image: Python 3.13.12, PyTorch 2.6.0, CUDA 12.6.}
\label{tab:env}
\centering\footnotesize
\setlength{\tabcolsep}{4pt}
\begin{tabular}{@{}lccc@{}}
\toprule
 & cluster A & cluster B & workstation \\
\midrule
CPU & x86-64 & aarch64 & x86-64 \\
GPUs & RTX 4090, 5090, & GH200 & RTX 3080 \\
 & A100, H100, V100$^*$ & & \\
Python & \multicolumn{2}{c}{3.14.6} & 3.14.4 \\
PyTorch & \multicolumn{2}{c}{2.13.0} & 2.11.0 \\
its CUDA / cuDNN & \multicolumn{2}{c}{13.2 / 9.20} & 13.0 / 9.19 \\
torchvision & \multicolumn{2}{c}{0.28.0} & 0.26.0 \\
NumPy / SciPy & \multicolumn{2}{c}{2.4.4 / 1.18.0} & 2.4.3 / 1.17.1 \\
Pillow & \multicolumn{2}{c}{12.2.0} & 12.1.1 \\
clean-fid / LPIPS & \multicolumn{2}{c}{0.1.35 / 0.1.4} & 0.1.35 / 0.1.4 \\
timm & \multicolumn{2}{c}{1.0.28} & 1.0.26 \\
\bottomrule
\end{tabular}
\end{table}

\section{Generation: further results}\label{sec:supp-gen}
\Cref{tab:supp-seeds} reports each run behind \cref{tab:gen}. The training seed varies within each
variant, with one set of token maps per variant and the same sampling seed and reference set for every run. A second
sampling seed changes each \INk epoch-100 gFID by at most 0.11 (see ``The literature's evaluator'' below).
The pre-registered two-seed rule for \INh credits a gain if the difference in means exceeds both
seed ranges. Under equal means and normally distributed seed noise of equal variance, the rule
credits a given direction with probability 10.8\% and either direction with probability 21.6\%
(numerical integration, checked with $2\cdot10^6$ Monte Carlo draws).

We registered third seeds after observing the two-seed results. With two seeds, \ours met the
pre-registered rule on \INh ($-10.9\%$). On \INk the difference was $-15.4\%$ (Welch $p=0.069$).
For AR-B, the two-seed rule credited a $-4.0\%$ difference. With three seeds the difference is
$-4.9\%$ ($p=0.17$).

\begin{table}[t]
\caption{\textbf{Per-seed gFID-50k for \cref{tab:gen} and \MAPo.}
  GPT-B trains for 100 epochs unless marked. Each run uses the cfg selected in its 10k sweep (2.5 for \INh,
  2.0 for \INk \ours, and bracketed values for \INk \NN). Held-out gFID uses validation images
  (5,000 for \INh, 50,000 for \INk). For \INk \NN seeds 0 and 2, the cfg-2.0 evaluations use the same
  checkpoints (post hoc). The pre-registered 300-epoch rows continue the same runs and evaluate held-out
  at cfg~2.0, using the companion evaluation for \NN seed 0 (6.09 against training images).}
\label{tab:supp-seeds}
\centering\footnotesize
\setlength{\tabcolsep}{3.2pt}
\begin{tabular}{@{}llrrr@{}}
\toprule
 & arm & seed 0 & seed 1 & seed 2 \\
\midrule
\multirow{4}{*}{\INh} & \NN & 12.18 & 12.09 & 11.88 \\
 & \RX & 11.41 & 11.56 & 11.66 \\
 & \ours & 10.81 & 10.81 & 11.08 \\
 & \MAPo & 11.20 & 11.61 & 11.80 \\
\midrule
\multirow{2}{*}{\INk} & \NN & 7.11 (2.5) & 6.82 (2.0) & 6.76 (2.5) \\
 & \ours & 5.93 & 5.85 & 5.53 \\
\midrule
\INk, & \NN, registered cfg & 7.99 (2.5) & 6.81 (2.0) & 7.64 (2.5) \\
held-out & \NN, cfg~2.0 & 6.83 & 6.81 & 6.72 \\
 & \ours & 6.38 & 6.32 & 5.88 \\
\midrule
\INk, & \NN & 6.38 (2.5) & 5.75 (2.0) & 5.63 (2.0) \\
300 epochs & \ours & 5.33 & 5.04 & 5.25 \\
 & \NN, held-out & 6.29 & 6.01 & 5.80 \\
 & \ours, held-out & 6.17 & 5.79 & 6.30 \\
\bottomrule
\end{tabular}
\end{table}

\paragraph{Held-out FID and guidance.}
Held-out gFID uses the stored Inception features of each 50k final with the validation split as
reference (5,000 images for \INh and 50,000 for \INk). Ten disjoint 5k subsets of \INh's training images
reproduce the training-reference gap ($-1.18\pm0.04$). This check points to the reference split as
the source of the smaller held-out gap.

Guidance affects the comparison. On \INk, \NN seeds 0 and 2 score 7.99 and 7.64 against held-out
images at their selected cfg~2.5, versus 6.83 and 6.72 at cfg~2.0. The held-out column of \cref{tab:crops} uses each
run's selected cfg, so the \INk two-view \NN baseline there is 7.48, against 6.79 when every run uses cfg~2.0 as in
\cref{tab:gen}. On \INh both variants score better
at their chosen cfg~2.5 than at 2.0 against both references. With cfg fixed at 2.0, the \ours
minus \NN difference on \INh is $-1.50$ $[-2.11,-0.90]$ against held-out images ($p=0.004$) and
$-2.06$ against training images ($p=0.003$). On \INk it is $-0.60$ $[-1.23,+0.04]$ ($p=0.057$)
and $-1.13$ $[-1.60,-0.66]$ ($p=0.0064$), respectively. Each \ours seed scores below every \NN
seed against both references in the \INk comparison. All these comparisons are post hoc.

\begin{table}[t]
\caption{\textbf{\Cref{tab:robust} at epoch~300}: the same \INk GPT-B runs continued to LlamaGen's
  300 epochs (pre-registered). Mean $\pm$ SD over three seeds per arm; $p$: Welch, two-sided, uncorrected; $^{\dagger}$post hoc.
  Bold: the better arm where $p<0.05$.
  \NN's registered finals use cfg~2.5, 2.0 and 2.0, \ours's cfg~2.0; the cfg-2.0 rows use a companion final for
  \NN's seed~0; the cfg-2.5 row regenerates every final except \NN's seed~0 at cfg~2.5.
  In gFID-50k, the ADM FID, the second sampling seed, FID-10k without guidance, and FD$_\mathrm{DINOv2}$ at cfg~2.0
  and 2.5, every \ours seed lies below every \NN seed; with three seeds per arm only FD$_\mathrm{DINOv2}$ at cfg~2.0
  and 2.5 reaches $p<0.05$.
  Values from \cref{tab:supp-seeds,tab:adm,tab:gfd} and this section.}
\label{tab:supp-300}
\centering\footnotesize
\setlength{\tabcolsep}{4pt}
\begin{tabular}{@{}lrrl@{}}
\toprule
\INk, GPT-B, epoch~300 & \NN & \ours & $p$ \\
\midrule
gFID-50k (registered) & 5.92 $\pm$ 0.40 & 5.21 $\pm$ 0.15 & 0.079 \\
OpenAI's evaluator, FID & 5.89 $\pm$ 0.43 & 5.18 $\pm$ 0.15 & 0.093 \\
second sampling seed & 5.93 $\pm$ 0.45 & 5.17 $\pm$ 0.13 & 0.089 \\
FID-10k, no guidance & 31.68 $\pm$ 0.26 & 26.05 $\pm$ 2.31 & 0.050 \\
held-out, cfg~2.0 & 6.03 $\pm$ 0.25 & 6.09 $\pm$ 0.27 & 0.80 \\
FD$_\mathrm{DINOv2}$, cfg~2.0$^{\dagger}$ & 296.4 $\pm$ 4.5 & \best{268.1 $\pm$ 8.3} & 0.013 \\
FD$_\mathrm{DINOv2}$, cfg~2.5$^{\dagger}$ & 248.6 $\pm$ 3.1 & \best{233.3 $\pm$ 5.3} & 0.020 \\
\bottomrule
\end{tabular}
\end{table}

\paragraph{The full schedule and proximity to training images.}
\Cref{tab:supp-300} lists every comparison of \cref{tab:robust} at epoch~300.
From epoch~100 to 300 the \INk gap in FID-10k at cfg~2.0 narrows from $-13.3\%$ to $-8.3\%$, with every \ours seed below
every \NN seed at every checkpoint; at a matched cfg~2.0 the 50k finals differ by $-10.5\%$ in-sample ($p=0.027$, post
hoc) and $+0.9\%$ held-out (pre-registered, $p=0.80$).
Without guidance (FID-10k at cfg~1.0) the epoch-300 gap is $-17.8\%$ (26.05 against 31.68; $p=0.050$, a pre-registered
secondary comparison without correction).
At epoch~300 (post hoc), held-out minus in-sample gFID at cfg~2.0 is 0.88 for \ours, 0.21 for \NN and 0.65 for the
released GPT-B; from epoch
100 to 300 held-out gFID improves by 0.76 for \NN (6.79 to 6.03) and by 0.11 for \ours (6.19 to 6.09), and \ours's validation bits
stay 0.18 lower throughout.

A pre-registered nearest-neighbour check measures how close the samples are to training images.
We reserve 50,000 training images as a reference subset $H$ and compute the exact nearest-neighbour
distance to the other 1.23M training images in Inception feature space. At epoch~300, 0.008\% of
\ours samples and 0.015\% of \NN samples fall below the first percentile of distances for $H$
($p=0.29$). The check gives no evidence of copying under this criterion.

In the descriptive comparison of the full distance distributions against $H$, \ours samples lie closer to the
training set (AUC 0.523 for \ours and 0.500 for \NN, $p=0.022$), and validation images lie farther away than $H$
(AUC 0.480).

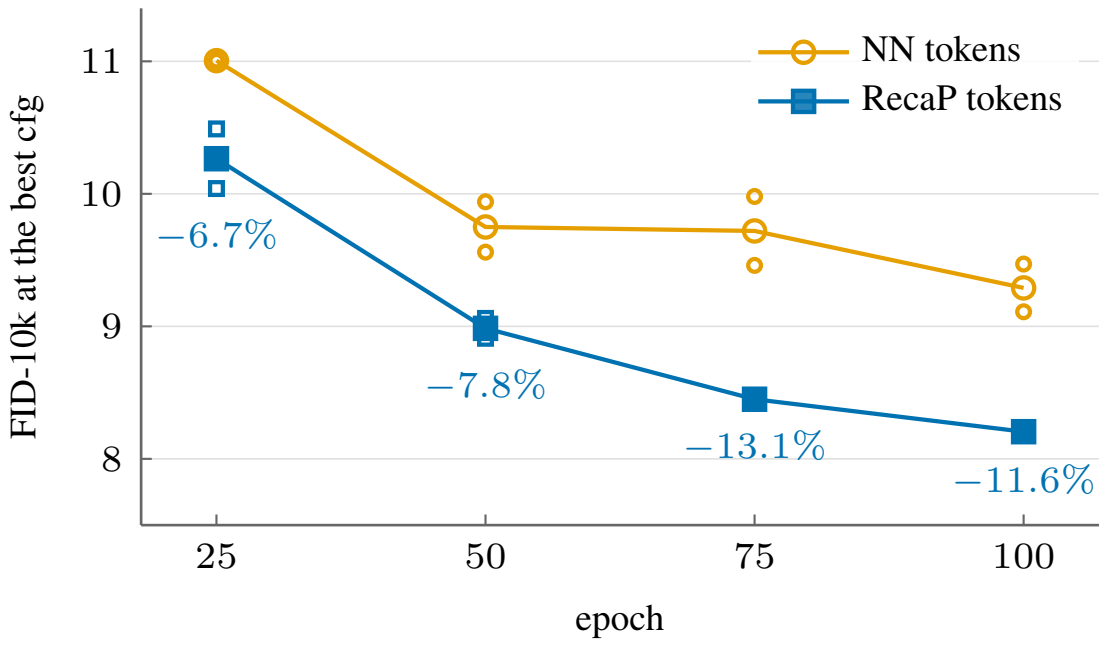
\begin{figure}[t]
\centering
\begin{tikzpicture}
\begin{axis}[
  width=0.95\columnwidth, height=5cm,
  xlabel={epoch}, ylabel={FID-10k at the best cfg},
  xtick={25,50,75,100}, xmin=18, xmax=107, ymin=7.5, ymax=11.4, paper,
  legend style={at={(0.98,0.98)}, anchor=north east}]
\addplot[color=nncol, mark=o] coordinates {(25,11.005) (50,9.75) (75,9.72) (100,9.29)};
\addlegendentry{\NN tokens}
\addplot[color=mapcol, mark=square*] coordinates {(25,10.265) (50,8.985) (75,8.45) (100,8.205)};
\addlegendentry{\ours tokens}
\addplot[only marks, mark=o, mark size=1.2pt, color=nncol, forget plot] coordinates {(25,11.00) (25,11.01)
  (50,9.94) (50,9.56) (75,9.98) (75,9.46) (100,9.47) (100,9.11)};
\addplot[only marks, mark=square, mark size=1.2pt, color=mapcol, forget plot] coordinates {(25,10.04) (25,10.49)
  (50,8.91) (50,9.06) (75,8.46) (75,8.44) (100,8.23) (100,8.18)};
\node[font=\scriptsize, color=mapcol, anchor=north] at (axis cs:25,9.95) {$-6.7\%$};
\node[font=\scriptsize, color=mapcol, anchor=north] at (axis cs:50,8.82) {$-7.8\%$};
\node[font=\scriptsize, color=mapcol, anchor=north] at (axis cs:75,8.36) {$-13.1\%$};
\node[font=\scriptsize, color=mapcol, anchor=north] at (axis cs:100,8.10) {$-11.6\%$};
\end{axis}
\end{tikzpicture}
\caption{\textbf{Early \INk training.} We plot each run's best FID-10k over cfg~1.0--3.0.
  Lines are the means of seeds 0 and 1 and marks show individual seeds. Percentages give the relative
  FID reduction with \ours tokens. The largest reduction in this sweep occurs at epoch~75.
  \Cref{fig:costgen} shows epochs 100--300 at cfg~2.0 with three seeds.}
\label{fig:epochs}
\end{figure}

\begin{figure*}[t]
\centering
\begin{tikzpicture}
\begin{axis}[
  width=0.5\textwidth, height=5.4cm, ymode=log, title={(a) \INh (100 classes)}, title style={font=\small},
  xlabel={cfg scale}, ylabel={FID-10k}, xtick={1,1.5,2,2.5,3}, xmin=0.5, xmax=3.1, ymin=10, ymax=75,
  ytick={12,15,20,30,50}, yticklabels={12,15,20,30,50}, minor ytick={}, paper,
  legend style={at={(0.98,0.98)}, anchor=north east}]
\addplot[color=nncol, mark=o] coordinates {(1.0,62.031) (1.5,28.952) (2.0,16.929) (2.5,14.213) (3.0,14.493)};
\addlegendentry{\NN}
\addplot[color=black!55, mark=triangle, densely dashed] coordinates {(1.0,62.767) (1.5,28.151) (2.0,16.179) (2.5,13.568) (3.0,14.127)};
\addlegendentry{\RXplot}
\addplot[color=mapcol, mark=square*] coordinates {(1.0,57.894) (1.5,25.500) (2.0,14.876) (2.5,13.001) (3.0,13.879)};
\addlegendentry{\ours}
\addplot[only marks, mark=o, color=nncol, mark size=1.1pt, forget plot] coordinates {(1.0,62.60) (1.5,29.01) (2.0,17.13) (2.5,14.40) (3.0,14.67) (1.0,60.60) (1.5,28.84) (2.0,16.63) (2.5,14.15) (3.0,14.55) (1.0,62.90) (1.5,29.01) (2.0,17.02) (2.5,14.09) (3.0,14.26)};
\addplot[only marks, mark=triangle, color=black!55, mark size=1.1pt, forget plot] coordinates {(1.0,62.96) (1.5,28.34) (2.0,16.08) (2.5,13.32) (3.0,14.08) (1.0,62.51) (1.5,27.84) (2.0,16.06) (2.5,13.54) (3.0,14.20) (1.0,62.84) (1.5,28.27) (2.0,16.40) (2.5,13.84) (3.0,14.10)};
\addplot[only marks, mark=square, color=mapcol, mark size=1.1pt, forget plot] coordinates {(1.0,57.04) (1.5,25.49) (2.0,14.49) (2.5,12.90) (3.0,13.86) (1.0,57.40) (1.5,24.95) (2.0,14.87) (2.5,12.78) (3.0,13.63) (1.0,59.24) (1.5,26.05) (2.0,15.27) (2.5,13.32) (3.0,14.15)};
\node[font=\scriptsize, color=mapcol, anchor=east, xshift=-3pt] at (axis cs:1.0,57.894) {$-6.7\%$};
\node[font=\scriptsize, color=mapcol, anchor=north, yshift=-2pt] at (axis cs:1.5,25.5) {$-11.9\%$};
\node[font=\scriptsize, color=mapcol, anchor=north, yshift=-2pt] at (axis cs:2.0,14.876) {$-12.1\%$};
\node[font=\scriptsize, color=mapcol, anchor=north east, yshift=-2pt, xshift=-1pt] at (axis cs:2.5,13.001) {$-8.5\%$};
\node[font=\scriptsize, color=mapcol, anchor=north east, yshift=-2pt, xshift=-1pt] at (axis cs:3.0,13.879) {$-4.2\%$};
\end{axis}
\end{tikzpicture}\hfill
\begin{tikzpicture}
\begin{axis}[
  width=0.5\textwidth, height=5.4cm, ymode=log, title={(b) \INk (all 1000 classes)}, title style={font=\small},
  xlabel={cfg scale}, ylabel={FID-10k}, xtick={1,1.5,2,2.5,3}, xmin=0.5, xmax=3.1, ymin=6, ymax=42,
  ytick={7,10,15,20,30}, yticklabels={7,10,15,20,30}, minor ytick={}, paper,
  legend style={at={(0.98,0.98)}, anchor=north east}]
\addplot[color=nncol, mark=o] coordinates {(1.0,34.511) (1.5,14.343) (2.0,9.318) (2.5,9.221) (3.0,10.558)};
\addlegendentry{\NN}
\addplot[color=mapcol, mark=square*] coordinates {(1.0,29.554) (1.5,11.606) (2.0,8.082) (2.5,8.885) (3.0,10.444)};
\addlegendentry{\ours}
\addplot[only marks, mark=o, color=nncol, mark size=1.1pt, forget plot] coordinates {(1.0,35.01) (1.5,14.46) (2.0,9.52) (2.5,9.47) (3.0,10.63) (1.0,33.22) (1.5,13.50) (2.0,9.11) (2.5,9.39) (3.0,10.92) (1.0,35.30) (1.5,15.07) (2.0,9.32) (2.5,8.81) (3.0,10.12)};
\addplot[only marks, mark=square, color=mapcol, mark size=1.1pt, forget plot] coordinates {(1.0,29.28) (1.5,11.36) (2.0,8.23) (2.5,9.08) (3.0,10.65) (1.0,29.04) (1.5,11.44) (2.0,8.18) (2.5,9.14) (3.0,10.75) (1.0,30.34) (1.5,12.02) (2.0,7.83) (2.5,8.43) (3.0,9.92)};

\node[font=\scriptsize, color=mapcol, anchor=east, xshift=-3pt] at (axis cs:1.0,29.554) {$-14.4\%$};
\node[font=\scriptsize, color=mapcol, anchor=north, yshift=-2pt] at (axis cs:1.5,11.606) {$-19.1\%$};
\node[font=\scriptsize, color=mapcol, anchor=north, yshift=-2pt] at (axis cs:2.0,8.082) {$-13.3\%$};
\node[font=\scriptsize, color=mapcol, anchor=north, yshift=-2pt] at (axis cs:2.5,8.885) {$-3.6\%$};
\node[font=\scriptsize, color=mapcol, anchor=north east, yshift=-2pt, xshift=-1pt] at (axis cs:3.0,10.444) {$-1.1\%$};
\end{axis}
\end{tikzpicture}
\caption{\textbf{FID-10k over the registered cfg grid at epoch~100}: 10k samples per point (sampling seed 0) against the training
  reference; lines are means of three seeds, small marks single seeds; labels: \ours's change from \NN (descriptive).
  (a) \INh: \NN, \RX and \ours. (b) \INk: \NN and \ours.
  On \INh \ours is below \NN at every cfg ($-4$ to $-12\%$), every seed below every \NN seed; on \INk up to cfg~2.0
  ($-13$ to $-19\%$, likewise), while at 2.5 and 3.0 the gap shrinks to $-3.6\%$ and $-1.1\%$, inside the seeds'
  spread.
  gFID-50k (\cref{tab:gen}) is computed at each run's best cfg of this grid.}
\label{fig:cfg}
\end{figure*}
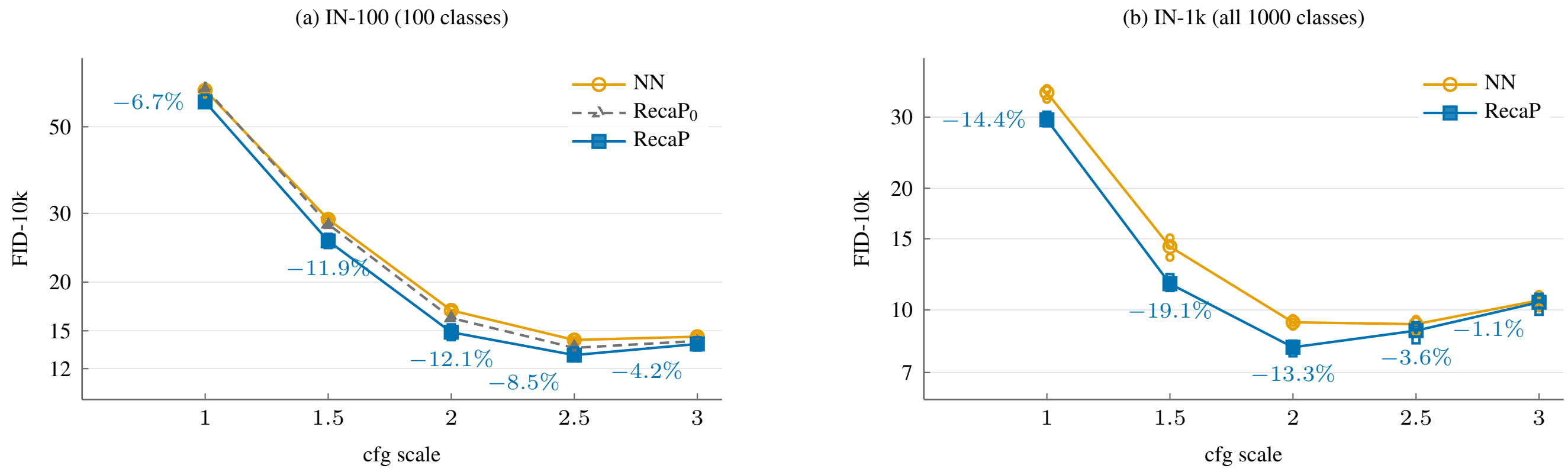

\begin{figure*}[tp]
\centering
\begin{tikzpicture}
  \node[inner sep=0, anchor=south west] (s) at (0,0) {\includegraphics[width=0.70\textwidth]{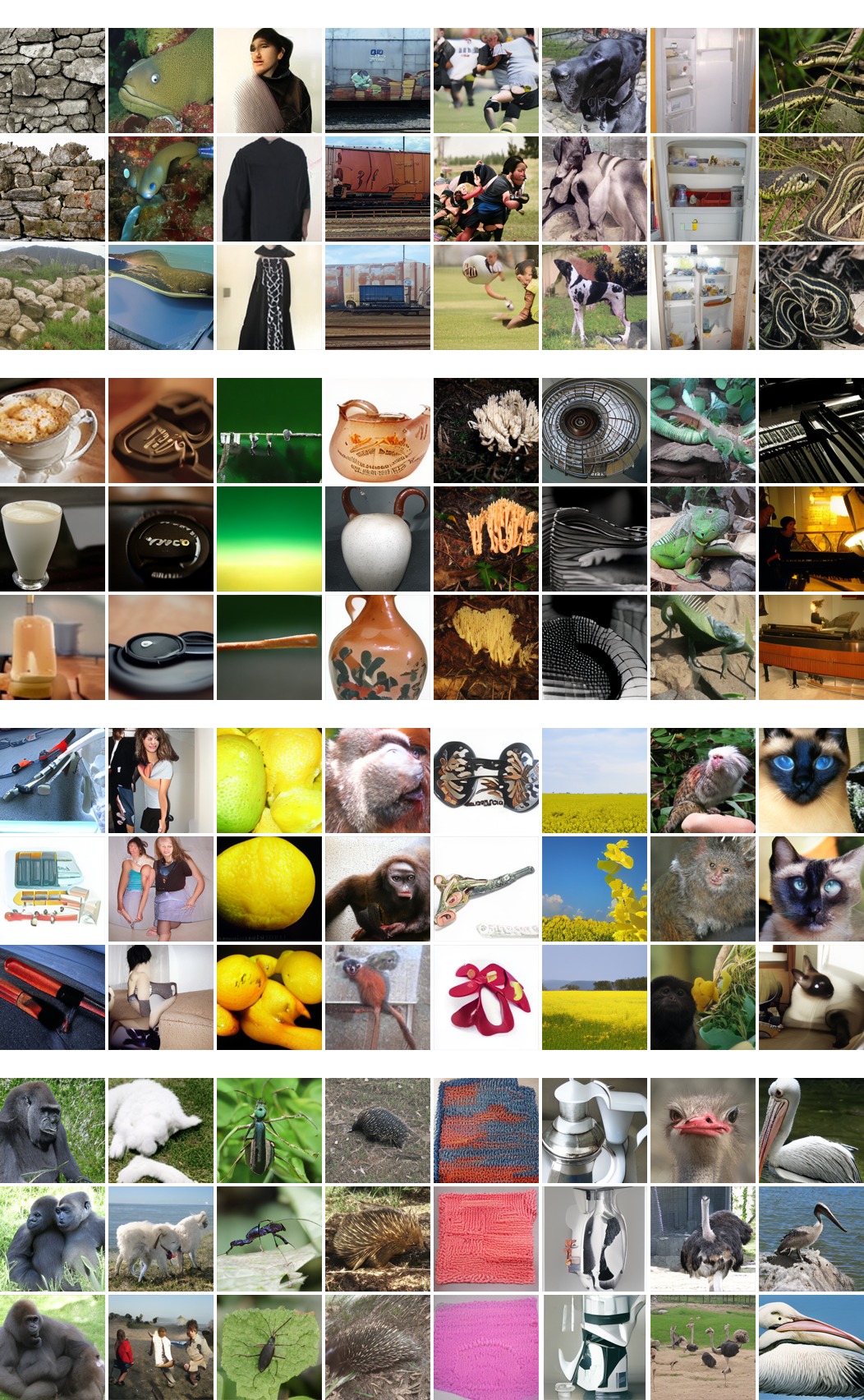}};
  \begin{scope}[x={(s.south east)}, y={(s.north west)}]
    \foreach \g/\names in {0/{stone wall,eel,abaya,freight car,rugby ball,Great Dane,refrigerator,garter snake},
                           1/{eggnog,lens cap,nail,whiskey jug,coral fungus,coil,common iguana,grand piano},
                           2/{screwdriver,miniskirt,lemon,titi,hair slide,rapeseed,marmoset,Siamese cat},
                           3/{gorilla,kuvasz,long-horned beetle,echidna,dishrag,coffeepot,ostrich,pelican}}
      \foreach \n [count=\c from 0] in \names
        \node[font=\scriptsize, inner sep=1pt, text height=1.6ex, text depth=0.4ex] at ({(64+132*\c)/1052}, {1-(\g*426+18)/1704}) {\n};
    \foreach \g in {0,1,2,3}
      \foreach \m/\t in {0/released, 1/{\NN}, 2/{\ours}}
        \node[anchor=south, rotate=90, font=\scriptsize, inner sep=3pt] at (0, {1-(\g*426+34+132*\m+64)/1704}) {\t};
  \end{scope}
\end{tikzpicture}
\caption{\textbf{Samples on \INk}: the first 32 images of each 50k final at cfg~2.0.
  Each column shares one class label (above) and one sampling seed.
  Rows: the released LlamaGen GPT-B (300 epochs, ten crops; generated anew with the final's labels and sampling seed), and GPT-B trained for 100 epochs on \NN tokens (seed 0;
  its cfg-2.0 companion final, as the registered one ran at cfg~2.5) and on \ours tokens (seed 0).
  Each image is downscaled from 256 to 128\,px.}
\label{fig:samples}
\end{figure*}

\paragraph{Training curves and samples.}
\Cref{fig:epochs} follows \INk through training with two seeds; with three seeds the epoch-50 values are
9.72 $\pm$ 0.19 (\NN) and 8.96 $\pm$ 0.09 (\ours) in FID-10k at the best cfg.
In gFID-50k, \ours at epoch~50 scores 6.59 $\pm$ 0.11, compared with 6.90 $\pm$ 0.19 for \NN at
epoch~100 ($p=0.09$, with all \ours seeds below all \NN seeds).
Every \INh run of \NN, \RX, \ours and \MAPo overfits (validation NLL ends 0.37--0.46 bit above its minimum; on one
view per image, 1.2 bits); the students on ten crops do not (excess 0), those on three maps per view hardly (0.06 bit),
and the KD students' validation bits still fall from epoch~25 to 100; at epoch~50 no difference among those four arms is significant (\ours against \NN: Holm~$p=0.71$).
\Cref{fig:samples} shows the first 16 samples of three 50k finals, with the same class labels and sampling seed.
\Cref{fig:cfg} shows the FID-10k of each epoch-100 checkpoint of \cref{tab:gen}'s runs over the registered cfg grid: on \INh \ours is ahead at every cfg,
on \INk up to cfg~2.0, and at 2.5--3.0 the \INk arms are within their seeds' spread.

\paragraph{The literature's evaluator, CMMD and the sampling seed (pre-registered).}
We regenerated every \INk final with its checkpoint, cfg, labels and sampling seed and scored the samples with
OpenAI's ADM evaluator~\cite{dhariwal2021diffusion}, unchanged, and with CMMD~\cite{jayasumana2024rethinking} against
the 50k validation images (\cref{tab:adm}).
Each regeneration of a registered final reproduces its clean-fid gFID-50k within 1\% (a post hoc cfg-2.0 companion
of \NN's seed 0: $+1.6\%$).
At epoch~100 the evaluator confirms the gain of \cref{tab:gen} ($-17.2\%$ against clean-fid's $-16.4\%$, $p=0.002$,
every \ours seed below every \NN seed), and the released GPT-B falls within the registered range around its published scores (FID 5.23 and IS 192.8
against 5.46 and 193.6).
At epoch~300, the ADM evaluator gives a difference of $-12.1\%$ ($p=0.09$), close to clean-fid's
$-12.0\%$. The sFID, IS, precision and recall comparisons are inconclusive at both epochs.

CMMD gives differences of $+7.6\%$ ($p=0.21$) at epoch~100 and $-0.9\%$ ($p=0.88$) at epoch~300. As in DINOv2 space,
\NN's finals score better at cfg~2.5 than at 2.0: at epoch~100 its seeds 0 and 2 score 0.467 and 0.475 at cfg~2.5 and 0.547 and 0.556 at 2.0.
At a matched cfg~2.0 (post hoc) CMMD differs by $-2.9\%$ ($p=0.17$) and $-5.6\%$ ($p=0.060$, every \ours seed below
every \NN seed). 

A second sampling seed moves each final's gFID-50k by at most 0.11 at epoch~100 (registered bound 0.10, missed by
one \NN final) and 0.08 at epoch~300, against
differences of 1.13 and 0.71 between the arms; with it the gain is $-15.1\%$ (6.84 against 5.81, $p=0.005$) and
$-12.8\%$ (5.93 against 5.17, $p=0.09$).

\begin{table*}[t]
\caption{\textbf{The \INk finals under OpenAI's ADM evaluator} (FID, sFID, IS, precision, recall; reference
  VIRTUAL\_imagenet256\_labeled) and CMMD (CLIP ViT-L/14 at 336\,px against ImageNet's 50k validation images, computed
  in float64), each final regenerated with its checkpoint, cfg, labels and sampling seed (pre-registered).
  Bottom rows: the relative difference of the arms' means; $^{*}$Welch $p<0.05$ ($p=0.002$; every other $p\ge0.09$).
  Registered finals only (the post hoc cfg-2.0 companions are left out); per-seed rows are not ranked.}
\label{tab:adm}
\centering\footnotesize
\setlength{\tabcolsep}{4pt}
\begin{tabular}{@{}llrrrrrr@{}}
\toprule
 & run (cfg) & IS $\uparrow$ & FID $\downarrow$ & sFID $\downarrow$ & Prec. $\uparrow$ & Rec. $\uparrow$ & CMMD $\downarrow$ \\
\midrule
\multirow{7}{*}{epoch~100} & \NN seed 0 (2.5) & 202.2 & 7.15 & 6.70 & 0.885 & 0.358 & 0.467 \\
 & \NN seed 1 (2.0) & 163.4 & 6.76 & 5.93 & 0.850 & 0.424 & 0.533 \\
 & \NN seed 2 (2.5) & 204.1 & 6.86 & 6.25 & 0.888 & 0.357 & 0.475 \\
 & \ours seed 0 (2.0) & 178.9 & 5.91 & 6.36 & 0.848 & 0.433 & 0.530 \\
 & \ours seed 1 (2.0) & 180.2 & 5.80 & 5.81 & 0.852 & 0.427 & 0.517 \\
 & \ours seed 2 (2.0) & 178.5 & 5.49 & 5.76 & 0.843 & 0.447 & 0.539 \\
 & released GPT-B (2.0) & 192.8 & 5.23 & 7.53 & 0.838 & 0.450 & 0.519 \\
\midrule
\multirow{6}{*}{epoch~300} & \NN seed 0 (2.5) & 218.6 & 6.38 & 7.03 & 0.885 & 0.381 & 0.435 \\
 & \NN seed 1 (2.0) & 179.3 & 5.73 & 6.08 & 0.849 & 0.445 & 0.519 \\
 & \NN seed 2 (2.0) & 176.2 & 5.56 & 6.17 & 0.846 & 0.450 & 0.502 \\
 & \ours seed 0 (2.0) & 195.1 & 5.32 & 6.35 & 0.855 & 0.442 & 0.488 \\
 & \ours seed 1 (2.0) & 195.7 & 5.03 & 5.74 & 0.852 & 0.443 & 0.492 \\
 & \ours seed 2 (2.0) & 206.4 & 5.19 & 6.99 & 0.861 & 0.433 & 0.464 \\
\midrule
\multicolumn{2}{@{}l}{\ours $-$ \NN, epoch~100} & $-$5.6\% & $-$17.2\%$^{*}$ & $-$5.0\% & $-$3.0\% & +14.7\% & +7.6\% \\
\multicolumn{2}{@{}l}{\ours $-$ \NN, epoch~300} & +4.0\% & $-$12.1\% & $-$1.1\% & $-$0.5\% & +3.3\% & $-$0.9\% \\
\bottomrule
\end{tabular}
\end{table*}

\paragraph{Per-class generation results (descriptive).}
Per class, \ours's \INk finals have lower KID against the training images than \NN's: per-class KID ($\times10^3$) is 16.75
against 18.86 at the registered cfgs ($-11\%$, lower in 73\% of the classes) and against 20.69 at a matched cfg~2.0
($-19\%$; 71 classes gain beyond three standard deviations, one loses).
At the registered cfgs the Inception classifier's top-1 agrees with the conditioning class on 82.2\% of \NN's samples
and 80.5\% of \ours's; at a matched cfg~2.0 \NN's falls to 78.0\%, and \ours is higher in 63\% of the classes (the
released GPT-B: 81.8\%; real training images: 91.9\%).
We group the classes by fixed ranges of the class index: dogs, birds, food, nature scenes, 17 text-bearing
classes, other animals and other objects.
At matched cfg~2.0, per-class KID improves most in absolute terms for other objects, text-bearing
objects, and food. Nature scenes are the only group without an improvement.

\paragraph{Published ImageNet generators.}
\Cref{tab:context} gives published scores for class-conditional generation on ImageNet at 256\,px.
The models differ in tokenizer, architecture, training length, guidance and evaluation. Sampling
50 images per class alone can lower FID by about 0.1 relative to random labels~\cite{zheng2025rae}.
At epoch~300, our GPT-B with two-view \ours tokens scores close to the released GPT-B with ten
crops against the training images. Its held-out score is 6.09 vs.\ 5.86. Published guided models
with 200--310M parameters and different architectures reach 1.95--3.30, and the lowest published
scores are near 1.
For our GPT-L, the differences fall within the pre-registered tie range (\cref{sec:gen}).
\ours's 100-epoch GPT-B scores 5.77, 0.57 above the released GPT-B (300 epochs, ten crops), which scores 5.20
under our evaluator (published 5.46; 5.31 with i.i.d.\ labels, \cref{prop:supp-gfid}); the released GPT-L scores 3.93, against 3.95 for our
100-epoch GPT-L on \ours's tokens with the released GPT-L as prior (held-out at each model's best cfg, 1.5 and 2.0:
4.40 against 5.19).

\begin{table}[t]
\caption{\textbf{Class-conditional generation on ImageNet at 256\,px} (FID-50k, with guidance unless noted).
  Ours: GPT-B and GPT-L on LlamaGen VQ-16's $16\times16$ tokens of two views per image (each run's best cfg; three
  seeds for GPT-B, one for GPT-L), and the released LlamaGen models under our evaluator; clean-fid with
  class-balanced labels.
  Published: each paper's own table (LlamaGen: its $16\times16$, 256\,px models; VAR: arXiv v2, whose v1 gives 3.60 and
  1.97; RAE: 1.28 with random labels).
  Bold: the lowest FID; underlined: the second lowest (protocols differ, so the ranking is indicative).}
\label{tab:context}
\centering\footnotesize
\setlength{\tabcolsep}{2.6pt}
\begin{tabular}{@{}lrlr@{}}
\toprule
generator & params & training & FID-50k $\downarrow$ \\
\midrule
\multicolumn{4}{@{}l}{\emph{ours (OpenAI's evaluator in brackets)}} \\
GPT-B, \NN tokens & 111M & 100 ep. & 6.90 (6.92) \\
GPT-B, \ours tokens & 111M & 100 ep. & 5.77 (5.73) \\
GPT-B, \NN tokens & 111M & 300 ep. & 5.92 (5.89) \\
GPT-B, \ours tokens & 111M & 300 ep. & 5.21 (5.18) \\
GPT-L, \NN tokens & 343M & 100 ep. & 4.15 \\
GPT-L, \ours (GPT-B prior) & 343M & 100 ep. & 4.13 \\
GPT-L, \ours (GPT-L prior) & 343M & 100 ep. & 3.95 \\
released GPT-B (ten crops) & 111M & 300 ep. & 5.20 (5.23) \\
released GPT-L (ten crops) & 343M & 300 ep. & 3.93 \\
\midrule
\multicolumn{4}{@{}l}{\emph{published, each paper's protocol}} \\
LlamaGen GPT-B~\cite{sun2024llamagen} & 111M & 300 ep. & 5.46 \\
LlamaGen GPT-L~\cite{sun2024llamagen} & 343M & 300 ep. & 3.80 \\
MaskGIT~\cite{chang2022maskgit} & 227M & no guidance & 6.18 \\
VAR-d16~\cite{tian2024var} & 310M & & 3.30 \\
MAR-B~\cite{li2024mar} & 208M & & 2.31 \\
RAR-B~\cite{yu2024rar} & 261M & & 1.95 \\
DiT-XL/2~\cite{peebles2023dit} & 675M & & 2.27 \\
VAR-d30~\cite{tian2024var} & 2.0B & & 1.92 \\
MAR-H~\cite{li2024mar} & 943M & & 1.55 \\
RAR-XXL~\cite{yu2024rar} & 1.5B & & \second{1.48} \\
RAE, DiT$^\mathrm{DH}$-XL~\cite{zheng2025rae} & 839M & AutoGuidance & \best{1.13} \\
\bottomrule
\end{tabular}
\end{table}

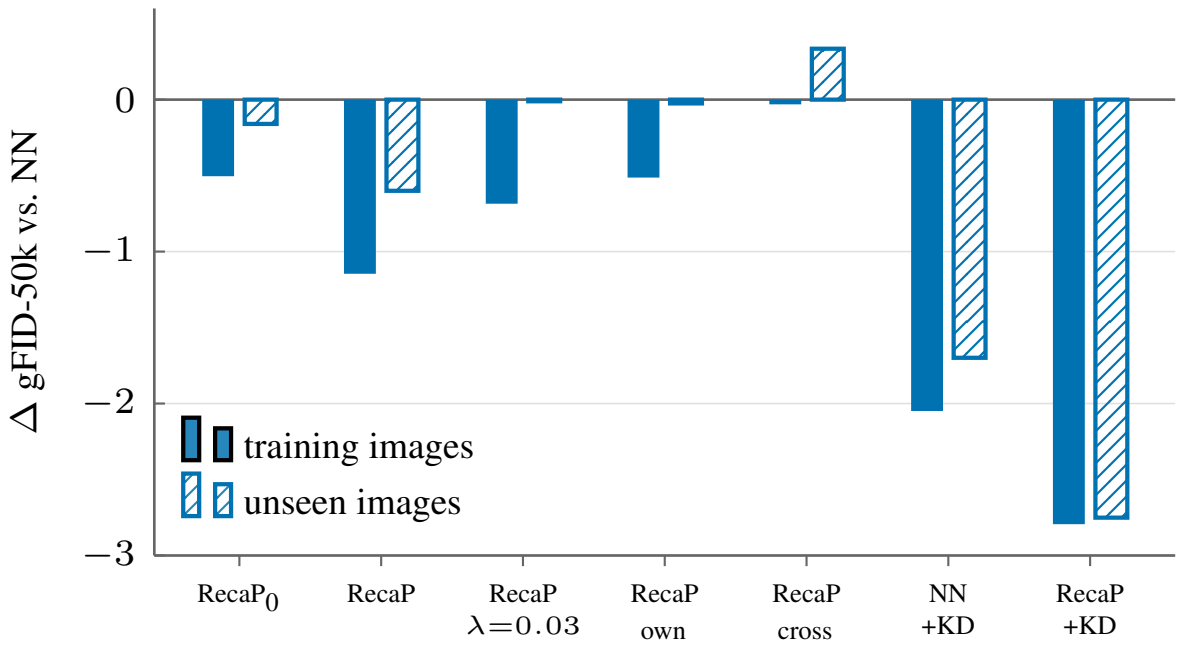
\begin{figure}[t]
\centering
\begin{tikzpicture}
\begin{axis}[
  ybar, width=\columnwidth, height=5.2cm, bar width=6pt,
  ymin=-3.0, ymax=0.6, ylabel={$\Delta$ gFID-50k vs.\ NN},
  xtick={1,2,3,4,5,6,7}, xmin=0.4, xmax=7.6,
  xticklabels={\oursname\recapplotsubscript{0}, \oursname, \oursname\\$\lambda{=}0.03$, \oursname\\\recapplotsubscript{own},
    \oursname\\\recapplotsubscript{cross}, NN\\+KD, \oursname\\+KD},
  xticklabel style={font=\tiny, align=center}, paper,
  legend style={at={(0.02,0.03)}, anchor=south west}]
\draw[black!60, line width=0.5pt] (axis cs:0.4,0) -- (axis cs:7.6,0);
\addplot[fill=mapcol, draw=none] coordinates {(1,-0.504) (2,-1.146) (3,-0.685) (4,-0.513) (5,-0.031) (6,-2.049)
  (7,-2.794)};
\addlegendentry{training images}
\addplot[fill=hocol, draw=mapcol, pattern=north east lines, pattern color=mapcol] 
coordinates {(1,-0.160)
  (2,-0.600) (3,-0.011) (4,-0.025) (5,0.335) (6,-1.699) (7,-2.751)};
\addlegendentry{unseen images}
\end{axis}
\end{tikzpicture}
\caption{\textbf{Change in gFID-50k against \NN tokens on \INh} (three seeds per arm; values in \cref{tab:mech};
  bar colours mark the reference, training or unseen images, not the token set).
  Without distillation, only \ours (released prior, $\lambda=0.01$) keeps a gain on unseen images with $p<0.05$
  (0.018, post hoc and unadjusted).
  Better reconstruction alone (\RX), a larger rate weight and own or cross-fitted priors gain in-sample or not at all.}
\label{fig:ablation}
\end{figure}

\paragraph{Generation in DINOv2 space.}
On \INh \ours lowers the Fr\'echet distance in DINOv2 feature space; on \INk it does so when both arms use the
same cfg. We regenerate the \cref{tab:gen} and AR-B samples using the same checkpoints, cfg, labels and
sampling seeds. Each regeneration reproduces its registered Inception FID within tolerance.
We compute FD$_\mathrm{DINOv2}$ against both training and validation images in a pre-registered
comparison (\cref{tab:gfd}).

On \INh, \ours scores 3.0\% below \NN (Holm $p=0.025$), with a 2.7\% held-out reduction (Holm
$p=0.024$). On \INk at the registered cfgs, \ours scores 3.0\% above \NN ($p=0.60$). Two \NN
runs score 16\% lower at cfg~2.5 than in their companion evaluations at cfg~2.0, while their
Inception FID changes by at most 0.14. At matched cfg~2.0, \ours scores 8.5\% below \NN
(post hoc, $p=0.009$), and at matched cfg~2.5, which lowers both arms' scores, 6.1\% below (post hoc, $p=0.011$;
held-out $-5.6\%$, $p=0.011$), with every \ours seed below every \NN seed at both cfgs. At epoch~300, FD$_\mathrm{DINOv2}$ differs by $-4.2\%$ ($p=0.49$) against training images and
$-3.9\%$ against validation images; at a matched cfg~2.0 (post hoc) it differs by $-9.6\%$ ($p=0.013$) and $-8.9\%$,
and at cfg~2.5 by $-6.1\%$ ($p=0.020$) and $-5.3\%$ ($p=0.029$), each time with every \ours seed below every \NN seed. At the registered cfgs the epoch-300 improvement is clearest with Inception
features.
For AR-B, the difference is $-0.9\%$ ($p=0.55$, unadjusted).
On \INh, \RX scores 4.1\% above \NN in DINOv2 space (descriptive, $p=0.0045$) and 4.6\% below
\NN in Inception space on the same regenerated samples. The generation gains therefore depend
on the feature space and the sampling protocol.

\begin{table}[t]
\caption{\textbf{FD$_\mathrm{DINOv2}$ of the generation finals} (50k samples each, regenerated; mean $\pm$ SD over three seeds)
  against the training images; held-out: the mean against the validation images; DINOv2 precision and recall ($k=3$,
  10k samples).
  \INk's registered \NN finals ran at cfg~2.5, 2.0 and 2.5; the cfg-2.0 row uses companion finals for seeds 0 and 2;
  the cfg-2.5 rows regenerate \NN's seed~1 and every \ours final at cfg~2.5 (post hoc).
  Ep.~300: the same runs continued to epoch~300 (registered \NN finals at cfg~2.5, 2.0 and 2.0; the cfg-2.0 row uses
  a companion final for seed~0, and the cfg-2.5 rows regenerate \NN's seeds 1 and 2 and every \ours final at cfg~2.5,
  post hoc).
  Released: LlamaGen's GPT-B. \ours{}+energy: with the local energy term (\cref{sec:supp-sharp}).
  Bold: the lowest value of a block; underlined: the second lowest (blocks of three or more ranked rows); the released
  model and P / R are not ranked.}
\label{tab:gfd}
\centering\footnotesize
\setlength{\tabcolsep}{2.2pt}
\begin{tabular}{@{}llrrr@{}}
\toprule
 & arm (cfg) & FD$_\mathrm{DINOv2}$ $\downarrow$ & held-out $\downarrow$ & P / R \\
\midrule
\multirow{4}{*}{\INh} & \NN (2.5) & 542.0 $\pm$ 4.9 & 615.2 & 0.474 / 0.141 \\
 & \RX (2.5) & 564.0 $\pm$ 4.1 & 636.4 & 0.473 / 0.131 \\
 & \ours (2.5) & \second{525.6 $\pm$ 4.2} & \second{598.6} & 0.477 / 0.126 \\
 & \ours{}+energy (2.5) & \best{522.9 $\pm$ 7.5} & \best{596.1} & 0.479 / 0.131 \\
\midrule
\multirow{6}{*}{\INk} & \NN (registered) & 291.2 $\pm$ 24.3 & 301.2 & 0.630 / 0.413 \\
 & \NN (2.0) & 327.7 $\pm$ 7.3 & 336.5 & 0.602 / 0.439 \\
 & \ours (2.0) & 299.9 $\pm$ 4.0 & 309.7 & 0.616 / 0.450 \\
 & \NN (2.5) & \second{274.4 $\pm$ 4.9} & \second{285.0} & 0.642 / 0.400 \\
 & \ours (2.5) & \best{257.6 $\pm$ 3.1} & \best{269.1} & 0.652 / 0.395 \\
 & released (2.0) & 270.0 & 279.9 & 0.627 / 0.478 \\
\midrule
\multirow{5}{*}{\begin{tabular}[c]{@{}l@{}}\INk,\\ep.~300\end{tabular}} & \NN (registered) & 280.0 $\pm$ 24.3 & 290.1 & 0.639 / 0.450 \\
 & \NN (2.0) & 296.4 $\pm$ 4.5 & 305.9 & 0.623 / 0.463 \\
 & \ours (2.0) & 268.1 $\pm$ 8.3 & 278.8 & 0.643 / 0.458 \\
 & \NN (2.5) & \second{248.6 $\pm$ 3.1} & \second{259.8} & 0.667 / 0.419 \\
 & \ours (2.5) & \best{233.3 $\pm$ 5.3} & \best{245.9} & 0.683 / 0.410 \\
\midrule
\multirow{2}{*}{AR-B} & \NN (3.0) & 728.1 $\pm$ 10.0 & 797.5 & 0.448 / 0.247 \\
 & \ours (3.0) & \best{721.3 $\pm$ 14.5} & \best{790.8} & 0.449 / 0.243 \\
\bottomrule
\end{tabular}
\end{table}

\paragraph{Distillation on \INk.}
We use the same token-level distillation as on \INh. Here the released GPT-B and its students
both saw 1.28M images. The comparison is pre-registered and uses three seeds per variant, with
all distilled models evaluated at cfg~2.0.
Held-out gFID-50k is 7.48 $\pm$ 0.61 (\NN at its registered cfgs, two finals at 2.5; 6.79 at cfg~2.0), 6.19 $\pm$
0.27 (\ours), 6.59 $\pm$ 0.15 (\KDNN) and 6.48 $\pm$ 0.21
(\KDMAP), and no difference is significant (Holm~$p>0.4$).
In-sample (6.90, 5.77, 6.09 and 5.74) KD improves \NN by 11.7\% (Holm~$p=0.03$), but every \KDNN seed lies above
every \ours seed (Holm~$p=0.21$), and \KDMAP lies 0.5\% below \ours.
On \NN tokens KD thus recovers 71\% of \ours's in-sample gain and gives \ours's tokens no credited gain; on \INh, where the
teacher had seen ten times more images, it added to them (\cref{sec:mechanism}), and the cfg did not favour KD there:
five of its six finals ran at cfg~2.0, where the arms without KD score worse than at their chosen 2.5.
With the released GPT-L as teacher (pre-registered; one seed per token set, credited as the GPT-L generators,
\cref{sec:supp-impl}), \NN's tokens reach 5.65: 18.0\% below \NN (credited), and level with \ours ($-2.0\%$) and
with \KDNN ($-7.2\%$).
\ours's tokens reach 5.59, level with \ours ($-3.2\%$).
Held-out at cfg~2.0 every comparison is a tie: 6.48 (\NN's tokens) and 6.64 (\ours's) against 6.79 (\NN), 6.19
(\ours) and 6.59 (\KDNN).
The VarKD study reports $-24\%$ for a GPT-L teacher with its own protocol and evaluator~\cite{peruzzo2026varkd}; our
students train from scratch for 100 epochs.
Post hoc, on \NN's tokens KD keeps about a quarter of its in-sample gain on unseen images with either teacher, \ours
about half.

\paragraph{Prior NLL and generation gains (post hoc).}
\Cref{tab:prior-predictability} compares prior and student NLL on
validation \NN sequences for \INh. A prior trained on the same images
may favour its original targets~\cite{menon2021statistical}.
On \INk the released GPT-B costs 10.97 bits: GPT-B students reach 11.07 and gain 16.4\%, whereas a GPT-L student
(343M, 100 epochs, one seed per arm) reaches 10.96 and gains 0.5\% (4.13 against 4.15); its \NN arm already samples
better than the prior (gFID 4.15 against 5.20).
Distillation behaves partly alike: a teacher far ahead of the students (1.21 bits on \INh) adds to the tokens, and on
\INk neither teacher adds a credited gain to them (GPT-B 0.10 bit ahead; GPT-L 0.38 bit ahead, one seed).
On \NN's tokens the GPT-L teacher's gain, 18.0\%, ties the GPT-B teacher's 11.7\% at one seed.
For GPT-L without guidance, \ours raises precision from 0.57 to 0.60 in the 10k sweeps at cfg~1.0,
with similar recall.
At its FID-best cfg~2.0 both arms have the same precision and recall.
GPT-B's recall there is slightly higher on the 10k sweeps (0.533 against 0.517) and level on the 50k finals (0.532
against 0.526).

\paragraph{Token surprise relative to prior entropy (pre-registered).}
For a token map $z$ of class $y$ and a prior $p$ let
$\kappa(z)=\frac{1}{256}\sum_t\big[-\log_2 p(z_t\mid z_{<t},y)-H\big(p(\cdot\mid z_{<t},y)\big)\big]$, the surprise beyond
the prior's own expectation in the same contexts; its mean is zero for the prior's own samples (measured: $+0.002$ to
$+0.011$ on 2,000 samples per prior, cfg~1.0).
Under the released GPT-B, the average conditional entropy along \ours and \NN token maps is similar
(10.80 and 10.78 bits on \INh validation images). The 0.24-bit reduction in rate comes from
the lower $\kappa$ (0.22 to $-0.03$ on \INh, and 0.23 to $-0.03$ on the full 50k ImageNet
validation set). Both values miss the pre-registered ranges ($-0.25$ to $-0.05$ for \ours and
within $\pm0.10$ for \NN). \Cref{tab:kappa} includes other priors. Under the other priors four registered predictions hold: \NN's tokens carry excess surprise also under the
released GPT-L and under \INh's \NN-trained GPT-B ($\kappa$ above 0.10; both predicted after a self-test on 50
images), a GPT-B trained on \ours's tokens does not find them typical, and \ours's tokens are less surprising than
\NN's under these two priors.
Every prior is overconfident on the \NN tokens of these unseen images ($\kappa$ 0.22--0.61), and every prior of
\cref{tab:kappa} other than the search's own on \ours's (0.39--0.49), whereas the \INh students are calibrated on
their own training tokens ($\kappa=-0.06$ for \NN's, $-0.04$ for \ours's).
On the validation images \ours's tokens are less surprising than \NN's under every prior.

\begin{table}[t]
\caption{\textbf{Surprise beyond the prior's expectation}, $\kappa$ (bits per token), of \INh's 5,000 validation
  images' tokens (pre-registered). The \INh GPT-Bs: seed 0 at epoch~25; $^{1}$\MAPo's prior. 95\% CIs within
  $\pm0.01$. The three-maps student's prior is left out. Descriptive.}
\label{tab:kappa}
\centering\footnotesize
\setlength{\tabcolsep}{3pt}
\begin{tabular}{@{}lrrr@{}}
\toprule
prior & \NN & \ours & difference \\
\midrule
released GPT-B (search's prior) & 0.22 & $-0.03$ & $-0.25$ \\
released GPT-L & 0.50 & 0.39 & $-0.11$ \\
\INh GPT-B, \NN tokens$^{1}$ & 0.49 & 0.42 & $-0.07$ \\
\INh GPT-B, \ours tokens & 0.61 & 0.49 & $-0.12$ \\
\bottomrule
\end{tabular}
\end{table}

\begin{table}[t]
\caption{\textbf{gFID-50k of generators trained on tokens chosen under different priors} (mean $\pm$ SD over three
  seeds; $^{1}$one seed per arm).
  r: released by LlamaGen (300 epochs); o: ours, trained 100 epochs on \INk's \NN tokens.
  GPT-L$^\mathrm{o}$ was tested as a prior on \INh only: tokenizing \INk with it would take an estimated 950 RTX 4090
  GPU-h, and the released priors already give the gain on \INk. n/a: \NN tokens have no prior.
  Prior bits: per \NN validation token; the \NN-trained students reach 12.22 (\INh, GPT-B), 11.07 (\INk, GPT-B) and
  10.96 (\INk, GPT-L).
  Pre-registered; tests in the text.
  Bold: the best gFID in a row; underlined: the second best.}
\label{tab:prior}
\centering\footnotesize
\setlength{\tabcolsep}{2.4pt}
\begin{tabular}{@{}lrrrr@{}}
\toprule
 & \NN & \multicolumn{3}{c}{\ours, prior:} \\
\cmidrule(l){3-5}
generator & & GPT-B$^\mathrm{r}$ & GPT-L$^\mathrm{r}$ & GPT-L$^\mathrm{o}$ \\
\midrule
\INh: prior bits & n/a & 11.01 & 10.73 & 10.99 \\
\quad GPT-B & 12.05$\pm$0.15 & \best{10.90$\pm$0.16} & 11.25$\pm$0.18 & \second{11.21$\pm$0.08} \\
\INk: prior bits & n/a & 10.97 & 10.69 & 10.96 \\
\quad GPT-B & 6.90$\pm$0.19 & \best{5.77$\pm$0.21} & \second{5.85$\pm$0.05} & -- \\
\quad GPT-L$^{1}$ & 4.15 & \second{4.13} & \best{3.95} & -- \\
\bottomrule
\end{tabular}
\end{table}

\paragraph{Stronger priors (\cref{tab:prior}).}
\Cref{tab:prior-predictability} summarizes the results for the \INh GPT-B
students. On \INk, NLL advantages of 0.10 and 0.38 bit per \NN validation
token for the released GPT-B and GPT-L priors correspond to gFID reductions
of 16.4 and 15.2\% in the pre-registered comparisons.
Every seed with a GPT-L prior lies below every \NN seed.
Held-out gFID-50k (secondary) is 16.20 (our GPT-L) and 16.30 (released GPT-L) on \INh, against 15.97 (released GPT-B)
and 16.57 (\NN); on \INk at cfg~2.0, 6.29 (released GPT-L) against 6.19 and 6.79.
Descriptively (one seed per arm; the difference between the two priors is not tested), the GPT-L student gains more
with a GPT-L prior (margin 0.27 bit) than with the GPT-B prior ($-0.01$): 4.7 against 0.5\% at the best cfg, 14.8
against 13.9\% at cfg~1.0 and 8.0 against 6.2\% at epoch~50, but held-out at cfg~2.0 both lie above \NN (5.19 and
5.26 against 5.01).
The GPT-L priors produce similar PSNR gains over \NN (1.20--1.22\,dB, compared with
1.22--1.23\,dB for the released GPT-B prior).
\paragraph{Ablations on \INh (pre-registered).}
\Cref{tab:supp-ablations} adds nine arms to \cref{tab:mech}, seven new token sets and two one-view arms on view 0 of
existing ones, each with three GPT-B seeds and the same recipe.
Ending the schedule at 35 steps keeps the gain ($-7.3\%$ against \NN, Holm~$p=0.032$; level with \MAP), so the
tokenization can be 30\% cheaper.
Of the weights tested, $\lambda=0.01$ gives the lowest gFID. Both 0.003 and 0.03 yield higher
gFID (Holm $p=0.026$ for 0.003 vs.\ 0.01). At a similar rate, omitting LPIPS-VGG doubles the
PSNR gain and raises downstream gFID by 39\% relative to \MAP. Using one view per image
raises gFID by 42\% for \NN and 47\% for \MAP. The difference between the one-view variants
is $-6.3\%$ (Holm $p=0.07$).
Three maps per view (\MAP's and two restarts of the relaxation from its initial logits plus standard normal noise;
each epoch draws one of the six per image) give 8.81: $-27\%$ against \NN and $-19\%$ against \MAP (Holm
$p=0.0001$ and 0.0003), and $-14\%$ against \MAP
held-out; on \INk they tie with \MAP (\cref{tab:crops}).
\NN tokens of LlamaGen's ten crops give 8.26 ($-17\%$ against \MAP held-out), and \ours's tokens of the same crops the
best \INh students of this paper (7.61; $-7.9\%$ against \NN on them, Holm~$p=0.022$; level held-out); the crops on
both datasets are discussed with \cref{tab:crops} in \cref{sec:crops}.
Tokens from an amortized encoder give no gain (\cref{sec:negative}).

\begin{table*}[t]
\caption{\textbf{Ablations on \INh}: GPT-B students (three seeds, mean $\pm$ SD) on token sets that change one setting of \MAP
  or of \NN (one view, ten crops).
  \MAP on ten crops is also tested against \NN on ten crops: better (Holm~$p=0.022$).
  bits: validation bits per token at epoch~100 on the arm's own tokens.
  Verdicts: Welch's test with Holm's correction within each pre-registered family; the one-view rows also compare the
  two one-view arms (\MAP, one view, against \NN, one view: $-6.3\%$, Holm~$p=0.07$, a tie); \NN, one view, against
  \NN: Welch, no correction.
  $\lambda=0.03$: \cref{tab:mech}.
  Bold: the lowest gFID-50k and held-out gFID; underlined: the second lowest; bits are on each arm's own tokens and are
  not ranked here, since across these settings (one view, ten crops, other objectives) they compare only within a
  group.}
\label{tab:supp-ablations}
\centering\footnotesize
\setlength{\tabcolsep}{4pt}
\begin{tabular}{@{}lrcrrl@{}}
\toprule
tokens & gFID-50k $\downarrow$ & cfg & held-out $\downarrow$ & bits & against \NN / \MAP (Holm) \\
\midrule
\NN & 12.05 $\pm$ 0.15 & 2.5 & 16.57 & 12.22 & -- \\
\MAP ($\lambda=0.01$, 50 steps, two views) & 10.90 $\pm$ 0.16 & 2.5 & 15.97 & 12.09 & better / -- \\
\MAP, $\lambda=0.03$ & 11.36 $\pm$ 0.05 & 2.5 & 16.55 & 11.35 & -- / worse \\
\MAP, $\lambda=0.003$ & 11.31 $\pm$ 0.11 & 2.5 & 16.24 & 12.47 & better / worse \\
\MAP, 35 steps & 11.17 $\pm$ 0.28 & 2.5 & 16.23 & 12.09 & better / tie \\
L2 only, at \MAP's rate & 15.13 $\pm$ 0.18 & 3.0 & 20.14 & 11.97 & worse / worse \\
\NN, one view & 17.08 $\pm$ 0.19 & 3.0 & 20.66 & 13.57 & worse / -- \\
\MAP, one view & 16.01 $\pm$ 0.57 & 3.0 & 20.08 & 13.43 & worse / worse \\
\MAP, three maps per view & 8.81 $\pm$ 0.22 & 2.0 & 13.75 & 11.60 & better / better \\
\NN, LlamaGen's ten crops & \second{8.26 $\pm$ 0.05} & 2.0 & \second{13.31} & 11.38 & better / better \\
\MAP, LlamaGen's ten crops & \best{7.61 $\pm$ 0.19} & 2.0 & \best{13.12} & 11.23 & -- / better \\
amortized encoder & 12.20 $\pm$ 0.16 & 2.5 & 17.12 & 11.94 & tie / worse \\
\bottomrule
\end{tabular}
\end{table*}

\section{Cost and rate--distortion}\label{sec:supp-cost}
\begin{figure*}[t]
\centering
\begin{tikzpicture}
\pgfplotsset{every axis/.append style={paper, scale only axis, width=4.15cm, height=2.75cm},
  lg/.style={black!85, mark=*, mark size=1.3pt}, ib/.style={black!45, mark=square*, mark size=1.3pt}}
\tikzset{lab/.style={font=\tiny, text=black!75, inner sep=1pt}}
\begin{axis}[name=a, xmin=0.2, xmax=1.22, ymin=0.45, ymax=1.08, xlabel={GPU time / time of 50 steps},
  ylabel={share of the PSNR gain}, title={(a) cost: steps, candidates, polish},
  legend style={at={(0.98,0.03)}, anchor=south east}]
\addplot[lg] coordinates {(0.2779,0.5419) (0.3572,0.7049) (0.5179,0.8575) (1.0000,1.0000)};
\addlegendentry{LlamaGen VQ-16}
\addplot[ib] coordinates {(0.2763,0.5148) (0.3562,0.7105) (0.5172,0.8681) (1.0000,1.0000)};
\addlegendentry{IBQ-$2^{14}$}
\addplot[only marks, mark=o, black!85, mark size=1.6pt, forget plot] coordinates {(0.9998,0.8994) (0.9999,0.9780) (0.8035,0.9985)};
\addplot[only marks, mark=square, black!45, mark size=1.6pt, forget plot] coordinates {(0.9998,0.9308) (0.9998,0.9913) (0.8048,1.0005)};
\node[lab, anchor=north west] at (axis cs:0.2779,0.5419) {5 steps};
\node[lab, anchor=north west] at (axis cs:0.3572,0.7049) {10};
\node[lab, anchor=north west] at (axis cs:0.5179,0.8575) {20};
\node[lab, anchor=south] at (axis cs:1.0000,1.0000) {50};
\node[lab, anchor=west] at (axis cs:1.02,0.8994) {4 cand.};
\node[lab, anchor=west] at (axis cs:1.02,0.9780) {8 cand.};
\node[lab, anchor=south] at (axis cs:0.8035,1.0040) {no polish};
\end{axis}
\begin{axis}[name=b, at={(a.south east)}, anchor=south west, xshift=1.35cm, xmin=-2.25, xmax=0.85, ymin=0, ymax=1.45,
  xlabel={$\Delta$ bits per token (generator)}, ylabel={$\Delta$PSNR vs \NN (dB)}, title={(b) rate--distortion over $\lambda$}]
\draw[black!35, line width=0.4pt] (axis cs:-2.25,0) -- (axis cs:0.85,0);
\draw[black!35, line width=0.4pt] (axis cs:0,0) -- (axis cs:0,1.45);
\addplot[lg] coordinates {(0.5357,1.2683) (0.2625,1.2654) (-0.2408,1.2197) (-1.0575,0.9647) (-1.7901,0.2521)};
\addplot[ib] coordinates {(0.6185,1.1810) (0.3131,1.1747) (-0.2461,1.1321) (-1.1725,0.9193) (-2.0514,0.3702)};
\addplot[only marks, mark=x, black!80, mark size=2pt] coordinates {(0,0)};
\node[lab, anchor=south west] at (axis cs:0.02,0.02) {\NN};
\node[lab, anchor=south] at (axis cs:0.5357,1.2683) {0};
\node[lab, anchor=south] at (axis cs:0.2625,1.2654) {.003};
\node[lab, anchor=south east] at (axis cs:-0.2408,1.2197) {.01};
\node[lab, anchor=south east] at (axis cs:-1.0575,0.9647) {.03};
\node[lab, anchor=west, xshift=2pt] at (axis cs:-1.7901,0.2521) {.1};
\end{axis}
\begin{axis}[name=c, at={(b.south east)}, anchor=south west, xshift=1.35cm, xmin=-2.25, xmax=0.85, ymin=0.5, ymax=2.15,
  xlabel={$\Delta$ bits per token (generator)}, ylabel={FID$_\infty$ of the reconstructions},
  title={(c) FID$_\infty$, \NN dashed}]
\draw[black!35, line width=0.4pt] (axis cs:0,0.5) -- (axis cs:0,2.15);
\draw[black!85, densely dashed, line width=0.6pt] (axis cs:-2.25,1.6882) -- (axis cs:0.85,1.6882);
\draw[black!45, densely dashed, line width=0.6pt] (axis cs:-2.25,0.9782) -- (axis cs:0.85,0.9782);
\node[lab, anchor=south east] at (axis cs:0.85,1.6882) {\NN};
\node[lab, anchor=south east] at (axis cs:0.85,0.9782) {\NN};
\addplot[lg] coordinates {(0.5357,0.8274) (0.2625,0.8610) (-0.2408,0.8782) (-1.0575,1.1761) (-1.7901,2.0171)};
\addplot[ib] coordinates {(0.6185,0.6335) (0.3131,0.6253) (-0.2461,0.6082) (-1.1725,0.6697) (-2.0514,1.0935)};
\node[lab, anchor=north] at (axis cs:0.5357,0.8274) {0};
\node[lab, anchor=south] at (axis cs:0.2625,0.8610) {.003};
\node[lab, anchor=north, yshift=-1pt] at (axis cs:-0.2408,0.8782) {.01};
\node[lab, anchor=south west] at (axis cs:-1.0575,1.1761) {.03};
\node[lab, anchor=west] at (axis cs:-1.7901,2.0171) {.1};
\end{axis}
\end{tikzpicture}
\caption{\textbf{Cost and rate--distortion of the search} on 10k ImageNet validation images, paired with \NN.
  (a)~Share of the 50-step setting's PSNR gain against GPU time (one RTX 5090): annealing steps 5--50, 4 or 8
  candidates (open markers), no polish.
  (b)~PSNR gain against the change in bits per token under each tokenizer's class-conditional generator (LlamaGen's
  GPT-B, IBQ's AR-B), over $\lambda$ (labels on
  LlamaGen's curve; IBQ's points lie close to them, in the same order).
  (c)~FID$_\infty$ of the reconstructions over the same $\lambda$; dashed: \NN (LlamaGen 1.69, IBQ 0.98).
  IBQ's FID$_\infty$ lies lower throughout and, as LlamaGen's, rises above \NN's only at $\lambda=0.1$.
  Grey level and marker distinguish the tokenizers.}
\label{fig:cost}
\end{figure*}
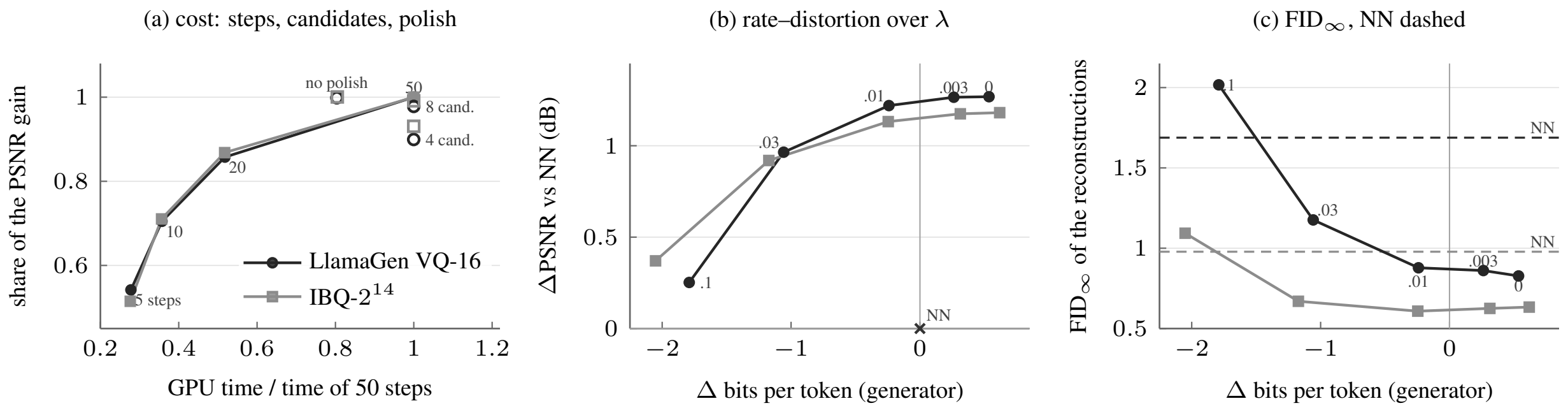

\Cref{fig:cost} reports the trade-off between encoding time, reconstruction and rate for the
settings in \cref{sec:relax,sec:solver-ablation}. On a GH200, LlamaGen \RX takes 0.55\,s per 256-px
image, including \NN and the metrics (\cref{tab:impl}). The \INh ten-crop \ours jobs with a GPT-B
prior and no refinement take 0.58\,s per crop. For \RX, 20 steps retain 86\% of the PSNR gain at
52\% of the time. For \ours, 35 steps ($\tau=0.04$) retain 99.4\% at 70\% of the time (2,000 images).
Eight candidates retain 98\% of the PSNR gain but do not reduce time appreciably, because decoder
and LPIPS passes dominate.

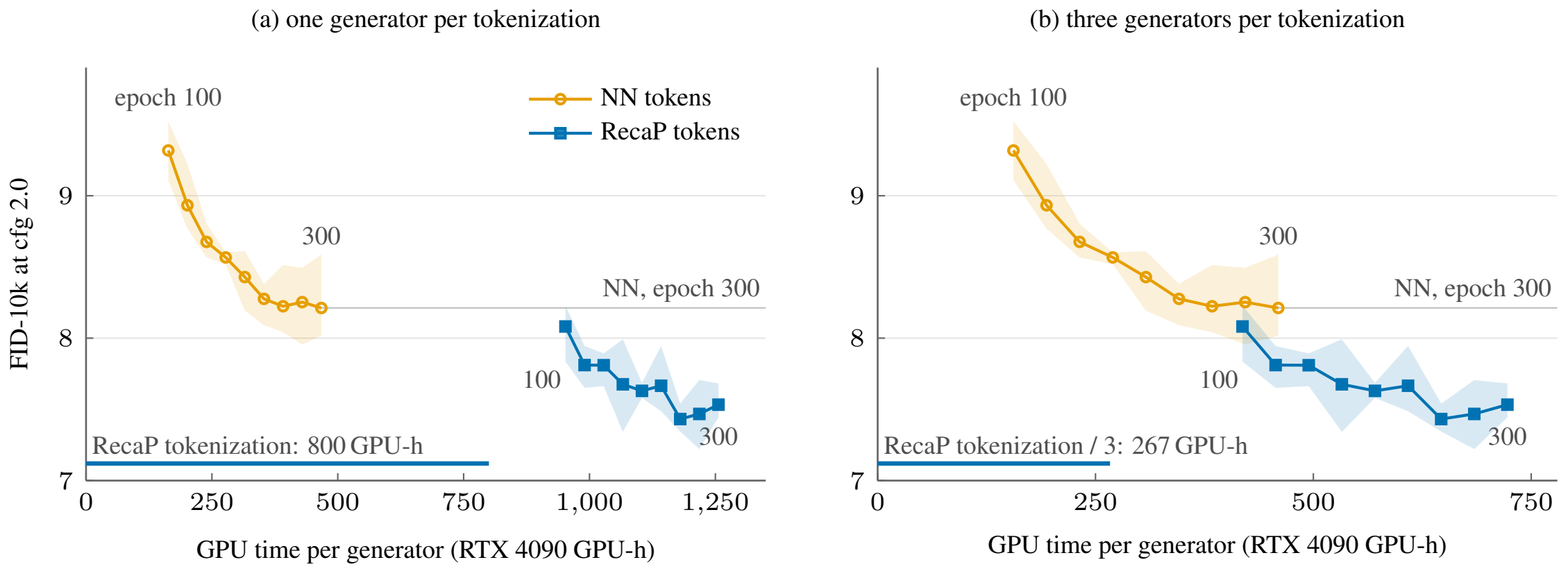
\begin{figure*}[t]
\centering
\begin{tikzpicture}
\begin{axis}[paper, name=a, scale only axis, width=2.55in, height=1.55in, ymin=7.0, ymax=9.9,
  xlabel={GPU time per generator (RTX 4090 GPU-h)}, scaled x ticks=false,
  legend style={font=\scriptsize, draw=none, fill=none, at={(0.98,0.98)}, anchor=north east}, legend cell align=left, xmin=0, xmax=1350, xtick={0,250,500,750,1000,1250}, ylabel={FID-10k at cfg~2.0},
  title={(a) one generator per tokenization}]
\addplot[fill=nncol, fill opacity=0.15, draw=none, forget plot] coordinates {(163.4,9.523) (201.4,9.223) (239.4,8.804) (277.4,8.602) (315.4,8.611) (353.4,8.381) (391.4,8.513) (429.4,8.494) (467.4,8.587) (467.4,8.017) (429.4,7.955) (391.4,8.042) (353.4,8.091) (315.4,8.193) (277.4,8.519) (239.4,8.567) (201.4,8.769) (163.4,9.109)} -- cycle;
\addplot[fill=mapcol, fill opacity=0.15, draw=none, forget plot] coordinates {(952.2,8.228) (990.2,7.945) (1028.2,7.893) (1066.2,7.992) (1104.2,7.686) (1142.2,7.946) (1180.2,7.541) (1218.2,7.706) (1256.2,7.682) (1256.2,7.444) (1218.2,7.221) (1180.2,7.344) (1142.2,7.487) (1104.2,7.580) (1066.2,7.342) (1028.2,7.663) (990.2,7.651) (952.2,7.835)} -- cycle;
\addplot[color=nncol, mark=o, mark size=1.3pt] coordinates {(163.4,9.318) (201.4,8.933) (239.4,8.676) (277.4,8.567) (315.4,8.429) (353.4,8.276) (391.4,8.224) (429.4,8.253) (467.4,8.212)};
\addlegendentry{\NN tokens}
\addplot[color=mapcol, mark=square*, mark size=1.3pt] coordinates {(952.2,8.082) (990.2,7.811) (1028.2,7.810) (1066.2,7.676) (1104.2,7.630) (1142.2,7.666) (1180.2,7.432) (1218.2,7.468) (1256.2,7.533)};
\addlegendentry{\ours tokens}
\draw[black!45, very thin] (axis cs:467.4,8.212) -- (axis cs:\pgfkeysvalueof{/pgfplots/xmax},8.212)
  node[pos=1, anchor=south east, font=\scriptsize, text=black!70, inner sep=1.5pt] {\NN, epoch~300};
\draw[mapcol, line width=1.2pt] (axis cs:0,7.12) -- (axis cs:800.2,7.12);
\node[font=\scriptsize, text=black!70, anchor=south west, inner sep=1.5pt] at (axis cs:0,7.14) {\ours tokenization: 800\,GPU-h};
\node[font=\scriptsize, text=black!70, inner sep=1pt, anchor=south] at (axis cs:163.4,9.563) {epoch~100};
\node[font=\scriptsize, text=black!70, inner sep=1pt, anchor=south] at (axis cs:467.4,8.627) {300};
\node[font=\scriptsize, text=black!70, inner sep=1pt, anchor=north east] at (axis cs:952.2,7.805) {100};
\node[font=\scriptsize, text=black!70, inner sep=1pt, anchor=north] at (axis cs:1256.2,7.404) {300};
\end{axis}
\begin{axis}[paper, name=b, scale only axis, width=2.55in, height=1.55in, ymin=7.0, ymax=9.9,
  xlabel={GPU time per generator (RTX 4090 GPU-h)}, scaled x ticks=false,
  legend style={font=\scriptsize, draw=none, fill=none, at={(0.98,0.98)}, anchor=north east}, legend cell align=left, xmin=0, xmax=780, xtick={0,250,500,750}, at={(a.south east)}, anchor=south west, xshift=0.42in,
  title={(b) three generators per tokenization}]
\addplot[fill=nncol, fill opacity=0.15, draw=none, forget plot] coordinates {(155.8,9.523) (193.8,9.223) (231.8,8.804) (269.8,8.602) (307.8,8.611) (345.8,8.381) (383.8,8.513) (421.8,8.494) (459.8,8.587) (459.8,8.017) (421.8,7.955) (383.8,8.042) (345.8,8.091) (307.8,8.193) (269.8,8.519) (231.8,8.567) (193.8,8.769) (155.8,9.109)} -- cycle;
\addplot[fill=mapcol, fill opacity=0.15, draw=none, forget plot] coordinates {(418.7,8.228) (456.7,7.945) (494.7,7.893) (532.7,7.992) (570.7,7.686) (608.7,7.946) (646.7,7.541) (684.7,7.706) (722.7,7.682) (722.7,7.444) (684.7,7.221) (646.7,7.344) (608.7,7.487) (570.7,7.580) (532.7,7.342) (494.7,7.663) (456.7,7.651) (418.7,7.835)} -- cycle;
\addplot[color=nncol, mark=o, mark size=1.3pt] coordinates {(155.8,9.318) (193.8,8.933) (231.8,8.676) (269.8,8.567) (307.8,8.429) (345.8,8.276) (383.8,8.224) (421.8,8.253) (459.8,8.212)};
\addplot[color=mapcol, mark=square*, mark size=1.3pt] coordinates {(418.7,8.082) (456.7,7.811) (494.7,7.810) (532.7,7.676) (570.7,7.630) (608.7,7.666) (646.7,7.432) (684.7,7.468) (722.7,7.533)};
\draw[black!45, very thin] (axis cs:459.8,8.212) -- (axis cs:\pgfkeysvalueof{/pgfplots/xmax},8.212)
  node[pos=1, anchor=south east, font=\scriptsize, text=black!70, inner sep=1.5pt] {\NN, epoch~300};
\draw[mapcol, line width=1.2pt] (axis cs:0,7.12) -- (axis cs:266.7,7.12);
\node[font=\scriptsize, text=black!70, anchor=south west, inner sep=1.5pt] at (axis cs:0,7.14) {\ours tokenization / 3: 267\,GPU-h};
\node[font=\scriptsize, text=black!70, inner sep=1pt, anchor=south] at (axis cs:155.8,9.563) {epoch~100};
\node[font=\scriptsize, text=black!70, inner sep=1pt, anchor=south] at (axis cs:459.8,8.627) {300};
\node[font=\scriptsize, text=black!70, inner sep=1pt, anchor=north east] at (axis cs:418.7,7.805) {100};
\node[font=\scriptsize, text=black!70, inner sep=1pt, anchor=north] at (axis cs:722.7,7.404) {300};
\end{axis}
\end{tikzpicture}
\caption{\textbf{Generation quality and compute on \INk.}
  FID-10k at cfg~2.0 against training images at epochs~100, 125, \dots, 300 (mean and range of three
  seeds). GPU hours include encoding two training views and the validation images (\NN 11.4, \ours 800.2;
  \cref{tab:compute}). We convert GPT-B training to RTX 4090 GPU hours at 1.52 per epoch, based on the
  four-GPU run (37--39\,h per 100 epochs).
  (a)~One generator per tokenization. (b)~Per-generator cost when three generators share the tokens
  (post hoc).
  In gFID-50k at the best cfg, \NN scores 6.90 $\pm$ 0.19 and 5.92 $\pm$ 0.40
  at epochs~100 and 300, and \ours scores 5.77 $\pm$ 0.21 and 5.21 $\pm$ 0.15.}
\label{fig:costgen}
\end{figure*}

\Cref{fig:costgen} compares FID-10k against \INk training images with the combined cost of encoding
and training. \ours at epoch~100 reaches 8.08, compared with 8.21 for \NN at epoch~300. At comparable training-reference FID-10k, a single generator costs 952\,GPU-h with \ours and
467\,GPU-h with \NN, including tokenization. With three runs sharing the token set, the estimated cost per generator
is 419\,GPU-h for \ours and 460\,GPU-h for \NN (post hoc).
Encoding the two training views with \ours takes 786 RTX 4090 GPU-h
(800.2 with the validation images, \cref{tab:compute}); LlamaGen's ten crops would need five times as many
encodings, about 3,900 GPU-h (\cref{tab:crops}). Training for 100 epochs with \ours rather than
300 with \NN saves an estimated 296--312 GPU-h at comparable FID-10k. These
costs give a break-even estimate of
2.5--2.7 runs sharing one token set.

\section{Alternative assignments and training procedures}\label{sec:negative}

\paragraph{Training-side estimators.}
We compare the straight-through estimator (STE) with the rotation trick (RT)~\cite{fifty2025rotation}
and a soft assignment term in LlamaGen VQ-16. The STE has the lowest EMA rFID-50k after ten epochs:
3.07, compared with 3.20 for RT, 3.29 for RT with soft assignment, and 3.43 for STE with soft
assignment. Each variant uses three seeds.

\paragraph{Amortizing the search.}
On LlamaGen,
a learned five-step optimizer recovers 59\% of the PSNR gain and five Adam steps recover 54\%.

We pre-register an amortized search on \INh. An 88M-parameter network picks one of the 16 candidates
per position in two rounds. It trains first to imitate \ours and then to minimize its objective.
It recovers 83\% of \ours's PSNR gain, reduces the rate 1.09 times as much as \ours does relative to \NN, and needs
0.018\,s per two-view image (\ours: 1.20\,s), with no \lpipsA gain. GPT-B trained on its tokens scores
$12.20\pm0.16$ in gFID-50k, 1.3\% above \NN (Holm $p=0.29$) and 12\% above \MAP (Holm $p=0.001$).

Training with the search in the loop diverges or lowers reconstruction PSNR (21.5 vs.\ 25.1\,dB
on CIFAR). Regressing the encoder onto searched codes also lowers held-out \NN PSNR.

\paragraph{Reconstruction-only search baselines.}
For the matched-budget comparison in \cref{sec:recon}, straight-through
refinement optimizes the encoder output through quantization. The other
baseline refines the continuous latent with Adam and applies \NN at the
end. Both use reconstruction loss alone. Their rates rise by 0.22 and
0.48 bit per token relative to \NN, respectively. The relaxation in
this comparison omits its final discrete refinement, which would add 0.002\,dB.

\paragraph{Re-fitting the encoder.}
On 100 LlamaGen validation images, a short encoder refit without a GAN term reduces the PSNR
gap from 1.25 to 0.43\,dB, close to the gap for a GAN-free tokenizer. With the refit, \lpipsA rises by 2\% on
LlamaGen and 9\% on IBQ-$2^{18}$. Reconstruction-only reassignment improves PSNR with no
corresponding increase in \lpipsA in this comparison.

\paragraph{A masked generator.}
We test a MaskGIT-style generator using GPT-B's network, masked training and eight-step parallel
decoding (pre-registered, 100 epochs on \INh, three seeds per variant). With \ours tokens,
gFID-50k is $17.95\pm0.73$ versus $17.72\pm0.48$ for \NN ($+1.3\%$, Holm $p=0.95$), and
held-out FID is 22.36 versus 22.02. Validation NLL decreases by 0.06 bit per token ($p=0.0004$).
Precision falls (0.648 vs.\ 0.678) while recall rises (0.280 vs.\ 0.245). The lower validation NLL does
not yield an FID improvement for this masked generator.

\paragraph{Soft targets stored once.}
We store the top 32 next-token probabilities of the released GPT-B on the final token maps
using one teacher-forced pass (3.8\,min on a GH200). We then train with these soft targets on
\INh (pre-registered). Renormalizing over the 32 tokens raises gFID-50k to 19.34 on \NN tokens
and 19.94 on \ours tokens, compared with 12.05 and 10.90 under standard training. Assigning
the missing probability mass to the true token gives scores closer to standard training (11.67
and 11.10). The top 32 tokens cover only 20\% of the teacher's probability mass, which limits
what the stored targets can retain.

\end{document}